\documentclass{article}

 \usepackage[preprint]{neurips_2026}

\usepackage[utf8]{inputenc} 
\usepackage[T1]{fontenc}    
\usepackage{hyperref}       
\usepackage{url}            
\usepackage{booktabs}       
\usepackage{amsfonts}       
\usepackage{nicefrac}       
\usepackage{microtype}      
\usepackage{xcolor}         

\title{BBOWP-Bench: Evaluating LLMs on Black-Box Optimization Word Problems}

\author{%
  Yutaro Yamada\thanks{Equal contribution.}
  \quad
  Kei Hiroshima\footnotemark[1]
  \quad
  Nozomu Yoshinari\footnotemark[1]
  \\
  \textbf{Kento Uchida}
  \quad
  \textbf{Shinichi Shirakawa} \\
  Yokohama National University \\
  \texttt{\{yamada-yutaro-dw,hiroshima-kei-st,yoshinari-nozomu-ry\}@ynu.jp} \\
  \texttt{\{uchida-kento-fz,shirakawa-shinichi-bg\}@ynu.ac.jp}
}

\usepackage[whole]{bxcjkjatype}
\usepackage{booktabs}
\usepackage{multirow}
\usepackage{colortbl}
\usepackage{url}
\usepackage{graphicx}
\usepackage{amsmath}
\usepackage{amssymb}
\usepackage{amsfonts}
\usepackage{mathtools}
\usepackage{amsthm}
\usepackage{stmaryrd}
\usepackage{tabularx}
\usepackage{comment}
\usepackage{wrapfig}

\usepackage[capitalize,noabbrev]{cleveref}
\makeatletter
\AddToHook{cmd/appendix/before}{%
  \def\cref@section@alias{appendix}%
  \def\cref@subsection@alias{subappendix}%
  \def\cref@subsubsection@alias{subsubappendix}}
\makeatother

\usepackage{tabularx}
\usepackage{array}

\usepackage[many]{tcolorbox}
\usepackage{enumitem}
\usepackage{ifthen}
\usepackage{xcolor}

\newcommand{\markupdraft}[2]{
    \ifthenelse{\equal{#1}{display}}{#2}{}
    \ifthenelse{\equal{#1}{color}}{\color{#2}}{}
}

\newcommand{\newcolored}[3][]{{\markupdraft{color}{#2}#3}
    \ifthenelse{\equal{#1}{}}{}{\markupdraft{display}{{\color{yellow!70!black}[#1]}}}}

\newcommand{\argmin}{\mathop{\rm argmin}\limits}

\newcommand{\E}{\mathbb{E}}

\makeatletter
\newcommand{\unbreakcitealp}[1]{\mbox{\citealp{#1}}}
\newcommand{\natbibsep}{\NAT@sep\space}
\makeatother

\begin{document}

\maketitle

\begin{abstract}
Formulating an optimization problem strongly affects the quality of the final solution, yet good formulations usually require substantial expertise.
Recent studies have therefore examined how to automatically derive optimization problems from natural-language descriptions, but existing benchmarks focus on settings where objectives and constraints can be written explicitly as mathematical expressions.
Many practically important problems, including hyperparameter optimization in machine learning, are naturally treated as black-box optimization (BBO) problems, in which only objective values are observable, and the functional form is unavailable. In BBO, the search space design, a part of the problem formulation, and the selection of the optimization algorithm are crucial for problem-solving. Automating these processes with large language models (LLMs) is a significant challenge.
This paper introduces Black-Box Optimization Word Problems (BBOWP), a novel problem setting in which a system must infer both a search space and an optimization algorithm from a natural-language description of a black-box optimization task.
To support research on this setting, we establish the BBOWP Benchmark Suite (BBOWP-Bench), a dataset and evaluation framework for BBOWP\@.
Each instance combines a natural-language problem description, an executable evaluation environment, and a human-designed reference formulation, allowing evaluation of both search-space design and algorithm selection.
The benchmark covers 23 instances from four different application domains and supports reproducible assessment through actual black-box optimization runs.
Using this benchmark, we evaluate LLMs on the joint formulation of search spaces and optimizers from natural-language descriptions across multiple BBO domains.
We show that current LLMs vary their optimizer choice with the evaluation budget, though they sometimes struggle with search space design, particularly in identifying important variables and balancing their ranges when the problem description is less informative or the search space is highly problem-specific.
Our code and dataset are available at \url{https://github.com/shiralab/bbowp-bench}.
\end{abstract}

\begin{figure}[t]
    \centering
    \includegraphics[width=\textwidth]{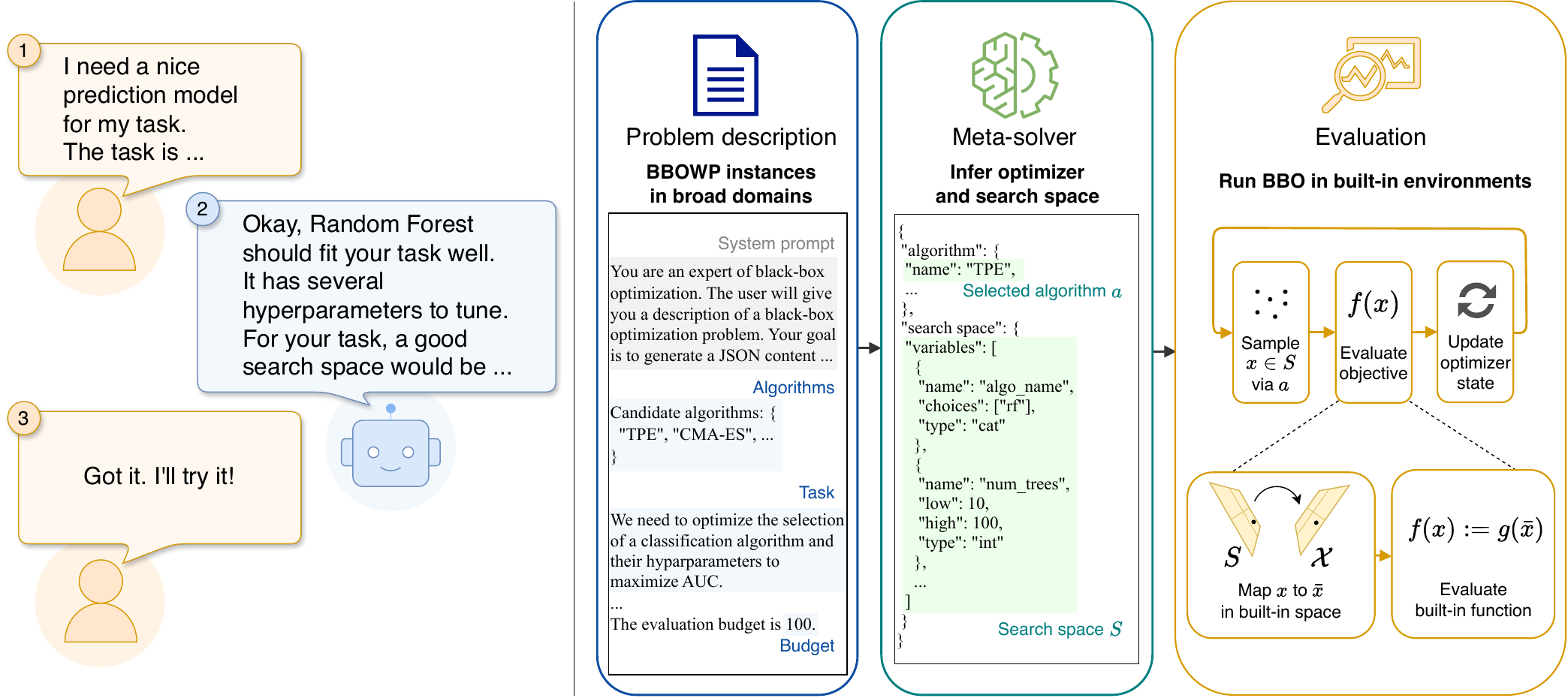}
    \caption{
        \textbf{(Left)} A typical scenario considered in BBOWP\@.
        Users ask a meta-solver to formulate a BBO problem for their task, but they may not have an implemented objective function when they describe the task.
        \textbf{(Right)} Overview of the BBOWP Benchmark Suite.
        Given a problem description $d$ that includes information such as the maximum number of function evaluations $N_{\mathrm{max}}$, a meta-solver determines an appropriate optimization algorithm $a$ and search space $S$.
        Then, the formulation is evaluated through actual optimization runs, using a built-in objective function via identifier mapping.
    }
    \label{fig:overview}
\end{figure}

\section{Introduction}

Formulating an optimization problem has a major impact on solution quality.
However, identifying the key elements of a problem, such as design variables and objectives, and then expressing them in a solver-ready form usually requires substantial expertise.
Prior work has therefore studied the automatic formulation of optimization problems from natural-language descriptions.
Recent datasets and benchmarks cover a range of optimization classes, from linear programming settings such as NL4Opt \citep{ramamonjison2022nl4opt} to more recent resources for nonlinear optimization \citetext{\unbreakcitealp{huang2025mamo}\natbibsep\unbreakcitealp{yang2024optibench}\natbibsep\unbreakcitealp{huang2025orlm}\natbibsep\unbreakcitealp{lu2025optmath}}, and support the evaluation on tasks such as extracting variables, objectives, and constraints from text.

This line of work has produced useful datasets and evaluation settings, but existing benchmarks still target problem classes in which objectives and constraints can be written explicitly as mathematical expressions.
In contrast, many important real-world tasks, including hyperparameter optimization, molecular design, and structural shape optimization, are more naturally cast as black-box optimization (BBO) problems.
In BBO, only objective values are observable, and the underlying functional form is unknown or intractable.
As a result, progress in auto-formulation for BBO requires not only new methods but also benchmark datasets that make formulation quality and downstream optimization performance measurable.

In parallel, it has been argued that there is a need for metadata-rich BBO benchmarks \citep{song2024position}.
Existing BBO benchmarks, such as COCO \citep{coco}, contain limited metadata describing the problem context.
However, in practice, utilizing prior knowledge is essential to realize efficient optimization, and leveraging large language models (LLMs) to exploit such information can be a key direction.
This viewpoint also motivates the development of a BBO benchmark with task information written in natural language.

Building on these research streams, we introduce Black-Box Optimization Word Problems (BBOWP), a novel problem setting for auto-formulation in BBO domains.
In BBOWP, a system receives a natural-language problem description and must infer both a search space and a suitable optimization algorithm.
To accelerate research on BBOWP, we develop the BBOWP Benchmark Suite (BBOWP-Bench), which pairs natural-language problem descriptions with design-variable domain definitions, executable evaluation environments, and human-designed reference formulations.
We also design an evaluation protocol that supports both comparison against a human reference and assessment through actual BBO runs.
Using this framework, we take a first step toward evaluating several LLMs in BBOWP.
We find that they can select suitable algorithms for the evaluation budget when provided with detailed problem descriptions. However, they often struggle to identify important variables and set appropriate ranges when problem descriptions are less informative.

Our contributions are summarized as follows:
(1) We propose BBOWP, a new problem setting in which a system must infer both a search space and a suitable optimization algorithm for a BBO task from its natural-language description.
(2) We introduce the BBOWP-Bench, a dataset for BBOWP that pairs natural-language problem descriptions with executable evaluation environments and human-designed reference formulations.
(3) We design an evaluation protocol that supports both formulation-level comparison and downstream assessment through actual BBO runs.
(4) We provide the first benchmark study of several LLMs on BBOWP and show that several task settings remain challenging for current LLMs.

\section{Related Work}

Constructing solver-ready formulations from natural-language problem descriptions requires substantial expertise.
To reduce this reliance on experts, recent work has developed datasets and benchmarks for auto-formulation of optimization problems \citep{jiang2025llmopt,ramamonjison2022augmenting,ramamonjison2022nl4opt,xiao2023complexor,ahmaditeshnizi2024nlp4lp,huang2025mamo,yang2024optibench,huang2025orlm,lu2025optmath}.
Most of these studies focus on linear programming, integer programming, and mixed-integer linear programming \citep{ramamonjison2022augmenting,ramamonjison2022nl4opt,huang2025mamo}, with more recent extensions to nonlinear programming \citep{huang2025orlm,yang2024optibench,lu2025optmath}.
Despite this progress, many real-world optimization problems do not admit explicit symbolic objectives.
Hyperparameter optimization, expensive simulations, materials discovery, and human-in-the-loop design are all naturally modeled as BBO problems.
Our work extends the auto-formulation perspective to this broader setting.

Other studies have applied LLMs to BBO.
\citet{Liu2024-gu}, \citet{zhao2026}, and \citet{ngo2026} leverage LLMs to enhance Bayesian optimization (BO) for controlling a part of optimization: constructing surrogate models, configuring acquisition strategies, and sampling candidates.
\citet{Liu2025-ti} introduce a framework that leverages LLMs for hyperparameter optimization.
\citet{brunzema2026} and \citet{mahammadli2025} let LLMs refine the search space adaptively during optimization.
While these studies assume that design variables and their initial domains are given and restrict the optimizer to BO-based methods, BBOWP requires LLMs to infer the setup from a natural-language description.

Closely related but distinct text-to-optimization studies include \citep{li2025solver} and \citep{imajuku2026alebench}.
\citet{li2025solver} studied automated code generation for a part of the BBO formulation from natural-language descriptions, where the code computes objectives and constraints from simulation outputs.
\citet{imajuku2026alebench} introduced ALE-Bench, which evaluates whether agents can improve performance on hard optimization tasks from task descriptions.
In contrast, BBOWP focuses on an earlier formulation stage: given only a natural-language description, the system must infer not only a suitable search algorithm but also what search space should be exposed to the optimizer.
Our work is therefore complementary to these studies as part of an end-to-end automated pipeline for BBO.

Algorithm selection is a central issue in BBO.
Because the suitable optimizer depends on the search space, objective function landscape, and evaluation budget, the optimizer should be carefully chosen to match the task formulation rather than fixed in advance \citep{munoz2015algorithm,kerschke2019automated}.
BBOWP is closely related to this issue because solving a BBOWP instance requires inferring a search space and selecting an optimizer that is compatible with it.

\section{Black-Box Optimization Word Problem (BBOWP)}\label{sec:BBOWP}

BBOWP is formulated for settings in which the problem description may be written by a non-expert, so the underlying objective function is not assumed to be fully specified beforehand.
Hence, the users may prepare their objective function \emph{after} receiving an answer to BBOWP as shown in \cref{fig:overview} (left).
Because any maximization problem can be converted into a minimization problem by minimizing the negative of the objective, we consider only minimization in what follows.
In this paper, we consider unconstrained single-objective BBO problems.
Extending BBOWP to constrained and multi-objective optimization is left for future work.

In BBOWP, the input is a natural-language problem description $d$.
The description includes information such as the problem background, the target performance metric, the maximum number of function evaluations $N_{\mathrm{max}}$, and the set of available algorithms $\mathcal{A}$.
Because the instruction may be written at a high level, the design variables are not always stated explicitly in $d$.

The output for BBOWP is a search space $S \in \mathcal{S}$ and an optimization algorithm $a \in \mathcal{A}$.
Here, $\mathcal{S}$ denotes the class of admissible search spaces, and $S$ denotes a formulation of the search space suitable for the target BBO problem described in $d$. For instance, the system for solving BBOWP determines the number, names, types, and ranges of design variables as the search space design, aligning with the problem description.
We include algorithm selection in the output because the suitability of an optimizer depends on the problem formulation and the available evaluation budget.

The quality of the answer $(a, S)$ can be measured by the final solution quality when running the algorithm $a$ to minimize the objective function on the search space $S$ within the given budget $N_\mathrm{max}$.
Namely, BBOWP asks for an optimizer and a search space that minimize the expected objective value attained within the given budget:
$(a^*, S^*) = {\argmin}_{a \in \mathcal{A},\, S \in \mathcal{S}} \,\, \E \left[ \mathcal{V}(a, S, N_{\mathrm{max}}) \right]$,
where $\mathcal{V}(a, S, N_{\mathrm{max}})$ denotes the best objective value obtained by running algorithm $a$ under the search space $S$ with budget $N_{\mathrm{max}}$, evaluated on the underlying performance metric intended by $d$.
Because black-box optimizers are typically stochastic, $\mathcal{V}$ is generally a random variable.
We study the performance of policies that output a suitable optimization algorithm $a$ and a search space $S$ for a given input $d$.
We refer to such a policy as a meta-solver.

\section{BBOWP Benchmark Suite} \label{sec:bbowp_bench}

The input of BBOWP is a problem description written in natural language.
Existing BBO benchmark datasets such as COCO \citep{coco} and NeuroEvoBench \citep{neuroevo} do not include natural-language descriptions of the problems, which makes them unsuitable for evaluating performance on BBOWP\@.
To address this limitation, we propose the BBOWP-Bench, a benchmark dataset designed specifically for the BBOWP task.
An overview of the framework is shown in \cref{fig:overview} (right).

\subsection{Benchmark Design} \label{sec:benchmark_design}

At the abstract level, BBOWP does not assume that the search space is fixed in advance.
Therefore, a meta-solver is free to design the search space, deciding which variables to optimize and how to identify them, based on the natural-language description and its intended optimization objective.
Ideally, the evaluation should be able to assess any search space generated by a meta-solver.
In principle, this could be done by having a human interpret the meta-solver's output and implement the corresponding objective interface for that formulation.
However, such a procedure is impractical for a benchmark that must rely on pre-built execution environments and support automatic evaluation.
The central design choice of the BBOWP-Bench is therefore to approximate this ideal evaluation protocol using a fixed executable objective together with an adapter that maps generated search spaces to the variable space expected by the executable environment.

The benchmark provides pre-built executable objectives and their problem descriptions derived from existing real-world BBO benchmarks (see Section~\ref{sec:data_description}).
For a given description $d$, the objective $g_d : \mathcal{X}_d \to \mathbb{R}$ is defined on a sufficiently large canonical design space $\mathcal{X}_d = \mathcal{X}_{l_1} \times \dots \times \mathcal{X}_{l_I}$.
Each component is indexed by an identifier $l_i$ representing the semantic role of a design variable. For example, one such identifier may be \texttt{learning\_rate}.
The benchmark also specifies default values $x_{l_i}^{\mathrm{def}} \in \mathcal{X}_{l_i}$.

By contrast, a meta-solver outputs a search space $S = S_{l'_1} \times \dots \times S_{l'_J}$ together with an optimization algorithm.
Because $\mathcal{X}_d$ is not always given to the meta-solver directly, the generated identifiers $l'_j$ need not coincide with the canonical identifiers $l_i$.
The meta-solver may omit relevant variables, introduce irrelevant ones, or use alternative names for the canonical executable names.
Each generated variable additionally specifies its type, domain, and choice of sampling scale, either linear or logarithmic. If a logarithmic scale is selected, the optimization algorithm searches in the transformed space, and the realized value is mapped back before evaluation.
The domain is defined by the upper and lower bounds of the search range for a design variable or by a singleton set for a constant value, or by a set of choices for a categorical variable.

To evaluate a point $x = (x_{l'_1}, \dots, x_{l'_J}) \in S$, the benchmark resolves correspondence between generated and canonical variables through an equivalence relation $\sim$ over identifiers. For each canonical variable $l_i$, let $M_i = \{ j \in \{1, \dots, J\} \mid l_i \sim l'_j \}$ denote the set of generated variables matched to $l_i$. When this correspondence is not ambiguous, the benchmark constructs a canonical point $\bar{x} = (\bar{x}_{l_1}, \dots, \bar{x}_{l_I}) \in \mathcal{X}_d$ by
\begin{equation}
  \bar{x}_{l_i} =
  \begin{cases}
    x_{l'_j} & \text{if } M_i = \{j\} \enspace , \\
    x_{l_i}^{\mathrm{def}} & \text{if } M_i = \emptyset
  \end{cases} \enspace .
\end{equation}
The benchmark then evaluates the generated point through the induced wrapper objective
\begin{equation}
  f_d(x) =
  \begin{cases}
    g_d(\bar{x}) & \text{if } |M_i| \le 1 \text{ for all } i \text{, and if } x \text{ can be cast to } \bar{x} \in \mathcal{X}_d \\
    \omega_d & \text{otherwise}
  \end{cases} \enspace ,
  \label{eq:wrapper_obj}
\end{equation}
where $\omega_d$ denotes the worst objective value for task $d$ ($\infty$ for minimization and $-\infty$ for maximization).
Here, omitted variables are filled with task-specific default values, and irrelevant generated variables are ignored unless they create an ambiguity in the correspondence. 
We note that the worst objective value is returned when the canonical point $\bar{x}$ lies outside the canonical design space $\mathcal{X}_d$, i.e., $\bar{x} \notin \mathcal{X}_d$.

This construction is the key design choice in the BBOWP-Bench.
It enables evaluation of a broad range of generated search spaces using a fixed executable objective $g_d$.
In this sense, the benchmark approximates the ideal flexible objective needed by the abstract task definition without requiring a new execution environment for every generated formulation.
Further details of the benchmark design are discussed in \cref{app:design}.

BBOWP-Bench also includes the list of candidate black-box optimizers. The system for solving BBOWP is expected to select a suitable one from the candidate list in addition to designing the search space. Then, the output, a pair of a selected algorithm and a designed search space $(a, S)$, can be automatically evaluated by running the algorithm $a$ on the wrapper objective \eqref{eq:wrapper_obj} with $S$.

\subsection{Data Description}
\label{sec:data_description}

\begin{wraptable}{r}{0.65\textwidth}
  \vspace{-3.5\baselineskip} 
  \centering
  \caption{Examples of the instances in BBOWP-Bench.}
  {\small
    \setlength{\tabcolsep}{1pt}
    \renewcommand{\arraystretch}{0.95}
    \begin{tabular}{l c >{\raggedright\arraybackslash}p{0.25\columnwidth} c c c c}
      \toprule
      \textbf{Source} & \textbf{Task} & \textbf{Scenario} & \textbf{Dim} & \textbf{Float} & \textbf{Int} & \textbf{Cat} \\
      \midrule
      YAHPO & 3 & HPO of Ranger & 8 & $\checkmark$ & $\checkmark$ & $\checkmark$ \\
      & 5 & HPO of SVM & 6 & $\checkmark$ & $\checkmark$ & $\checkmark$ \\
      \midrule
      Olympus & 14 & S${}_N$Ar reaction & 4 & $\checkmark$ &  & \\
      & 15 & HPLC system & 6 & $\checkmark$ &  & \\
      \midrule
      MECHBench & 17 & Star-shaped crash box & 5 & $\checkmark$ &  & \\
      \midrule
      MuJoCo & 19 & Inverted Pendulum & 5 & $\checkmark$ &  & \\
      & 20 & Inverted Double Pendulum & 10 & $\checkmark$ &  & \\
      \bottomrule
    \end{tabular}
  }
  \label{tab:instance}
\end{wraptable}

We created BBOWP instances from four sources: YAHPO Gym~\citep{yahpo}, Olympus~\citep{olympus_2021, olympus_2023}, MECHBench~\citep{mechbench}, and MuJoCo-based continuous-control tasks~\citep{todorov2012mujoco}.
YAHPO Gym is a benchmark dataset for hyperparameter optimization (HPO) of machine learning models.
Olympus covers experimental science domains, such as materials science and drug discovery, and consists of tasks that emulate real-world physical experiments and chemical simulations.
MECHBench is a benchmark dataset for shape optimization in structural mechanics and consists of tasks that involve designing structures modeled after vehicle components to achieve high crash-safety performance.
MuJoCo provides physics-based continuous-control tasks in simulated robotic environments.

From these sources, we built 7 tasks from YAHPO Gym, 9 from Olympus, 2 from MECHBench, and 5 from MuJoCo, yielding 23 tasks in total.
We carefully extract the appropriate tasks for BBOWP-Bench by excluding those with ambiguous variable names (HPO of the glmnet scenario in YAHPO) and those with implementation problems inconsistent with the paper (Alkoxylation scenario in Olympus).
Examples of scenarios and design-variable types are shown in \cref{tab:instance} (see \cref{tab:all_instance} for complete details).
The execution environment for each benchmark is prepared as a separate Docker image for each problem domain.

For each problem instance, we prepare a JSON record containing the canonical variables, their domains, and default values used by the executable environment.
We also prepare three problem descriptions with different difficulty levels: \textit{Easy}, \textit{Intermediate}, and \textit{Hard}.
The \textit{Easy} description explicitly includes an explanation of the full canonical design space, the \textit{Intermediate} description includes partial clues from which the canonical design space can be inferred, and the \textit{Hard} description contains no direct information about the canonical design space and instead describes only the task background and optimization objective.

To define the equivalence relation $\sim$ among identifiers, we also create a list of semantically equivalent names for each design variable.
For example, for a design variable \texttt{learning\_rate}, we include synonymous identifiers such as \texttt{lr} and \texttt{step\_size}.
We evaluate the list of semantically equivalent names by analyzing the effect on the optimization results and quantifying the coverage of the variable names included in the search space designed by the meta-solvers in \cref{app:alias_mapping}.

Each task is accompanied by a search space and an optimization algorithm selected and verified by multiple BBO experts as the human reference.
It also provides the reasons behind the choice of variables, ranges, scales, and optimizer under different settings for evaluation budget: \textit{Small} with $N_{\mathrm{max}}=100$, \textit{Large} with $N_{\mathrm{max}}=10{,}000$, where $N_{\mathrm{max}}$ is the maximum number of function evaluations.
The experts designed the search space to avoid missing important variables and to balance its size with the given budget, and selected the optimization algorithm based on variable types and budget. Note that these human references were created by referring to the \textit{Easy} description.
This information makes the benchmark useful as a reference baseline, a source of few-shot examples, and a basis for further analysis.
We emphasize that the expert formulation is not intended to represent the optimal answer.
Rather, it serves as a high-quality reference for validation and analysis.
See \cref{app:input_output} for details of the input and output of BBOWP-Bench.

\section{Experiments} \label{sec:experiment}

We assess the performance of representative LLMs as meta-solvers and demonstrate the usefulness of BBOWP-Bench.

\subsection{Experimental Setup} \label{sec:exp_setup}

\paragraph{Evaluated Meta-Solvers.}
To investigate the impact of model size and reasoning capability on BBOWP, we evaluated four representative LLMs: Gemini 3 Flash \citep{google2025gemini3flash}, Gemini 3.1 Pro \citep{google2026gemini31pro}, GPT-5 mini \citep{openai2025gpt5systemcard}, and GPT-5.2 \citep{openai2025gpt52systemcard}.

\paragraph{Prompting and Difficulty Levels.}
For each problem, the LLM receives a prompt containing an overview of the BBOWP task, the required JSON output format, the problem description, and the target maximum number of function evaluations (budget).
To rigorously assess the models' capabilities to build a search space, we evaluated the LLMs across three difficulty levels of problem descriptions: \textit{Easy}, \textit{Intermediate}, and \textit{Hard} (see Section~\ref{sec:data_description}).
We evaluated them in both \textit{zero-shot} and \textit{one-shot} settings and discuss the results of the \textit{zero-shot} setting in this section.
The results of the \textit{one-shot} setting are provided in \cref{app:experiment:oneshot}.
Complete prompt templates are provided in \cref{app:prompt}.
We do not evaluate sensitivity to prompt rephrasings, which remains a limitation of the current evaluation.

\paragraph{Candidate Algorithms and Budget Conditions.} 
The LLMs are instructed to select an optimization algorithm from a predefined pool of standard optimizers for BBO: CMA-ES~\citep{hansen:cma}, CMA-ES with Margin (CMA-ESwM)~\citep{hamano:cmaeswm}, CatCMA~\citep{hamano:catcma}, CatCMA with Margin (CatCMAwM)~\citep{hamano:catcmawm}, Differential Evolution (DE)~\citep{1997:de}, Nelder-Mead~\citep{1965:nelder}, Particle Swarm Optimization (PSO)~\citep{1995:pso}, Bayesian optimization with Gaussian process regression (GP)~\citep{1998:gp}, SMAC3~\citep{smac3}, and Tree-structured Parzen Estimator (TPE)~\citep{2011:tpe}.
We query the LLMs to design search spaces and select algorithms for two distinct budgets.

\paragraph{Execution of Optimization.}
For the end-to-end evaluation through actual optimization runs, we use the predefined script to execute the optimization according to the JSON specification generated by an LLM\@.
The optimization script is implemented using the OptunaHub library~\citep{2026:optunahub}.
While the optimization runs on the generated search space, actual function evaluations are performed using the built-in function as described in \cref{sec:benchmark_design}.
We report the mean and standard deviation of the best evaluation values obtained over three optimization runs.
For \textit{Small} budget settings, the optimization runs are executed across all 23 tasks.
For \textit{Large} budget settings, the optimization runs are executed across 21 tasks other than the MECHBench tasks (Tasks 17 and 18) since each evaluation takes a minute or more in those tasks.

\paragraph{LLM Generation Settings.}
We set the temperature to $1$ for all models and generate one output for each experimental condition.
We additionally analyze the effect of variability across the LLM generations in \cref{app:genvar}.

\subsection{Behavioral Analysis of LLMs as Meta-Solvers}
\label{sec:results_optimization}

\begin{table}
\centering
\small
\caption{Comparison of optimization performance in \textit{Easy} and \textit{Hard} difficulties with \textit{Small} budget. We present the results for several tasks here; the complete results are shown in \cref{app:experiment:zeroshot}. Each cell reports the mean and standard deviation of the best evaluation values obtained over three optimization runs.}
\label{tab:results_optimization_score_diff}
\setlength{\tabcolsep}{5pt}
\begin{tabular}{@{}llrrrrr@{}}
\toprule
& & \multicolumn{2}{c}{YAHPO} & \multicolumn{2}{c}{Olympus} & MECHBench \\
\cmidrule(lr){3-4} \cmidrule(lr){5-6} \cmidrule(lr){7-7}
Difficulty & Model & Task 3 ($\uparrow$) & Task 5 ($\uparrow$) & Task 14 ($\downarrow$) & Task 15 ($\uparrow$) & Task 17 ($\downarrow$) \\
\midrule
\multirow{4}{*}{\textit{Easy}} & Gemini 3 Flash  & 0.94 {\scriptsize $\pm$ 0.02} & \textbf{0.98} {\scriptsize $\pm$ 0.00} & \textbf{0.19} {\scriptsize $\pm$ 0.00} & \textbf{2521.76} {\scriptsize $\pm$ 155.43} & \textbf{-12044.69} {\scriptsize $\pm$ 548.96} \\
 & Gemini 3.1 Pro  & 0.86 {\scriptsize $\pm$ 0.00} & 0.95 {\scriptsize $\pm$ 0.01} & 0.20 {\scriptsize $\pm$ 0.01} & 2436.03 {\scriptsize $\pm$ 64.52} & \textbf{-12044.69} {\scriptsize $\pm$ 548.96} \\
 & GPT-5 mini & 0.94 {\scriptsize $\pm$ 0.00} & 0.97 {\scriptsize $\pm$ 0.01} & 0.32 {\scriptsize $\pm$ 0.02} & 2282.10 {\scriptsize $\pm$ 32.82} & -10823.81 {\scriptsize $\pm$ 682.55} \\
 & GPT-5.2 & \textbf{0.95} {\scriptsize $\pm$ 0.01} & 0.95 {\scriptsize $\pm$ 0.01} & \textbf{0.19} {\scriptsize $\pm$ 0.01} & 2277.24 {\scriptsize $\pm$ 33.38} & -10804.95 {\scriptsize $\pm$ 709.63} \\
\midrule
\multirow{4}{*}{\textit{Hard}} & Gemini 3 Flash  & \textbf{0.88} {\scriptsize $\pm$ 0.00} & \textbf{0.95} {\scriptsize $\pm$ 0.00} & 0.91 {\scriptsize $\pm$ 0.01} & $-\infty$ & \textbf{-8528.06} {\scriptsize $\pm$ 6.86} \\
 & Gemini 3.1 Pro  & 0.82 {\scriptsize $\pm$ 0.00} & \textbf{0.95} {\scriptsize $\pm$ 0.01} & 1.56 {\scriptsize $\pm$ 0.75} & $-\infty$ & -8471.56 {\scriptsize $\pm$ 98.17} \\
 & GPT-5 mini & 0.84 {\scriptsize $\pm$ 0.00} & 0.93 {\scriptsize $\pm$ 0.01} & 0.80 {\scriptsize $\pm$ 0.01} & $-\infty$ & $\infty$ \\
 & GPT-5.2 & 0.82 {\scriptsize $\pm$ 0.00} & \textbf{0.95} {\scriptsize $\pm$ 0.00} & \textbf{0.78} {\scriptsize $\pm$ 0.01} & \textbf{1311.06} {\scriptsize $\pm$ 27.18} & -8525.20 {\scriptsize $\pm$ 15.63} \\
\midrule
\multicolumn{2}{c}{Human Reference} & 0.89 {\scriptsize $\pm$ 0.03} & 0.98 {\scriptsize $\pm$ 0.01} & 0.20 {\scriptsize $\pm$ 0.01} & 2455.28 {\scriptsize $\pm$ 79.88} & -11513.26 {\scriptsize $\pm$ 534.10} \\
\bottomrule
\end{tabular}
\end{table}

\paragraph{Sensitivity to Problem Domains and Difficulties.}

A critical requirement for a meta-solver is the ability to infer the variables to be optimized when problem descriptions are ambiguous.
\cref{tab:results_optimization_score_diff} presents the optimization performance in \textit{Easy} and \textit{Hard} difficulties with \textit{Small} budget.
In the \textit{Easy} setting, where the full canonical design spaces are explicitly provided in the prompt, most models act as highly capable instruction followers, successfully mapping the text to valid search spaces across all domains.
However, in the \textit{Hard} setting, where models must infer the design variables and bounds entirely from an ambiguous background description, a significant gap across application domains emerges.
For Tasks 3 and 5 in HPO (YAHPO Gym), we observed that LLMs successfully reconstruct valid search spaces even in the \textit{Hard} setting.
For example, for Task 5 (HPO of SVM), all LLMs include the key variables \texttt{cost}, \texttt{kernel}, and \texttt{gamma} with appropriate ranges (see \cref{tab:search_space_svm_app} for details).
Conversely, for tasks in Olympus and MECHBench, models frequently design invalid or ineffective search spaces in the \textit{Hard} setting. 
For example, Task 17 in MECHBench defines \texttt{thickness}, the uniform wall thickness over the canonical domain $[0.7, 3.0]$, while its \textit{Hard} description states only the loading condition and gives neither the variable names nor their scales.
All four LLMs introduce a wall-thickness variable, but specify substantially different search ranges: $[1.0, 3.0]$ (Gemini 3 Flash), $[0.8, 4.0]$ (GPT-5.2), $[1.0, 5.0]$ (Gemini 3.1 Pro), and $[0.0005, 0.02]$ (GPT-5 mini), which is specified on a markedly different scale from the other ranges.
\cref{app:searchspace} provides the details of the search spaces designed by LLMs.
Given that the worst value is returned when a solution lies outside the canonical domain, the score of the first three models degrades as the feasible fraction of their intervals decreases ($100\%$, $82\%$, and $50\%$).
In contrast, the interval designed by GPT-5 mini contains no feasible values, resulting in the worst score ($\infty$).
We note that several search spaces designed by LLMs from \textit{Hard} and \textit{Intermediate} descriptions contain design variables that lack a mapping to their canonical executable variable names. Our framework incorporates alias dictionaries that map each design variable in an LLM-generated search space to its canonical counterpart by covering as broad a range of name variations as possible. 
Developing an effective method to map variables more precisely is left for future work.

\paragraph{Budget Awareness in Algorithm Selection.}

A desired capability in BBOWP is the ability to select an algorithm suited to the computational budget.
We analyze the distribution of optimization algorithms selected by the LLMs with \textit{Small} ($N_{\max} = 100$) versus \textit{Large} ($N_{\max} = 10{,}000$) evaluation budgets.
The detailed figure of comparison can be found in \cref{app:algorithm_selection}.
We observe a highly rational shift in model behavior against the evaluation budget, confirming that all four LLMs possess a strong degree of budget awareness.
For example, the proportion of selecting Bayesian optimization algorithms (GP, SMAC3, and TPE) by GPT-5.2 is 56.5\% for \textit{Small} budget, whereas it decreases to 30.4\% for \textit{Large} budget.
More clearly, Gemini 3.1 Pro selects these algorithms at a rate of 78.3\% for \textit{Small} budget, but only 30.4\% for \textit{Large} budget.
This behavior indicates that LLMs correctly recognize that with an abundant evaluation budget, population-based evolutionary algorithms are preferable for exploiting their scalability and convergence ability.

\subsection{Scale-Invariant Evaluation of Search Space and Selected Algorithm}
\label{sec:results_metric}

Evaluating meta-solver performance across vastly different application domains presents a significant aggregation challenge, as raw objective values have incommensurable scales and bounds.
To evaluate the quality of the meta-solvers' formulations composed of the search space $S$ and the selected algorithm $a$ in a scale-invariant manner, we propose a quantile-based evaluation metric.

\paragraph{Formulation Ranking Score (FRS).}

We measure the probability that the best objective function value found based on the meta-solver's answer is superior to the results obtained from search spaces and optimization algorithms obeying a certain distribution.
Formally, let $y(a, S)$ denote the best objective function value achieved by running an optimization algorithm $a \in \mathcal{A}$ on a search space $S \in \mathcal{S}$ for a given task.
Assuming a minimization problem without loss of generality, we define \textit{Formulation Ranking Score (FRS)} for a specific configuration $(a, S)$ as follows:
\begin{align}
    \text{FRS} (a, S) = \Pr_{(a', S') \sim P_{(a, S)}} \left( y(a, S) \leq y(a', S') \right)
    \enspace ,
    \label{eq:def_metric}
\end{align}
where $P_{(a, S)}$ indicates the reference distribution of algorithms and search spaces.
This metric quantifies the quality of the configuration $(a, S)$ relative to those distributed according to $P_{(a, S)}$.

In practice, since this exact probability cannot be computed analytically, we employ a Monte Carlo approximation to \cref{eq:def_metric} to estimate it:
\begin{align}
    \widehat{\text{FRS}}(a, S) = \frac{1}{|\mathcal{R}|} \sum_{(a', S') \in \mathcal{R}} \mathbb{I} \left\{ y(a, S) \leq y(a', S') \right\}
    \enspace ,
\end{align}
where $\mathbb{I}(\cdot)$ is the indicator function, and $\mathcal{R}$ represents a finite reference set of optimization algorithms and search spaces drawn from $P_{(a, S)}$.
In our experiments, to construct the reference set $\mathcal{R}$, we sampled the search space $S'$ by sampling variable bounds and categorical choices from the canonical design space according to a uniform distribution, while the optimization algorithm $a'$ was randomly sampled from the candidate algorithms defined in \cref{sec:exp_setup}, excluding high-overhead solvers: SMAC3 and Bayesian optimization with Gaussian process.
The number of samples is $|\mathcal{R}| = 2{,}000$.
The approximated FRS indicates the proportion of elements in the reference set $\mathcal{R}$ that perform no better than the given configuration $(a, S)$. As the range of FRS is $[0, 1]$, we can measure the performance of meta-solvers on tasks with different domains on the same scale, regardless of the scales or landscape complexities of the underlying objective functions.
For simplicity, we hereafter denote $\widehat{\text{FRS}}$ as FRS.

\paragraph{FRS-based Comparison.}

\begin{table}[t]
\centering
\small
\caption{Comparison of FRS between different models under the \textit{Small} budget. For each task, the FRS is averaged over three optimization runs. The reported values are the mean across 22 tasks, excluding Task 17, and standard deviation across three optimization seeds. The mean FRS obtained using the human reference is \textbf{0.927}. Full results are provided in \protect\cref{app:results_frs}. \textbf{Bold} indicates the best score among the four models.}
\label{tab:results_metric_small}
\begin{tabular}{ccccc}
\toprule
Difficulty & Gemini 3 Flash & Gemini 3.1 Pro & GPT-5 mini & GPT-5.2 \\
\midrule
\textit{Easy} & \textbf{0.940} {\scriptsize $\pm$ 0.003} & 0.871 {\scriptsize $\pm$ 0.038} & 0.849 {\scriptsize $\pm$ 0.011} & 0.903 {\scriptsize $\pm$ 0.018} \\
\textit{Intermediate} & 0.420 {\scriptsize $\pm$ 0.019} & \textbf{0.446} {\scriptsize $\pm$ 0.024} & 0.354 {\scriptsize $\pm$ 0.004} & 0.439 {\scriptsize $\pm$ 0.002} \\
\textit{Hard} & \textbf{0.491} {\scriptsize $\pm$ 0.010} & 0.443 {\scriptsize $\pm$ 0.018} & 0.426 {\scriptsize $\pm$ 0.013} & 0.446 {\scriptsize $\pm$ 0.008} \\
\midrule
Average & \textbf{0.617} {\scriptsize $\pm$ 0.004} & 0.587 {\scriptsize $\pm$ 0.006} & 0.543 {\scriptsize $\pm$ 0.002} & 0.596 {\scriptsize $\pm$ 0.008} \\
\bottomrule
\end{tabular}
\end{table}

Using the FRS metric, we evaluate the performance of the LLMs across the 22 tasks, excluding Task 17 due to the long computation time required to construct its reference set, under the \textit{Small} budget ($N_{\textrm{max}}=100$).
\cref{tab:results_metric_small} summarizes the results.
\mbox{Gemini 3 Flash} achieves the highest overall average FRS (0.617), followed by \mbox{GPT-5.2} (0.596).

Analyzing performance across difficulty levels reveals a sharp contrast.
In the \textit{Easy} setting, all LLMs achieve high FRS, and Gemini 3 Flash even exceeds the human reference.
For instance, in Task 19, Gemini 3 Flash restricts the gains for the pole angle and angular velocity to non-negative ranges based on PD-control principles, and achieves the best objective value among the LLMs and the human reference (see \cref{tab:search_space_pendulum_easy_app} for details).

In contrast, performance drops markedly in the \textit{Intermediate} and \textit{Hard} settings.
This drop can be mainly attributed to a mismatch between the generated search spaces and the task contexts.
For example, in Task 19 under the \textit{Hard} setting, GPT-5.2 names each gain after the corresponding observation (e.g., \texttt{w\_angle}) and attains the best objective value.
However, Gemini 3.1 Pro names the gains only by index (\texttt{w\_0}, \dots, \texttt{w\_3}) alongside a separate \texttt{bias}, making their correspondence to the actual variables ambiguous and resulting in no valid evaluations (see \cref{tab:search_space_pendulum_hard_app}).

\subsection{Isolating Search Space Quality}
\label{sec:results_search_space}

\begin{figure}[t]
  \begin{minipage}{0.35\textwidth}
    \centering
    \includegraphics[width=0.99\linewidth]{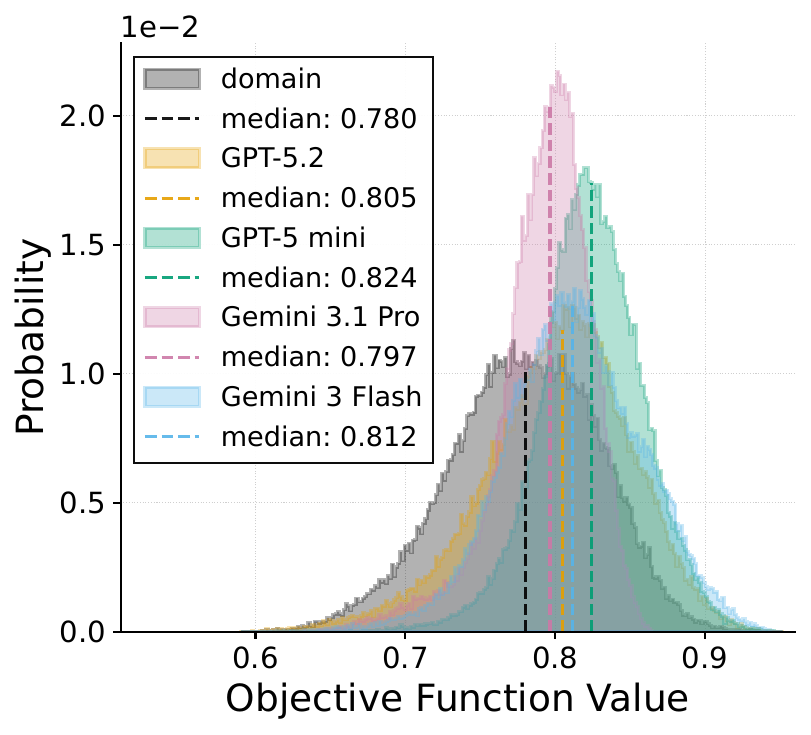}
  \end{minipage}%
  \begin{minipage}{0.65\textwidth}
    \centering
    \footnotesize
    \begin{tabularx}{\linewidth}{
      >{\raggedright\arraybackslash}X 
      >{\raggedright\arraybackslash}X 
      >{\raggedright\arraybackslash}X 
      >{\raggedright\arraybackslash}X}
      \toprule
      Variable & Domain & GPT-5 mini & Gemini 3.1 Pro \\
      \midrule
      sample.fraction   & [0.1, 1.0] & [0.1, 1.0] & [0.1, 1.0] \\
      num\_trees & [1, 2000] & [30, 500] & \underline{500} \\
      mtry.power & [0.0, 1.0] & [0.0, 1.0] & [0.0, 1.0] \\
      num.random splits & [1, 100] & [1, 50] & [1, 100] \\
      min.node.size & [1, 100] & [1, 100] & [1, 100] \\
      splitrule & \{gini, \newline extratrees\} & \{gini, \newline extratrees\} & \{gini, \newline extratrees\} \\
      respect.unordered factors & \{ignore, order, partition\} & \{ignore, order, partition\} & \{ignore, order\} \\
      imputation & \{impute.mean, impute.median, impute.hist\} & \{impute.mean, impute.median, impute.hist\} & \{impute.mean, impute.median\} \\
      \bottomrule
    \end{tabularx}
  \end{minipage}
  \caption{Search space comparison designed by LLMs for Task 3 in \textit{Easy} difficulty. \textbf{(Left)} Empirical probability density functions (EPDFs) of objective function value over the canonical domain (gray) versus the search spaces designed by four LLMs. \textbf{(Right)} Comparison table of ranges or categorical choices for each variable in the canonical domain and the search spaces designed by GPT-5 mini (achieved the highest median) and Gemini 3.1 Pro (exhibited the worst median). \underline{Underline} in the table indicates that the variable is set to a constant value. Gemini~3.1 Pro aggressively reduced the search space.}
  \label{fig:search_space_dist_yahpo}
\end{figure}

While the FRS metric evaluates the quality of a pair consisting of a selected algorithm $a$ and a search space $S$, it does not disentangle whether high performance stems from a robust optimization algorithm or a well-designed search space.
In this section, we therefore isolate the quality of the generated search space $S$ independently of the optimization algorithm. 
We achieve this by analyzing the empirical probability density function (EPDF) of objective function values within the search space designed by the LLMs. 
For three selected tasks (Tasks 3, 14, 20), we uniformly sample 100,000 points from both the canonical domain and each LLM-designed search space, evaluating the objective function value for every point.

\paragraph{Search Space Reduction by LLMs.}

The left panel of \cref{fig:search_space_dist_yahpo} illustrates the EPDFs of the objective function values for Task 3, HPO of Ranger in YAHPO\@. 
The gray curve represents the canonical domain, which exhibits a wide variance and a median objective value of 0.780.
Because the canonical domain is broad, an optimization algorithm operating under a strict budget of $N_{\mathrm{max}}=100$ is highly likely to waste evaluations in low-performing regions. 

The EPDFs over the search spaces designed by LLMs reveal how the quality of the designed search spaces differs among LLMs.
GPT-5 mini (green curve) shifts the bulk of the probability mass rightward to a median of 0.824.
It successfully covers a desirable region with high-quality solutions.
Conversely, Gemini 3.1 Pro (red curve) yields a lower median of 0.797.
Its EPDF exhibits exceptionally high kurtosis, with a sharp and narrow peak strictly concentrated around its median, resulting in highly homogeneous objective function values.

The right panel of \cref{fig:search_space_dist_yahpo} compares the canonical design space with the search spaces designed by GPT-5 mini and Gemini 3.1 Pro.
GPT-5 mini reduced the search ranges of two variables, \texttt{num\_trees} and \texttt{num.random.splits}, whereas Gemini 3.1 Pro restricted the choices of two categorical variables, \texttt{respect.unordered.factors} and \texttt{imputation}.
Additionally, Gemini 3.1 Pro fixed \texttt{num\_trees} to a constant value, thereby reducing the dimensionality of the search space by one.
Gemini 3.1 Pro thus constructed a narrower search space, which may explain the smaller variance in its EPDF, while its lower median compared to GPT-5 mini suggests that reducing the search space does not necessarily yield a region containing a larger proportion of high-quality solutions.
See \cref{app:results_sample_space_quality} for more results.
A 2$\times$2 ablation that separately replaces the generated search space and the selected algorithm supports this observation with actual optimization runs (Appendix~\ref{app:ss_algo_ablation}).

\section{Conclusion}
\label{sec:conclusion}

This paper proposed the Black-Box Optimization Word Problem (BBOWP), a novel problem setting requiring LLMs to infer both a search space and an optimization algorithm from the natural-language description of the target BBO task, and established the BBOWP-Bench as the first dataset and evaluation framework for the BBOWP.
Each problem instance in BBOWP-Bench provides a natural-language problem description, an executable evaluation environment, and a human reference formulation, enabling the evaluation of both search-space design and algorithm selection.
Through the alias-based alignment of generated variables with a predefined canonical design space, BBOWP-Bench approximates the evaluation of open-ended formulation, in which meta-solvers can freely design search spaces.
Using BBOWP-Bench, we evaluated several LLMs and assessed their ability on BBOWP.
We also introduced FRS, a quantitative and intuitive evaluation metric, to evaluate search configurations, and analyzed the EPDF of objective function values to assess the generated search space.
The empirical evaluations demonstrate that current LLMs vary their choice of optimization algorithms according to the evaluation budget, whereas they sometimes struggle with search space design, particularly in identifying important variables and balancing their ranges.

\section{Limitations and Future Work}
\label{sec:limitation}

BBOWP-Bench is an initial benchmark with a small number of tasks and has several limitations that suggest future directions.
A major limitation lies in the alias-based mapping: due to the limited coverage of the alias dictionaries, some generated variables that should be mapped to canonical variables are left unmapped, and the corresponding canonical variables are fixed to their default values rather than optimized (see \cref{app:alias_mapping}).
This coverage issue may be mitigated by extending the alias dictionaries or by using an additional LLM to map the generated identifier to a pre-defined identifier more appropriately.
Beyond this coverage issue, there are structural limitations that cannot be addressed by the mapping alone: formulations based on different coordinate systems, units, or parameterizations, as well as variables without canonical counterparts, cannot be evaluated, since BBOWP-Bench assumes that a valid formulation lies within the predefined canonical design space.
Regarding the evaluation protocol, the optimization results and the FRS over 22 tasks are based on a single LLM generation per model and task.
The human reference is constructed from the \textit{Easy} descriptions and serves as a non-optimal reference rather than an upper bound.
Moreover, BBOWP-Bench does not support providing conditional search space information to optimizers, which often appears in HPO tasks, indicating the need to improve our benchmark.
In addition, expanding the fixed set of candidate algorithms is a possible future direction, and generating new algorithms with an agent-based meta-solver is an alternative approach, which needs to be evaluated separately from the formulation ability.
Within our framework, an agent could instead iteratively revise its formulation based on observed optimization outcomes without modifying the objective, under a total budget fixed to $N_{\max}$.

\begin{ack}
This study was partially funded by JST SPRING (JPMJSP2178), JST BOOST (JPMJBS2427), JSPS KAKENHI (JP23K28156, JP23H00491, JP24K20857, JP26K02993), and JST Grant Number JPMJPF2203.
\end{ack}

\bibliographystyle{plainnat}
\bibliography{reference}

\newpage

\appendix

\section{Details of the instances in BBOWP-Bench}
\cref{tab:all_instance} presents the list of all instances in BBOWP-Bench.

\begin{table}[htbp]
  \centering
  \caption{All instances in BBOWP-Bench. \textbf{Dim} indicates the number of canonical design variables. \textbf{Float}, \textbf{Int}, and \textbf{Cat} indicate whether floating (continuous), integer, and categorical variables are included, respectively.}
  {\small
    \setlength{\tabcolsep}{2.5pt}
    \renewcommand{\arraystretch}{0.95}
    \begin{tabular}{l c >{\raggedright\arraybackslash}p{0.5\columnwidth} c c c c}
      \toprule
      \textbf{Source} & \textbf{Task} & \textbf{Scenario} & \textbf{Dim} & \textbf{Float} & \textbf{Int} & \textbf{Cat} \\
      \midrule
      YAHPO & 1 & HPO of MLP & 7 & $\checkmark$ & $\checkmark$ & \\
      & 2 & HPO of RcppHNSW & 6 &  & $\checkmark$ & $\checkmark$ \\
      & 3 & HPO of Ranger & 8 & $\checkmark$ & $\checkmark$ & $\checkmark$ \\
      & 4 & HPO of RPART & 5 & $\checkmark$ & $\checkmark$ & $\checkmark$ \\
      & 5 & HPO of SVM & 6 & $\checkmark$ & $\checkmark$ & $\checkmark$ \\
      & 6 & HPO of XGBoost & 14 & $\checkmark$ & $\checkmark$ & $\checkmark$ \\
      & 7 & Algorithm selection, HPO & 36 & $\checkmark$ & $\checkmark$ & $\checkmark$ \\
      \midrule
      Olympus & 8 & N-benzylation & 4 & $\checkmark$ &  & \\
      & 9 & Color mixing BOB & 5 & $\checkmark$ &  & \\
      & 10 & Color mixing N9 & 3 & $\checkmark$ &  & \\
      & 11 & Fullerenes synthesis & 3 & $\checkmark$ &  & \\
      & 12 & Photo PCE10 blend & 4 & $\checkmark$ &  & \\
      & 13 & Photo WF3 blend & 4 & $\checkmark$ &  & \\
      & 14 & S${}_N$Ar reaction & 4 & $\checkmark$ &  & \\
      & 15 & HPLC system & 6 & $\checkmark$ &  & \\
      & 16 & Suzuki reaction & 4 & $\checkmark$ &  & \\
      \midrule
      MECHBench & 17 & Star-shaped crash box & 5 & $\checkmark$ &  & \\
      & 18 & Three-point bending of layered beam & 10 & $\checkmark$ &  & \\
      \midrule
      MuJoCo & 19 & Inverted Pendulum & 5 & $\checkmark$ &  & \\
      & 20 & Inverted Double Pendulum & 10 & $\checkmark$ &  & \\
      & 21 & Hopper & 36 & $\checkmark$ &  & \\
      & 22 & Reacher & 22 & $\checkmark$ &  & \\
      & 23 & Swimmer & 18 & $\checkmark$ &  & \\
      \bottomrule
    \end{tabular}
  }
  \label{tab:all_instance}
\end{table}

\section{Code and Data Availability} \label{app:code_dataset}
BBOWP-Bench includes four existing assets: YAHPO Gym~\citep{yahpo}, Olympus~\citep{olympus_2021, olympus_2023}, MECHBench~\citep{mechbench}, and MuJoCo~\citep{todorov2012mujoco}.
YAHPO Gym and MuJoCo are licensed under the Apache-2.0 License.
Olympus and MECHBench are licensed under the MIT License.
Our code is available under the MIT License at \url{https://github.com/shiralab/bbowp-bench}.
Our benchmark and dataset are also released under the Creative Commons Attribution 4.0 International License (CC BY 4.0) at \url{https://github.com/shiralab/bbowp-bench/tree/main/benchmark}.

\section{Further Discussion on Benchmark Design} \label{app:design}

In building BBOWP-Bench, we made several decisions about the benchmark design.
Here, we discuss other possibilities.

\paragraph{Evaluation Using Prebuilt Functions.}

Instead of using prebuilt functions, we had another option: building an objective function according to a given search space on the fly, using coding agents.
This would make it possible to evaluate broader search spaces than those we considered.
We did not adopt this approach due to two concerns.
First, automatically generating objective functions is itself a challenging task and has been studied as a research topic in \citep{li2025solver}.
This could confound the evaluation results, as it would entangle the capabilities of meta-solvers with those of coding agents.
Second, coding agents are large, expensive, and frequently updated, which detracts from the portability and stability of the benchmark.
We therefore prefer a more compact and stable design.
For the same reason, we did not adopt LLM-as-a-Judge.

\paragraph{Identifier Matching Using Alias Lists.}

An obvious limitation of BBOWP-Bench is that valid design variables may be overlooked due to the limited coverage of the alias lists (see \cref{app:alias_mapping} for details).
Using vector representations of identifiers, such as Word2Vec \citep{mikolov2013distributed}, could alleviate this limitation.
However, doing so would introduce a dependency on another model, which is generally difficult to control.
As a first step in the BBOWP research, we chose a more controllable approach.
This simple procedure allows us to easily control the precision--recall trade-off in identifier matching, while leaving room for improvement through additional alias lists.
Combining these approaches could be a promising direction for further development.

\section{Details for Input and Output of BBOWP-Bench}
\label{app:input_output}

\subsection{Prompt Template for LLMs}
\label{app:prompt}
\begin{figure}[htbp]
    \begin{tcolorbox}[
        title=Prompt Template,
        colback=gray!5!white,
        colframe=gray!75!black,
        fonttitle=\bfseries,
        boxrule=0.5pt,
        left=10pt, right=10pt,
        top=10pt, bottom=10pt,
        sharp corners,
    ]
    \footnotesize
    \textbf{System Prompt:}\\
    {\raggedright
    \# Task
    
    You are an expert in black-box optimization.
    The user will give you a task description of a black-box optimization problem.
    Your goal is to generate a JSON content that defines all of the following items based on the description:
    
    \begin{enumerate}[leftmargin=20pt]
        \item search space
        \item the best choice of an optimization algorithm, which must be selected from the predefined algorithms in the schema
    \end{enumerate}
    
    \# Policy
    
    To achieve the best result within the limited budget, please follow the policy below when generating the JSON content.
    \begin{itemize}[leftmargin=20pt]
        \item For efficient optimization under the given budget, you can flexibly control the size of the search space by
        \begin{enumerate}[leftmargin=20pt]
            \item reducing the dimensionality of the search space (e.g., by fixing some parameters to be constant).
            \item adjusting the range of numerical parameters.
            \item reducing the number of choices for categorical parameters.
        \end{enumerate}
        \item You should choose the optimization algorithm that is most suitable for the defined search space and the given budget.
    \end{itemize}
    
    \# Output Format
    
    Your output must be a JSON content that strictly follows the schema defined below.
    
    \{SCHEMA\}\\
    
    Additionally, the search space in the JSON content must be defined according to the following schema, where you can choose any combination of the variable types (int, float, bool, cat) and the constraints mentioned below:
    \begin{itemize}[leftmargin=20pt]
        \item int: \{"actual\_name": "...", "type": "int", "low": 1, "high": 10, "step": 1, "log": false\}
        \item float: \{"actual\_name": "...", "type": "float", "low": 1e-4, "high": 1.0, "log": true\}
        \item bool: \{"actual\_name": "...", "type": "bool"\}
        \item cat: \{"actual\_name": "...", "type": "cat", "choices": [\{"actual\_name": "a"\}, \{"actual\_name": "b"\}]\}
        \item const\_int: \{"actual\_name": "...", "type": "const\_int", "value": 5\}
        \item const\_float: \{"actual\_name": "...", "type": "const\_float", "value": 0.1\}
        \item const\_bool: \{"actual\_name": "...", "type": "const\_bool", "value": true\}
        \item const\_cat: \{"actual\_name": "...", "type": "const\_cat", "value": "gbtree"\}
    \end{itemize}
    
    Make sure to include reasons for your search space design and algorithm choice in the JSON content, each of which should be written in the following aspects:
    \begin{itemize}[leftmargin=20pt]
        \item search space design
        \begin{itemize}[leftmargin=20pt]
          \item the reason why each variable is included in the search space
          \item the reason why each variable is optimized/fixed
          \item the reason for the range/choices of each variable
        \end{itemize}
        \item algorithm choice
        \begin{itemize}[leftmargin=20pt]
            \item the reason why the chosen algorithm is suitable for the defined search space and the given budget
        \end{itemize}
    \end{itemize}
    
    \# Constraints
    \begin{enumerate}[leftmargin=20pt]
        \item The name of each variable in the search space must be in snake\_case.
        \item The name of each choice for categorical variables must be in snake\_case.
    \end{enumerate}
    \par}
    
    \vspace{1em}
    
    \textbf{User Prompt:}\\
    {\raggedright
    Based on the task description given below, generate a JSON content according to the defined schema and system instructions.

    \vspace{0.5em}
    
    \# Task Description
    
    \{TASK\_DESCRIPTION\}
    \par}
    \end{tcolorbox}
    \caption{Prompt template. \{SCHEMA\} defines the JSON schema for the output, specifying the available optimization algorithms, the required properties for each variable type, and the reasons behind the search space design and algorithm selection.}
    \label{box:prompt}
\end{figure}

\begin{figure}[htbp]
    \begin{tcolorbox}[
        title=Prompt Template,
        colback=gray!5!white,
        colframe=gray!75!black,
        fonttitle=\bfseries,
        boxrule=0.5pt,
        left=10pt, right=10pt,
        top=10pt, bottom=10pt,
        sharp corners,
    ]
    \textbf{System Prompt:}\\
    {\raggedright
    \# Task

    ...

    \vspace{0.5em}

    \# Policy

    ...

    \vspace{0.5em}

    \# Output Format

    ...

    \vspace{0.5em}

    \# Constraints

    ...

    \vspace{0.5em}

    \textbf{\# Example}

    \vspace{0.3em}
    
    \#\# Task Description
    
    The objective is to maximize the number of timesteps an inverted pendulum remains balanced, simulated in MuJoCo via the Gymnasium interface (InvertedPendulum-v5). The control is performed by a linear policy that computes a horizontal force applied to the cart based on the current observation at each timestep. The episode terminates early if the absolute value of the pole angle exceeds 0.2 radians, and the maximum episode length is 1000 timesteps. The observation variables are: position (position of the cart along the linear track, in meters), angle (vertical angle of the pole relative to the upright position, in radians), velocity (linear velocity of the cart, in meters per second), and angular\_velocity (angular velocity of the pole, in radians per second). The action variable force (horizontal force applied to the cart, in Newtons) is clipped to [-3.0, 3.0] by the simulator. The design variables consist of 5 continuous variables. Four of them are feedback gains that multiply each observation variable, and the remaining one is a bias term. The feedback gains are named as 'observation variable'\_to\_'action variable', and the bias term is named as bias\_to\_force. Each design variable ranges over [-1000, 1000]. Please note that while these specified ranges define the absolute implementational boundaries for each parameter, they serve primarily as a reference frame; you can flexibly control the actual search space depending on your optimization strategy. The evaluation budget is 100.

    \vspace{0.5em}
    
    \#\# Output
    
    \{
        "search\_space": \{
            "variables": [
                \{
                    "actual\_name": "position\_to\_force",
                    "value": 0.0,
                    "type": "const\_float"
                \},
                \{
                    "actual\_name": "angle\_to\_force",
                    "low": -150.0,
                    "high": 150.0,
                    "log": false,
                    "type": "float"
                \},
                \{
                    "actual\_name": "velocity\_to\_force",
                    "low": -6.0,
                    "high": 6.0,
                    "log": false,
                    "type": "float"
                \},
                \{
                    "actual\_name": "angular\_velocity\_to\_force",
                    "low": -30.0,
                    "high": 30.0,
                    "log": false,
                    "type": "float"
                \},
                \{
                    "actual\_name": "bias\_to\_force",
                    "value": 0.0,
                    "type": "const\_float"
                \}
            ],
            "reason": "The search space is designed based on the scale of the action variable, which is bounded to [-3.0, 3.0] by the simulator, and has a margin to allow broad exploration. The reduction of the search space is needed due to the limited evaluation budget. The gain of angle, velocity, and angular velocity can be considered the three most important parameters for the control of the inverted pendulum, and the gain of position is fixed to 0.0 to reduce the search space. The bias term is fixed to 0.0 as this is expected to be less important."
        \},
        "algorithm": \{
            "name": "TPE",
            "reason": "The search space involves three continuous variables. TPE is suitable for finding a reasonable solution within a limited budget and can handle continuous optimization problems."
        \}
    \}
    
    \par}
    \end{tcolorbox}
    \caption{System prompt for one-shot settings. The same parts as the prompt of zero-shot settings are omitted.}
    \label{box:one_shot}
\end{figure}

We use the prompt template provided in \cref{box:prompt} as the input of BBOWP-Bench.
For one-shot settings, we use the system prompt with the example based on Task 19, provided in \cref{box:one_shot}.

\subsection{Task Description}

\begin{figure}[htbp]
    \begin{tcolorbox}[
        title=Task Descriptions of Task 3 (HPO of Ranger):,
        colback=gray!5!white,
        colframe=gray!75!black,
        fonttitle=\bfseries,
        boxrule=0.5pt,
        left=10pt, right=10pt,
        top=10pt, bottom=10pt,
        sharp corners,
    ]
    \textbf{\textit{Easy} difficulty:}\\
    {\raggedright
    We need to optimize the hyperparameters of a random forest (Ranger) model to maximize AUC for a classification task. This optimization is formulated within an 8-dimensional mixed search space comprising continuous, integer, and categorical variables. It is important to consider the mechanics of ensemble learning and decision tree construction. Achieving robust predictive performance heavily relies on the discrete size of the tree ensemble, explicitly controlled by the variable 'num.trees' (typically evaluated within the bounds of 1 to 2000), alongside the continuous proportion of the training data sampled to build each individual tree, represented by the variable 'sample.fraction' (ranging from 0.1 to 1.0). Furthermore, the model's ability to capture complex patterns while maintaining diversity is governed by several factors: the strategy used to determine the subset of features evaluated at each split, defined by the variable 'mtry.power' (ranging from 0.0 to 1.0), the categorical rule chosen to measure split quality, corresponding to the variable 'splitrule' (with options being 'gini' or 'extratrees'), and the discrete number of random threshold evaluations per feature, controlled by the variable 'num.random.splits' (ranging from 1 to 100). To prevent the individual trees from growing too deep and overfitting, you must also carefully tune the discrete stopping criterion that defines the minimum allowed number of samples in a terminal leaf, which is managed by the variable 'min.node.size' (ranging from 1 to 100). Finally, the categorical approach taken to handle non-ordinal features, determined by the variable 'respect.unordered.factors' (accepting values like 'ignore', 'order', or 'partition'), combined with the preliminary strategy chosen to resolve missing data points, specified by the variable 'num.impute.selected.cpo' (with options 'impute.mean', 'impute.median', or 'impute.hist'), plays a critical role in the stability and ultimate performance of the resulting model. Please note that while these specified ranges and categorical options define the absolute implementational boundaries and available choices for each parameter, they serve primarily as a reference frame; you can flexibly control the actual search space depending on your optimization strategy. \\
    The evaluation budget is \{MAX\_EVAL\_VALUE\}.
    \par}

    \vspace{1.5em}

    \textbf{\textit{Intermediate} difficulty:}\\
    {\raggedright
    We need to optimize the hyperparameters of a random forest (Ranger) model to maximize AUC for a classification task. It is important to consider the mechanics of ensemble learning and decision tree construction. Achieving robust predictive performance heavily relies on the discrete size of the tree ensemble, alongside the continuous proportion of the training data sampled to build each individual tree. Furthermore, the model's ability to capture complex patterns while maintaining diversity is governed by several factors: the strategy used to determine the subset of features evaluated at each split, the categorical rule chosen to measure split quality, and the discrete number of random threshold evaluations per feature. To prevent the individual trees from growing too deep and overfitting, you must also carefully tune the discrete stopping criterion that defines the minimum allowed number of samples in a terminal leaf. Finally, the categorical approach taken to handle non-ordinal features, combined with the preliminary strategy chosen to resolve missing data points, plays a critical role in the stability and ultimate performance of the resulting model. \\
    The evaluation budget is \{MAX\_EVAL\_VALUE\}.
    \par}

    \vspace{1.5em}

    \textbf{\textit{Hard} difficulty:}\\
    {\raggedright
    We need to optimize the hyperparameters of Ranger (random forest) to maximize AUC. \\
    The evaluation budget is \{MAX\_EVAL\_VALUE\}.
    \par}
    \end{tcolorbox}
    \caption{Task descriptions of Task 3 (HPO of Ranger in YAHPO) for the different difficulties, \textit{Easy}, \textit{Intermediate}, and \textit{Hard}.}
    \label{box:task_desc}
\end{figure}

As described in \cref{sec:data_description}, we prepare three descriptions for each task with different difficulty levels (Easy, Intermediate, and Hard). 
We illustrate an example of the task descriptions of Task 3 (HPO of Ranger in YAHPO Gym) for the three difficulties in \cref{box:task_desc}.

\subsection{Output Example}
\begin{figure}[htbp]
    \begin{tcolorbox}[
        colback=gray!5!white,
        colframe=gray!75!black,
        boxrule=0.5pt,
        left=10pt, right=10pt,
        top=10pt, bottom=10pt,
        sharp corners,
    ]
    \small
    {\raggedright
    \{\\
    \hspace{1em} "search\_space": \{\\
    \hspace{2em} "variables": [\\
    \hspace{3em} \{"actual\_name": "cp", "low": 0.0009118819655545162, "high": 1.0, \\
    \hspace{3em} "log": true, "type": "float"\},\\
    \hspace{3em} \{"actual\_name": "maxdepth", "low": 1, "high": 30, "step": 1, \\
    \hspace{3em} "log": false, "type": "int"\},\\
    \hspace{3em} \{"actual\_name": "minsplit", "low": 1, "high": 100, "step": 1, \\
    \hspace{3em} "log": false, "type": "int"\},\\
    \hspace{3em} \{"actual\_name": "minbucket", "low": 1, "high": 100, "step": 1, \\
    \hspace{3em} "log": false, "type": "int"\},\\
    \hspace{3em} \{"actual\_name": "num\_impute\_selected\_cpo",\\
    \hspace{3em} "choices": [\\
    \hspace{4em} \{"actual\_name": "impute\_mean"\}, \\
    \hspace{4em} \{"actual\_name": "impute\_median"\}, \\
    \hspace{4em} \{"actual\_name": "impute\_hist"\}\\
    \hspace{3em} ],\\
    \hspace{3em} "type": "cat"\\
    \hspace{3em} \}\\
    \hspace{2em} ],\\
    \hspace{2em} "reason": "The search space consists of 5 dimensions (one continuous, three integers, and one \\
    \hspace{2em} categorical variable). Given the generous evaluation budget of 10,000 trials, there is ample capacity \\
    \hspace{2em} to comprehensively explore the entire space without needing to freeze or narrow down any \\
    \hspace{2em} hyperparameters. Thus, all 5 variables are selected to be optimized across their originally \\
    \hspace{2em} prescribed ranges and choices, giving the optimization algorithm maximum flexibility to uncover \\
    \hspace{2em} complex interactions and effectively maximize the AUC. Variable names and categorical choices \\
    \hspace{2em} originally containing periods were converted to strictly follow snake\_case format constraints."\\
    \hspace{1em} \},\\
    \hspace{1em} "algorithm": \{\\
    \hspace{2em} "name": "TPE",\\
    \hspace{2em} "reason": "Tree-structured Parzen Estimator (TPE) natively and robustly handles mixed search \\
    \hspace{2em} spaces encompassing continuous, integer, and categorical design variables. For this \\
    \hspace{2em} 5-dimensional problem with a large evaluation budget of 10,000, TPE demonstrates superior \\
    \hspace{2em} scalability and computational efficiency compared to standard Gaussian Process-based \\
    \hspace{2em} techniques. It maintains an effective balance between exploration and exploitation across the \\
    \hspace{2em} thousands of trials, securing excellent convergence toward optimal hyperparameters while \\
    \hspace{2em} keeping algorithmic overhead manageable."\\
    \hspace{1em} \}\\
    \}\\
    \par}
    \end{tcolorbox}
    \caption{Output for Task 4 with \textit{Easy} difficulty and \textit{Large} budget. Generated by Gemini 3.1 Pro Preview.}
    \label{box:example_output}
\end{figure}

\cref{box:example_output} shows an example of an output for Task 4 with \textit{Easy} difficulty and \textit{Large} budget.

\section{Experimental Environment} \label{app:resource}
\paragraph{Compute Resources.}
We executed the experiments on multiple computer resources as follows:

\begin{itemize}
    \item Two computers running Ubuntu 22.04 with AMD EPYC7763 $\times$ 2 CPUs, 1024GB memory, and SSD 3.8TB.
    \item Two computers running Ubuntu 20.04 with AMD EPYC7763 $\times$ 1 CPU, 512GB memory, and SSD 3.8TB.
    \item Two computers running Ubuntu 22.04 with AMD EPYC7742 $\times$ 1 CPU, 512GB memory, and SSD 3.8TB.
    \item Three computers running Ubuntu 22.04 with Xeon E-2186G $\times$ 1 CPU, 64GB memory, and SSD 2TB.
\end{itemize}

\paragraph{Execution Time.}

For the \textit{Small} budget ($N_{\max} = 100$), a single optimization run requires less than a minute for all tasks except Tasks 17 and 18 in MECHBench.
For Tasks 17 and 18, some settings last for about 3 hours.
For the \textit{Large} budget ($N_{\max} = 10{,}000$), a single optimization run takes about 3 hours when the optimizer is not Bayesian optimization with Gaussian process (GP) or SMAC3.
Some settings require more than 48 hours when using GP or SMAC3 as the optimizer, and we terminated optimization runs for such settings.

\section{Results of Algorithm Selection}
\label{app:algorithm_selection}

\begin{figure}[htbp]
    \centering
    \includegraphics[width=0.7\linewidth]{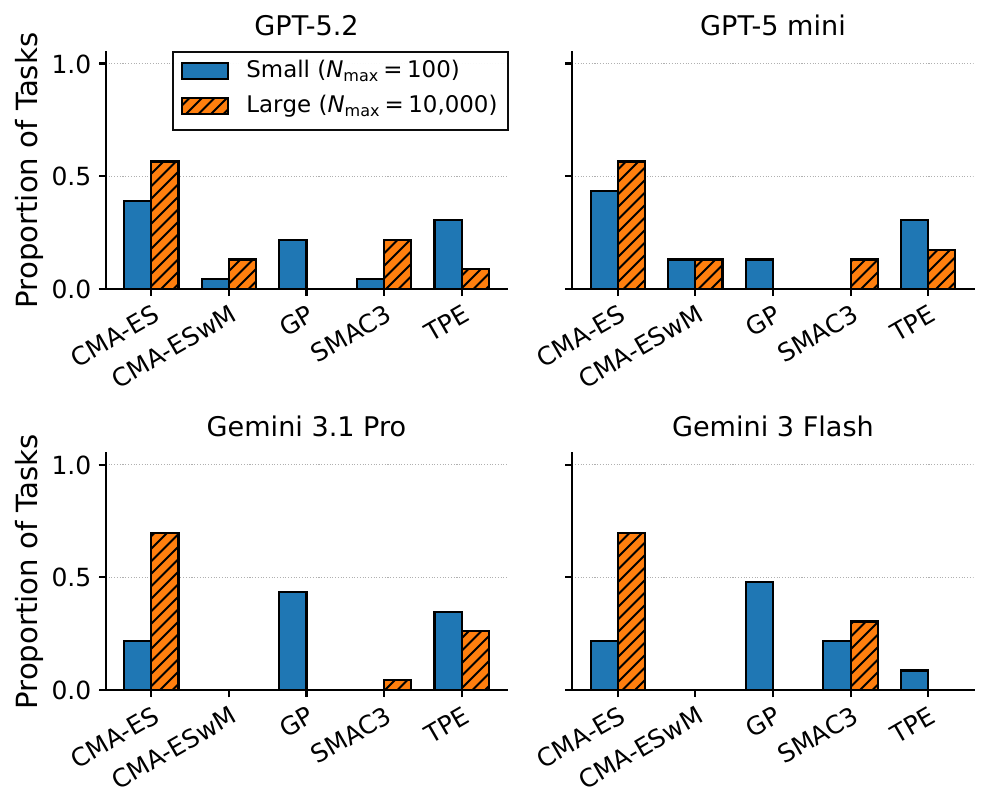}
    \caption{Distribution of optimization algorithms selected by the four evaluated LLMs across different evaluation budgets. Solid blue bars represent the \textit{Small} budget ($N_{\max} = 100$), while striped orange bars represent the \textit{Large} budget ($N_{\max} = 10{,}000$).}
    \label{fig:results_algorithm_selection_app}
\end{figure}

\cref{fig:results_algorithm_selection_app} shows the proportion of tasks for which each algorithm is selected by each LLM with the \textit{Small} and \textit{Large} budgets.
The results demonstrate strong budget awareness across all models: sample-efficient surrogate methods (GP, SMAC3, and TPE) are favored under \textit{Small} budget, whereas population-based evolution strategies (CMA-ES) dominate under \textit{Large} budget.

\section{Results of Optimization}
\label{app:experiment}

\subsection{Zero-Shot Settings}
\label{app:experiment:zeroshot}

We conducted the experiments for all 23 tasks in BBOWP-Bench for \textit{Small} budget ($N_{\max} = 100$).
The results are provided in \cref{tab:results_easy_small} for \textit{Easy} difficulty, \cref{tab:results_intermediate_small} for \textit{Intermediate} difficulty, and \cref{tab:results_hard_small} for \textit{Hard} difficulty.

Additionally, we conducted the experiments for 21 tasks for \textit{Large} budget ($N_{\max} = 10{,}000$).
Tasks 17 and 18 in MECHBench are excluded because each evaluation takes a minute or more in those tasks.
The results are provided in \cref{tab:results_easy_large} for \textit{Easy} difficulty, \cref{tab:results_intermediate_large} for \textit{Intermediate} difficulty, and \cref{tab:results_hard_large} for \textit{Hard} difficulty.
We terminated the optimization runs that took more than 48 hours, and omitted their results from each table, denoting such runs as ``--''.

Focusing on \textit{Small} budget, there is no significant difference across the model size (Gemini 3 Flash vs. Gemini 3.1 Pro and GPT-5 mini vs. GPT-5.2) for all difficulties.
We observe the same tendency for \textit{Large} budget.
In both budget settings, the best evaluation values on YAHPO tasks remain relatively stable, whereas they tend to decline as the settings become more challenging for Olympus, MECHBench, and MuJoCo.


\begin{table}[htbp]
\centering
\caption{Comparison of optimization performance for \textit{Easy} difficulty and \textit{Small} budget. Values are mean $\pm$ standard deviation over three runs.}
\label{tab:results_easy_small}
\setlength{\tabcolsep}{3.5pt}
\begin{tabular}{@{}llrrrr@{}}
\toprule
Domain & \# Task & Gemini 3 Flash & Gemini 3.1 Pro & GPT-5 mini & GPT-5.2 \\
\midrule
\multirow{7}{*}{YAHPO} & 1 ($\uparrow$) & \textbf{99.59} {\scriptsize $\pm$ 0.04} & 99.58 {\scriptsize $\pm$ 0.01} & 99.55 {\scriptsize $\pm$ 0.10} & 99.25 {\scriptsize $\pm$ 0.06} \\
 & 2 ($\uparrow$) & \textbf{0.96} {\scriptsize $\pm$ 0.01} & \textbf{0.96} {\scriptsize $\pm$ 0.01} & \textbf{0.96} {\scriptsize $\pm$ 0.01} & \textbf{0.96} {\scriptsize $\pm$ 0.01} \\
 & 3 ($\uparrow$) & 0.94 {\scriptsize $\pm$ 0.02} & 0.86 {\scriptsize $\pm$ 0.00} & 0.94 {\scriptsize $\pm$ 0.00} & \textbf{0.95} {\scriptsize $\pm$ 0.01} \\
 & 4 ($\uparrow$) & \textbf{0.92} {\scriptsize $\pm$ 0.00} & 0.91 {\scriptsize $\pm$ 0.00} & 0.91 {\scriptsize $\pm$ 0.00} & \textbf{0.92} {\scriptsize $\pm$ 0.00} \\
 & 5 ($\uparrow$) & \textbf{0.98} {\scriptsize $\pm$ 0.00} & 0.95 {\scriptsize $\pm$ 0.01} & 0.97 {\scriptsize $\pm$ 0.01} & 0.95 {\scriptsize $\pm$ 0.01} \\
 & 6 ($\uparrow$) & \textbf{1.00} {\scriptsize $\pm$ 0.00} & 0.99 {\scriptsize $\pm$ 0.00} & \textbf{1.00} {\scriptsize $\pm$ 0.00} & 0.99 {\scriptsize $\pm$ 0.00} \\
 & 7 ($\uparrow$) & \textbf{0.99} {\scriptsize $\pm$ 0.00} & 0.96 {\scriptsize $\pm$ 0.00} & \textbf{0.99} {\scriptsize $\pm$ 0.00} & \textbf{0.99} {\scriptsize $\pm$ 0.00} \\
\midrule
\multirow{9}{*}{Olympus} & 8 ($\downarrow$) & 2.54 {\scriptsize $\pm$ 0.00} & \textbf{2.48} {\scriptsize $\pm$ 0.09} & 2.56 {\scriptsize $\pm$ 0.04} & 2.53 {\scriptsize $\pm$ 0.01} \\
 & 9 ($\downarrow$) & \textbf{0.02} {\scriptsize $\pm$ 0.00} & 0.03 {\scriptsize $\pm$ 0.00} & 0.05 {\scriptsize $\pm$ 0.02} & 0.07 {\scriptsize $\pm$ 0.00} \\
 & 10 ($\downarrow$) & 0.05 {\scriptsize $\pm$ 0.02} & \textbf{0.03} {\scriptsize $\pm$ 0.02} & 0.07 {\scriptsize $\pm$ 0.01} & 0.07 {\scriptsize $\pm$ 0.00} \\
 & 11 ($\uparrow$) & 0.95 {\scriptsize $\pm$ 0.00} & \textbf{0.96} {\scriptsize $\pm$ 0.00} & 0.95 {\scriptsize $\pm$ 0.00} & \textbf{0.96} {\scriptsize $\pm$ 0.00} \\
 & 12 ($\downarrow$) & \textbf{0.00} {\scriptsize $\pm$ 0.00} & \textbf{0.00} {\scriptsize $\pm$ 0.00} & 0.05 {\scriptsize $\pm$ 0.04} & 0.01 {\scriptsize $\pm$ 0.02} \\
 & 13 ($\downarrow$) & \textbf{0.00} {\scriptsize $\pm$ 0.00} & \textbf{0.00} {\scriptsize $\pm$ 0.00} & \textbf{0.00} {\scriptsize $\pm$ 0.00} & \textbf{0.00} {\scriptsize $\pm$ 0.00} \\
 & 14 ($\downarrow$) & \textbf{0.19} {\scriptsize $\pm$ 0.00} & 0.20 {\scriptsize $\pm$ 0.01} & 0.32 {\scriptsize $\pm$ 0.02} & \textbf{0.19} {\scriptsize $\pm$ 0.01} \\
 & 15 ($\uparrow$) & \textbf{2521.76} {\scriptsize $\pm$ 155.43} & 2436.03 {\scriptsize $\pm$ 64.52} & 2282.10 {\scriptsize $\pm$ 32.82} & 2277.24 {\scriptsize $\pm$ 33.38} \\
 & 16 ($\uparrow$) & 99.87 {\scriptsize $\pm$ 0.12} & \textbf{100.67} {\scriptsize $\pm$ 0.90} & $-\infty$ & 98.31 {\scriptsize $\pm$ 0.12} \\
\midrule
\multirow{2}{*}{MECHBench} & 17 ($\downarrow$) & \textbf{-12044.69} {\scriptsize $\pm$ 548.96} & \textbf{-12044.69} {\scriptsize $\pm$ 548.96} & -10823.81 {\scriptsize $\pm$ 682.55} & -10804.95 {\scriptsize $\pm$ 709.63} \\
 & 18 ($\downarrow$) & 3.41 {\scriptsize $\pm$ 0.01} & \textbf{3.39} {\scriptsize $\pm$ 0.01} & 3.42 {\scriptsize $\pm$ 0.00} & \textbf{3.39} {\scriptsize $\pm$ 0.01} \\
\midrule
\multirow{5}{*}{MuJoCo} & 19 ($\uparrow$) & \textbf{1000.00} {\scriptsize $\pm$ 0.00} & 427.40 {\scriptsize $\pm$ 511.62} & 421.90 {\scriptsize $\pm$ 501.22} & 180.60 {\scriptsize $\pm$ 7.71} \\
 & 20 ($\uparrow$) & \textbf{139.14} {\scriptsize $\pm$ 14.10} & 135.38 {\scriptsize $\pm$ 15.30} & 81.53 {\scriptsize $\pm$ 6.60} & 81.80 {\scriptsize $\pm$ 2.74} \\
 & 21 ($\uparrow$) & 606.11 {\scriptsize $\pm$ 319.33} & 479.45 {\scriptsize $\pm$ 355.15} & \textbf{735.76} {\scriptsize $\pm$ 490.50} & 396.98 {\scriptsize $\pm$ 19.16} \\
 & 22 ($\uparrow$) & -55.14 {\scriptsize $\pm$ 1.51} & -59.20 {\scriptsize $\pm$ 1.77} & \textbf{-12.25} {\scriptsize $\pm$ 0.00} & -21.21 {\scriptsize $\pm$ 12.49} \\
 & 23 ($\uparrow$) & 277.23 {\scriptsize $\pm$ 15.74} & 173.60 {\scriptsize $\pm$ 116.48} & 277.23 {\scriptsize $\pm$ 15.74} & \textbf{325.10} {\scriptsize $\pm$ 2.19} \\
\bottomrule
\end{tabular}

\end{table}

\begin{table}[htbp]
\centering
\caption{Comparison of optimization performance for \textit{Intermediate} difficulty and \textit{Small} budget. Values are mean $\pm$ standard deviation over three runs.}
\label{tab:results_intermediate_small}
\begin{tabular}{llrrrr}
\toprule
Domain & \# Task & Gemini 3 Flash & Gemini 3.1 Pro & GPT-5 mini & GPT-5.2 \\
\midrule
\multirow{7}{*}{YAHPO} & 1 ($\uparrow$) & \textbf{99.42} {\scriptsize $\pm$ 0.09} & 97.34 {\scriptsize $\pm$ 0.10} & 97.80 {\scriptsize $\pm$ 0.96} & 99.07 {\scriptsize $\pm$ 0.19} \\
 & 2 ($\uparrow$) & 0.95 {\scriptsize $\pm$ 0.00} & \textbf{0.96} {\scriptsize $\pm$ 0.00} & 0.94 {\scriptsize $\pm$ 0.00} & 0.92 {\scriptsize $\pm$ 0.00} \\
 & 3 ($\uparrow$) & 0.86 {\scriptsize $\pm$ 0.03} & \textbf{0.88} {\scriptsize $\pm$ 0.00} & \textbf{0.88} {\scriptsize $\pm$ 0.00} & 0.80 {\scriptsize $\pm$ 0.01} \\
 & 4 ($\uparrow$) & 0.91 {\scriptsize $\pm$ 0.00} & \textbf{0.92} {\scriptsize $\pm$ 0.00} & 0.91 {\scriptsize $\pm$ 0.00} & 0.91 {\scriptsize $\pm$ 0.00} \\
 & 5 ($\uparrow$) & \textbf{0.95} {\scriptsize $\pm$ 0.00} & \textbf{0.95} {\scriptsize $\pm$ 0.00} & 0.94 {\scriptsize $\pm$ 0.00} & 0.94 {\scriptsize $\pm$ 0.00} \\
 & 6 ($\uparrow$) & \textbf{0.99} {\scriptsize $\pm$ 0.00} & \textbf{0.99} {\scriptsize $\pm$ 0.00} & \textbf{0.99} {\scriptsize $\pm$ 0.00} & \textbf{0.99} {\scriptsize $\pm$ 0.00} \\
 & 7 ($\uparrow$) & 0.91 {\scriptsize $\pm$ 0.00} & $-\infty$ & 0.91 {\scriptsize $\pm$ 0.00} & \textbf{0.96} {\scriptsize $\pm$ 0.00} \\
\midrule
\multirow{9}{*}{Olympus} & 8 ($\downarrow$) & $\infty$ & \textbf{3.36} {\scriptsize $\pm$ 0.06} & $\infty$ & $\infty$ \\
 & 9 ($\downarrow$) & 0.17 {\scriptsize $\pm$ 0.00} & 0.16 {\scriptsize $\pm$ 0.01} & 0.17 {\scriptsize $\pm$ 0.00} & \textbf{0.05} {\scriptsize $\pm$ 0.00} \\
 & 10 ($\downarrow$) & \textbf{0.31} {\scriptsize $\pm$ 0.01} & \textbf{0.31} {\scriptsize $\pm$ 0.00} & \textbf{0.31} {\scriptsize $\pm$ 0.01} & \textbf{0.31} {\scriptsize $\pm$ 0.00} \\
 & 11 ($\uparrow$) & 0.85 {\scriptsize $\pm$ 0.05} & $-\infty$ & $-\infty$ & \textbf{0.92} {\scriptsize $\pm$ 0.00} \\
 & 12 ($\downarrow$) & \textbf{0.33} {\scriptsize $\pm$ 0.00} & \textbf{0.33} {\scriptsize $\pm$ 0.00} & \textbf{0.33} {\scriptsize $\pm$ 0.01} & \textbf{0.33} {\scriptsize $\pm$ 0.01} \\
 & 13 ($\downarrow$) & 0.36 {\scriptsize $\pm$ 0.00} & 0.36 {\scriptsize $\pm$ 0.00} & \textbf{0.35} {\scriptsize $\pm$ 0.00} & \textbf{0.35} {\scriptsize $\pm$ 0.00} \\
 & 14 ($\downarrow$) & $\infty$ & $\infty$ & \textbf{0.78} {\scriptsize $\pm$ 0.04} & 0.94 {\scriptsize $\pm$ 0.01} \\
 & 15 ($\uparrow$) & $-\infty$ & $-\infty$ & $-\infty$ & \textbf{1288.30} {\scriptsize $\pm$ 9.40} \\
 & 16 ($\uparrow$) & $-\infty$ & $-\infty$ & $-\infty$ & \textbf{73.25} {\scriptsize $\pm$ 3.33} \\
\midrule
\multirow{2}{*}{MECHBench} & 17 ($\downarrow$) & -8392.55 {\scriptsize $\pm$ 167.46} & \textbf{-8518.21} {\scriptsize $\pm$ 18.20} & 800.31 {\scriptsize $\pm$ 0.00} & 800.31 {\scriptsize $\pm$ 0.00} \\
 & 18 ($\downarrow$) & \textbf{3.61} {\scriptsize $\pm$ 0.00} & \textbf{3.61} {\scriptsize $\pm$ 0.00} & \textbf{3.61} {\scriptsize $\pm$ 0.00} & \textbf{3.61} {\scriptsize $\pm$ 0.00} \\
\midrule
\multirow{5}{*}{MuJoCo} & 19 ($\uparrow$) & 315.37 {\scriptsize $\pm$ 45.43} & \textbf{1000.00} {\scriptsize $\pm$ 0.00} & \textbf{1000.00} {\scriptsize $\pm$ 0.00} & \textbf{1000.00} {\scriptsize $\pm$ 0.00} \\
 & 20 ($\uparrow$) & 102.95 {\scriptsize $\pm$ 11.99} & \textbf{103.19} {\scriptsize $\pm$ 2.78} & 88.26 {\scriptsize $\pm$ 5.26} & 76.34 {\scriptsize $\pm$ 6.72} \\
 & 21 ($\uparrow$) & 138.52 {\scriptsize $\pm$ 0.00} & \textbf{1008.22} {\scriptsize $\pm$ 13.75} & 138.52 {\scriptsize $\pm$ 0.00} & 138.52 {\scriptsize $\pm$ 0.00} \\
 & 22 ($\uparrow$) & \textbf{-12.25} {\scriptsize $\pm$ 0.00} & \textbf{-12.25} {\scriptsize $\pm$ 0.00} & $-\infty$ & $-\infty$ \\
 & 23 ($\uparrow$) & \textbf{0.02} {\scriptsize $\pm$ 0.00} & \textbf{0.02} {\scriptsize $\pm$ 0.00} & \textbf{0.02} {\scriptsize $\pm$ 0.00} & \textbf{0.02} {\scriptsize $\pm$ 0.00} \\
\bottomrule
\end{tabular}
\end{table}

\begin{table}[htbp]
\centering
\caption{Comparison of optimization performance for \textit{Hard} difficulty and \textit{Small} budget. Values are mean $\pm$ standard deviation over three runs.}
\label{tab:results_hard_small}
\begin{tabular}{llrrrr}
\toprule
Domain & \# Task & Gemini 3 Flash & Gemini 3.1 Pro & GPT-5 mini & GPT-5.2 \\
\midrule
\multirow{7}{*}{YAHPO} & 1 ($\uparrow$) & 98.31 {\scriptsize $\pm$ 0.29} & \textbf{99.61} {\scriptsize $\pm$ 0.00} & $-\infty$ & 97.67 {\scriptsize $\pm$ 0.58} \\
 & 2 ($\uparrow$) & 0.93 {\scriptsize $\pm$ 0.00} & \textbf{0.94} {\scriptsize $\pm$ 0.00} & \textbf{0.94} {\scriptsize $\pm$ 0.00} & 0.93 {\scriptsize $\pm$ 0.00} \\
 & 3 ($\uparrow$) & \textbf{0.88} {\scriptsize $\pm$ 0.00} & 0.82 {\scriptsize $\pm$ 0.00} & 0.84 {\scriptsize $\pm$ 0.00} & 0.82 {\scriptsize $\pm$ 0.00} \\
 & 4 ($\uparrow$) & 0.91 {\scriptsize $\pm$ 0.00} & \textbf{0.92} {\scriptsize $\pm$ 0.00} & \textbf{0.92} {\scriptsize $\pm$ 0.00} & \textbf{0.92} {\scriptsize $\pm$ 0.00} \\
 & 5 ($\uparrow$) & \textbf{0.95} {\scriptsize $\pm$ 0.00} & \textbf{0.95} {\scriptsize $\pm$ 0.01} & 0.93 {\scriptsize $\pm$ 0.01} & \textbf{0.95} {\scriptsize $\pm$ 0.00} \\
 & 6 ($\uparrow$) & \textbf{0.99} {\scriptsize $\pm$ 0.00} & \textbf{0.99} {\scriptsize $\pm$ 0.00} & \textbf{0.99} {\scriptsize $\pm$ 0.00} & \textbf{0.99} {\scriptsize $\pm$ 0.00} \\
 & 7 ($\uparrow$) & 0.91 {\scriptsize $\pm$ 0.00} & \textbf{0.92} {\scriptsize $\pm$ 0.00} & 0.91 {\scriptsize $\pm$ 0.00} & $-\infty$ \\
\midrule
\multirow{9}{*}{Olympus} & 8 ($\downarrow$) & $\infty$ & $\infty$ & $\infty$ & $\infty$ \\
 & 9 ($\downarrow$) & \textbf{0.03} {\scriptsize $\pm$ 0.00} & \textbf{0.03} {\scriptsize $\pm$ 0.00} & 0.04 {\scriptsize $\pm$ 0.01} & 0.04 {\scriptsize $\pm$ 0.01} \\
 & 10 ($\downarrow$) & 0.31 {\scriptsize $\pm$ 0.01} & \textbf{0.04} {\scriptsize $\pm$ 0.01} & 0.13 {\scriptsize $\pm$ 0.00} & 0.14 {\scriptsize $\pm$ 0.01} \\
 & 11 ($\uparrow$) & 0.94 {\scriptsize $\pm$ 0.00} & \textbf{0.95} {\scriptsize $\pm$ 0.00} & \textbf{0.95} {\scriptsize $\pm$ 0.00} & 0.80 {\scriptsize $\pm$ 0.00} \\
 & 12 ($\downarrow$) & \textbf{0.33} {\scriptsize $\pm$ 0.00} & \textbf{0.33} {\scriptsize $\pm$ 0.01} & \textbf{0.33} {\scriptsize $\pm$ 0.00} & \textbf{0.33} {\scriptsize $\pm$ 0.00} \\
 & 13 ($\downarrow$) & \textbf{0.35} {\scriptsize $\pm$ 0.00} & 0.36 {\scriptsize $\pm$ 0.00} & 0.36 {\scriptsize $\pm$ 0.01} & 0.36 {\scriptsize $\pm$ 0.00} \\
 & 14 ($\downarrow$) & 0.91 {\scriptsize $\pm$ 0.01} & 1.56 {\scriptsize $\pm$ 0.75} & 0.80 {\scriptsize $\pm$ 0.01} & \textbf{0.78} {\scriptsize $\pm$ 0.01} \\
 & 15 ($\uparrow$) & $-\infty$ & $-\infty$ & $-\infty$ & \textbf{1311.06} {\scriptsize $\pm$ 27.18} \\
 & 16 ($\uparrow$) & \textbf{59.05} {\scriptsize $\pm$ 4.95} & $-\infty$ & $-\infty$ & $-\infty$ \\
\midrule
\multirow{2}{*}{MECHBench} & 17 ($\downarrow$) & \textbf{-8528.06} {\scriptsize $\pm$ 6.86} & -8471.56 {\scriptsize $\pm$ 98.17} & $\infty$ & -8525.20 {\scriptsize $\pm$ 15.63} \\
 & 18 ($\downarrow$) & \textbf{3.61} {\scriptsize $\pm$ 0.00} & \textbf{3.61} {\scriptsize $\pm$ 0.00} & \textbf{3.61} {\scriptsize $\pm$ 0.00} & \textbf{3.61} {\scriptsize $\pm$ 0.00} \\
\midrule
\multirow{5}{*}{MuJoCo} & 19 ($\uparrow$) & 191.43 {\scriptsize $\pm$ 149.50} & $-\infty$ & 24.00 {\scriptsize $\pm$ 0.00} & \textbf{1000.00} {\scriptsize $\pm$ 0.00} \\
 & 20 ($\uparrow$) & \textbf{101.22} {\scriptsize $\pm$ 5.00} & $-\infty$ & 93.84 {\scriptsize $\pm$ 1.08} & 90.41 {\scriptsize $\pm$ 0.00} \\
 & 21 ($\uparrow$) & \textbf{138.52} {\scriptsize $\pm$ 0.00} & \textbf{138.52} {\scriptsize $\pm$ 0.00} & \textbf{138.52} {\scriptsize $\pm$ 0.00} & \textbf{138.52} {\scriptsize $\pm$ 0.00} \\
 & 22 ($\uparrow$) & \textbf{-12.25} {\scriptsize $\pm$ 0.00} & \textbf{-12.25} {\scriptsize $\pm$ 0.00} & \textbf{-12.25} {\scriptsize $\pm$ 0.00} & \textbf{-12.25} {\scriptsize $\pm$ 0.00} \\
 & 23 ($\uparrow$) & 0.02 {\scriptsize $\pm$ 0.00} & \textbf{14.33} {\scriptsize $\pm$ 12.99} & 0.02 {\scriptsize $\pm$ 0.00} & 0.02 {\scriptsize $\pm$ 0.00} \\
\bottomrule
\end{tabular}
\end{table}


\begin{table}[htbp]
\centering
\caption{Comparison of optimization performance for \textit{Easy} difficulty and \textit{Large} budget. Values are mean $\pm$ standard deviation over three runs. The symbol ``--'' indicates that the run was terminated due to more than 48 hours of computational time. The symbol ``$\dagger$'' marks cells backed by a single seed, for which the standard deviation is undefined.}
\label{tab:results_easy_large}
\begin{tabular}{llrrrr}
\toprule
Domain & \# Task & Gemini 3 Flash & Gemini 3.1 Pro & GPT-5 mini & GPT-5.2 \\
\midrule
\multirow{7}{*}{YAHPO} & 1 ($\uparrow$) & \textbf{99.66} {\scriptsize $\pm$ 0.00} & \textbf{99.66} {\scriptsize $\pm$ 0.00} & \textbf{99.66} {\scriptsize $\pm$ 0.00} & 99.57 {\scriptsize $\pm$ 0.00} \\
 & 2 ($\uparrow$) & \textbf{0.97} {\scriptsize $\pm$ 0.00} & 0.96 {\scriptsize $\pm$ 0.00} & \textbf{0.97} {\scriptsize $\pm$ 0.00} & \textbf{0.97} {\scriptsize $\pm$ 0.00} \\
 & 3 ($\uparrow$) & 0.96 {\scriptsize $\pm$ 0.00} & \textbf{0.97} {\scriptsize $\pm$ 0.00} & 0.96 {\scriptsize $\pm$ 0.00} & 0.96 {\scriptsize $\pm$ 0.00} \\
 & 4 ($\uparrow$) & \textbf{0.92} {\scriptsize $\pm$ 0.00} & \textbf{0.92} {\scriptsize $\pm$ 0.00} & \textbf{0.92} {\scriptsize $\pm$ 0.00} & \textbf{0.92} {\scriptsize $\pm$ 0.00} \\
 & 5 ($\uparrow$) & \textbf{0.98} {\scriptsize $\pm$ 0.00} & \textbf{0.98} {\scriptsize $\pm$ 0.00} & \textbf{0.98} {\scriptsize $\pm$ 0.00} & \textbf{0.98} {\scriptsize $\pm$ 0.00} \\
 & 6 ($\uparrow$) & \textbf{1.00} {\scriptsize $\pm$ 0.00} & 0.99 {\scriptsize $\pm$ 0.00} & 0.99 {\scriptsize $\pm$ 0.00} & -- \\
 & 7 ($\uparrow$) & \textbf{1.00} {\scriptsize $\pm$ 0.00} & -- & \textbf{1.00}$^{\dagger}$ & -- \\
\midrule
\multirow{9}{*}{Olympus} & 8 ($\downarrow$) & 0.58 {\scriptsize $\pm$ 1.01} & 1.57 {\scriptsize $\pm$ 0.59} & \textbf{0.46} {\scriptsize $\pm$ 0.75} & 0.94 {\scriptsize $\pm$ 0.88} \\
 & 9 ($\downarrow$) & \textbf{0.02} {\scriptsize $\pm$ 0.00} & 0.03 {\scriptsize $\pm$ 0.00} & 0.03 {\scriptsize $\pm$ 0.00} & \textbf{0.02} {\scriptsize $\pm$ 0.00} \\
 & 10 ($\downarrow$) & 0.03 {\scriptsize $\pm$ 0.01} & \textbf{0.01} {\scriptsize $\pm$ 0.01} & 0.03 {\scriptsize $\pm$ 0.01} & 0.02 {\scriptsize $\pm$ 0.01} \\
 & 11 ($\uparrow$) & \textbf{0.96} {\scriptsize $\pm$ 0.00} & \textbf{0.96} {\scriptsize $\pm$ 0.00} & \textbf{0.96} {\scriptsize $\pm$ 0.00} & \textbf{0.96} {\scriptsize $\pm$ 0.00} \\
 & 12 ($\downarrow$) & \textbf{0.00} {\scriptsize $\pm$ 0.00} & \textbf{0.00} {\scriptsize $\pm$ 0.00} & \textbf{0.00} {\scriptsize $\pm$ 0.00} & \textbf{0.00} {\scriptsize $\pm$ 0.00} \\
 & 13 ($\downarrow$) & \textbf{0.00} {\scriptsize $\pm$ 0.00} & \textbf{0.00} {\scriptsize $\pm$ 0.00} & \textbf{0.00} {\scriptsize $\pm$ 0.00} & \textbf{0.00} {\scriptsize $\pm$ 0.00} \\
 & 14 ($\downarrow$) & \textbf{0.19} {\scriptsize $\pm$ 0.01} & 0.21 {\scriptsize $\pm$ 0.03} & 0.20 {\scriptsize $\pm$ 0.01} & 0.22 {\scriptsize $\pm$ 0.01} \\
 & 15 ($\uparrow$) & 2652.48 {\scriptsize $\pm$ 26.79} & 2668.36 {\scriptsize $\pm$ 40.92} & \textbf{2709.47} {\scriptsize $\pm$ 73.65} & 2687.61 {\scriptsize $\pm$ 62.12} \\
 & 16 ($\uparrow$) & \textbf{100.60} {\scriptsize $\pm$ 0.99} & 99.80 {\scriptsize $\pm$ 0.43} & 100.10 {\scriptsize $\pm$ 1.14} & 100.15 {\scriptsize $\pm$ 0.47} \\
\midrule
\multirow{5}{*}{MuJoCo} & 19 ($\uparrow$) & \textbf{1000.00} {\scriptsize $\pm$ 0.00} & \textbf{1000.00} {\scriptsize $\pm$ 0.00} & \textbf{1000.00} {\scriptsize $\pm$ 0.00} & \textbf{1000.00} {\scriptsize $\pm$ 0.00} \\
 & 20 ($\uparrow$) & 264.61 {\scriptsize $\pm$ 22.92} & \textbf{280.25} {\scriptsize $\pm$ 120.64} & \textbf{280.25} {\scriptsize $\pm$ 120.64} & 254.31 {\scriptsize $\pm$ 60.15} \\
 & 21 ($\uparrow$) & 1141.18 {\scriptsize $\pm$ 70.04} & 1595.68 {\scriptsize $\pm$ 863.82} & 1256.67 {\scriptsize $\pm$ 219.39} & \textbf{1758.61} {\scriptsize $\pm$ 895.58} \\
 & 22 ($\uparrow$) & -66.86 {\scriptsize $\pm$ 8.80} & \textbf{-8.28} {\scriptsize $\pm$ 4.35} & -86.53 {\scriptsize $\pm$ 17.26} & -66.86 {\scriptsize $\pm$ 8.80} \\
 & 23 ($\uparrow$) & \textbf{354.84} {\scriptsize $\pm$ 0.40} & 296.08 {\scriptsize $\pm$ 111.11} & 0.02 {\scriptsize $\pm$ 0.00} & \textbf{354.84} {\scriptsize $\pm$ 0.40} \\
\bottomrule
\end{tabular}
\end{table}

\begin{table}[htbp]
\centering
\caption{Comparison of optimization performance for \textit{Intermediate} difficulty and \textit{Large} budget. Values are mean $\pm$ standard deviation over three runs. The symbol ``--'' indicates that the run was terminated due to more than 48 hours of computational time.}
\label{tab:results_intermediate_large}
\begin{tabular}{llrrrr}
\toprule
Domain & \# Task & Gemini 3 Flash & Gemini 3.1 Pro & GPT-5 mini & GPT-5.2 \\
\midrule
\multirow{6}{*}{YAHPO} & 1 ($\uparrow$) & -- & \textbf{99.66} {\scriptsize $\pm$ 0.00} & 99.63 {\scriptsize $\pm$ 0.00} & 99.57 {\scriptsize $\pm$ 0.00} \\
 & 2 ($\uparrow$) & -- & \textbf{0.96} {\scriptsize $\pm$ 0.00} & \textbf{0.96} {\scriptsize $\pm$ 0.00} & -- \\
 & 3 ($\uparrow$) & -- & \textbf{0.91} {\scriptsize $\pm$ 0.00} & -- & -- \\
 & 4 ($\uparrow$) & -- & \textbf{0.92} {\scriptsize $\pm$ 0.00} & \textbf{0.92} {\scriptsize $\pm$ 0.00} & \textbf{0.92} {\scriptsize $\pm$ 0.00} \\
 & 5 ($\uparrow$) & 0.95 {\scriptsize $\pm$ 0.00} & \textbf{0.96} {\scriptsize $\pm$ 0.00} & -- & -- \\
 & 6 ($\uparrow$) & -- & \textbf{0.99} {\scriptsize $\pm$ 0.00} & -- & -- \\
 & 7 ($\uparrow$) & -- & -- & -- & -- \\
\midrule
\multirow{9}{*}{Olympus} & 8 ($\downarrow$) & $\infty$ & $\infty$ & $\infty$ & $\infty$ \\
 & 9 ($\downarrow$) & 0.15 {\scriptsize $\pm$ 0.00} & 0.15 {\scriptsize $\pm$ 0.01} & 0.15 {\scriptsize $\pm$ 0.00} & \textbf{0.08} {\scriptsize $\pm$ 0.01} \\
 & 10 ($\downarrow$) & \textbf{0.30} {\scriptsize $\pm$ 0.00} & \textbf{0.30} {\scriptsize $\pm$ 0.00} & \textbf{0.30} {\scriptsize $\pm$ 0.00} & \textbf{0.30} {\scriptsize $\pm$ 0.00} \\
 & 11 ($\uparrow$) & \textbf{0.96} {\scriptsize $\pm$ 0.00} & \textbf{0.96} {\scriptsize $\pm$ 0.00} & 0.95 {\scriptsize $\pm$ 0.00} & 0.93 {\scriptsize $\pm$ 0.00} \\
 & 12 ($\downarrow$) & \textbf{0.31} {\scriptsize $\pm$ 0.01} & \textbf{0.31} {\scriptsize $\pm$ 0.00} & -- & \textbf{0.31} {\scriptsize $\pm$ 0.00} \\
 & 13 ($\downarrow$) & 0.34 {\scriptsize $\pm$ 0.00} & \textbf{0.33} {\scriptsize $\pm$ 0.00} & 0.34 {\scriptsize $\pm$ 0.00} & \textbf{0.33} {\scriptsize $\pm$ 0.00} \\
 & 14 ($\downarrow$) & \textbf{0.62} {\scriptsize $\pm$ 0.01} & $\infty$ & 0.71 {\scriptsize $\pm$ 0.00} & 1.15 {\scriptsize $\pm$ 0.52} \\
 & 15 ($\uparrow$) & $-\infty$ & $-\infty$ & \textbf{1420.01} {\scriptsize $\pm$ 30.67} & 1382.04 {\scriptsize $\pm$ 21.20} \\
 & 16 ($\uparrow$) & -- & $-\infty$ & \textbf{82.64} {\scriptsize $\pm$ 0.27} & -- \\
\midrule
\multirow{5}{*}{MuJoCo} & 19 ($\uparrow$) & \textbf{1000.00} {\scriptsize $\pm$ 0.00} & \textbf{1000.00} {\scriptsize $\pm$ 0.00} & \textbf{1000.00} {\scriptsize $\pm$ 0.00} & \textbf{1000.00} {\scriptsize $\pm$ 0.00} \\
 & 20 ($\uparrow$) & 100.72 {\scriptsize $\pm$ 0.00} & 100.72 {\scriptsize $\pm$ 0.00} & 124.30 {\scriptsize $\pm$ 6.62} & \textbf{129.54} {\scriptsize $\pm$ 13.98} \\
 & 21 ($\uparrow$) & \textbf{138.52} {\scriptsize $\pm$ 0.00} & \textbf{138.52} {\scriptsize $\pm$ 0.00} & \textbf{138.52} {\scriptsize $\pm$ 0.00} & \textbf{138.52} {\scriptsize $\pm$ 0.00} \\
 & 22 ($\uparrow$) & \textbf{-12.25} {\scriptsize $\pm$ 0.00} & \textbf{-12.25} {\scriptsize $\pm$ 0.00} & \textbf{-12.25} {\scriptsize $\pm$ 0.00} & \textbf{-12.25} {\scriptsize $\pm$ 0.00} \\
 & 23 ($\uparrow$) & \textbf{0.02} {\scriptsize $\pm$ 0.00} & \textbf{0.02} {\scriptsize $\pm$ 0.00} & \textbf{0.02} {\scriptsize $\pm$ 0.00} & \textbf{0.02} {\scriptsize $\pm$ 0.00} \\
\bottomrule
\end{tabular}
\end{table}

\begin{table}[htbp]
\centering
\caption{Comparison of optimization performance for \textit{Hard} difficulty and \textit{Large} budget. Values are mean $\pm$ standard deviation over three runs. The symbol ``--'' indicates that the run was terminated due to more than 48 hours of computational time. The symbol ``$\dagger$'' marks cells backed by a single seed, for which the standard deviation is undefined.}
\label{tab:results_hard_large}
\begin{tabular}{llrrrr}
\toprule
Domain & \# Task & Gemini 3 Flash & Gemini 3.1 Pro & GPT-5 mini & GPT-5.2 \\
\midrule
\multirow{7}{*}{YAHPO} & 1 ($\uparrow$) & \textbf{99.50} {\scriptsize $\pm$ 0.00} & $-\infty$ & 98.45 {\scriptsize $\pm$ 0.00} & \textbf{99.50} {\scriptsize $\pm$ 0.00} \\
 & 2 ($\uparrow$) & -- & \textbf{0.96} {\scriptsize $\pm$ 0.00} & -- & -- \\
 & 3 ($\uparrow$) & \textbf{0.91}$^{\dagger}$ & \textbf{0.91} {\scriptsize $\pm$ 0.00} & 0.89 {\scriptsize $\pm$ 0.00} & 0.86$^{\dagger}$ \\
 & 4 ($\uparrow$) & -- & \textbf{0.92} {\scriptsize $\pm$ 0.00} & -- & -- \\
 & 5 ($\uparrow$) & -- & \textbf{0.96} {\scriptsize $\pm$ 0.00} & \textbf{0.96} {\scriptsize $\pm$ 0.00} & 0.95 {\scriptsize $\pm$ 0.00} \\
 & 6 ($\uparrow$) & -- & \textbf{0.99} {\scriptsize $\pm$ 0.00} & \textbf{0.99} {\scriptsize $\pm$ 0.00} & $-\infty$ \\
 & 7 ($\uparrow$) & -- & -- & -- & \textbf{0.91}$^{\dagger}$ \\
\midrule
\multirow{9}{*}{Olympus} & 8 ($\downarrow$) & -- & $\infty$ & $\infty$ & $\infty$ \\
 & 9 ($\downarrow$) & 0.03 {\scriptsize $\pm$ 0.01} & \textbf{0.02} {\scriptsize $\pm$ 0.00} & \textbf{0.02} {\scriptsize $\pm$ 0.00} & 0.09 {\scriptsize $\pm$ 0.00} \\
 & 10 ($\downarrow$) & 0.10 {\scriptsize $\pm$ 0.00} & \textbf{0.02} {\scriptsize $\pm$ 0.01} & 0.18 {\scriptsize $\pm$ 0.01} & 0.10 {\scriptsize $\pm$ 0.00} \\
 & 11 ($\uparrow$) & -- & $-\infty$ & -- & \textbf{0.81} {\scriptsize $\pm$ 0.00} \\
 & 12 ($\downarrow$) & 0.31 {\scriptsize $\pm$ 0.00} & 0.31 {\scriptsize $\pm$ 0.00} & \textbf{0.30} {\scriptsize $\pm$ 0.01} & 0.31 {\scriptsize $\pm$ 0.01} \\
 & 13 ($\downarrow$) & 0.34 {\scriptsize $\pm$ 0.01} & 0.34 {\scriptsize $\pm$ 0.00} & \textbf{0.33} {\scriptsize $\pm$ 0.00} & 0.34 {\scriptsize $\pm$ 0.00} \\
 & 14 ($\downarrow$) & -- & \textbf{0.65} {\scriptsize $\pm$ 0.00} & -- & -- \\
 & 15 ($\uparrow$) & $-\infty$ & $-\infty$ & $-\infty$ & \textbf{1376.90} {\scriptsize $\pm$ 13.86} \\
 & 16 ($\uparrow$) & -- & $-\infty$ & $-\infty$ & -- \\
\midrule
\multirow{5}{*}{MuJoCo} & 19 ($\uparrow$) & \textbf{1000.00} {\scriptsize $\pm$ 0.00} & \textbf{1000.00} {\scriptsize $\pm$ 0.00} & 460.23 {\scriptsize $\pm$ 0.06} & 229.40 {\scriptsize $\pm$ 0.00} \\
 & 20 ($\uparrow$) & 123.50 {\scriptsize $\pm$ 9.59} & $-\infty$ & \textbf{325.58} {\scriptsize $\pm$ 3.57} & 90.41 {\scriptsize $\pm$ 0.00} \\
 & 21 ($\uparrow$) & -- & \textbf{138.52} {\scriptsize $\pm$ 0.00} & \textbf{138.52} {\scriptsize $\pm$ 0.00} & \textbf{138.52} {\scriptsize $\pm$ 0.00} \\
 & 22 ($\uparrow$) & \textbf{-12.25}$^{\dagger}$ & \textbf{-12.25} {\scriptsize $\pm$ 0.00} & \textbf{-12.25} {\scriptsize $\pm$ 0.00} & \textbf{-12.25} {\scriptsize $\pm$ 0.00} \\
 & 23 ($\uparrow$) & \textbf{0.02} {\scriptsize $\pm$ 0.00} & \textbf{0.02} {\scriptsize $\pm$ 0.00} & \textbf{0.02} {\scriptsize $\pm$ 0.00} & \textbf{0.02} {\scriptsize $\pm$ 0.00} \\
\bottomrule
\end{tabular}
\end{table}

\subsection{One-Shot Settings} \label{app:experiment:oneshot}
Similarly to zero-shot settings, we conducted experiments on 22 tasks, excluding Task 19, in BBOWP-Bench for \textit{Small} budget ($N_{\max} = 100$). We added Task 19's description and output into the system prompt as a one-shot example.
The results are provided in \cref{tab:results_oneshot_easy_small} for \textit{Easy} difficulty, \cref{tab:results_oneshot_intermediate_small} for \textit{Intermediate} difficulty, and \cref{tab:results_oneshot_hard_small} for \textit{Hard} difficulty.

Additionally, we conducted experiments on 20 tasks for \textit{Large} budget ($N_{\max} = 10{,}000$).
Tasks 17 and 18 in MECHBench are excluded.
Task 19 in MuJoCo is also omitted because it is included in the system prompt as the example.
The results are provided in \cref{tab:results_oneshot_easy_large} for \textit{Easy} difficulty, \cref{tab:results_oneshot_intermediate_large} for \textit{Intermediate} difficulty, and \cref{tab:results_oneshot_hard_large} for \textit{Hard} difficulty.
We terminated the optimization runs that exceeded 48 hours and omitted their results from each table, denoting such runs as ``--''.


\begin{table}[htbp]
\centering
\caption{Comparison of optimization performance with one-shot prompting for \textit{Easy} difficulty and \textit{Small} budget.}
\label{tab:results_oneshot_easy_small}
\begin{tabular}{llrrrr}
\toprule
Domain & Task & Gemini 3 Flash & Gemini 3.1 Pro & GPT-5 mini & GPT-5.2 \\
\midrule
\multirow{7}{*}{YAHPO} & Task 1 ($\uparrow$) & 97.01 & 96.12 & \textbf{99.55} & 98.11 \\
 & Task 2 ($\uparrow$) & 0.95 & \textbf{0.96} & \textbf{0.96} & \textbf{0.96} \\
 & Task 3 ($\uparrow$) & 0.88 & 0.86 & 0.91 & \textbf{0.94} \\
 & Task 4 ($\uparrow$) & \textbf{0.92} & 0.91 & \textbf{0.92} & \textbf{0.92} \\
 & Task 5 ($\uparrow$) & 0.95 & 0.95 & \textbf{0.98} & 0.95 \\
 & Task 6 ($\uparrow$) & \textbf{0.99} & \textbf{0.99} & \textbf{0.99} & \textbf{0.99} \\
 & Task 7 ($\uparrow$) & 0.98 & 0.97 & 0.99 & \textbf{1.00} \\
\midrule
\multirow{9}{*}{Olympus} & Task 8 ($\downarrow$) & 2.54 & \textbf{2.30} & 2.73 & 2.64 \\
 & Task 9 ($\downarrow$) & \textbf{0.03} & \textbf{0.03} & \textbf{0.03} & 0.06 \\
 & Task 10 ($\downarrow$) & \textbf{0.04} & 0.05 & 0.08 & 0.05 \\
 & Task 11 ($\uparrow$) & \textbf{0.96} & \textbf{0.96} & 0.95 & 0.95 \\
 & Task 12 ($\downarrow$) & \textbf{0.00} & \textbf{0.00} & 0.06 & 0.12 \\
 & Task 13 ($\downarrow$) & \textbf{0.00} & \textbf{0.00} & 0.09 & 0.01 \\
 & Task 14 ($\downarrow$) & \textbf{0.19} & 0.20 & 0.38 & 0.34 \\
 & Task 15 ($\uparrow$) & 2498.78 & \textbf{2511.32} & 2211.82 & 2295.25 \\
 & Task 16 ($\uparrow$) & 99.69 & \textbf{100.86} & 90.31 & 91.29 \\
\midrule
\multirow{2}{*}{MECHBench} & Task 17 ($\downarrow$) & \textbf{-11687.66} & -11602.07 & -11298.41 & -10086.18 \\
 & Task 18 ($\downarrow$) & 3.41 & 3.41 & 4.68 & \textbf{3.40} \\
\midrule
\multirow{4}{*}{MuJoCo} & Task 20 ($\uparrow$) & 243.67 & 156.76 & 236.18 & \textbf{376.58} \\
 & Task 21 ($\uparrow$) & 999.21 & 836.11 & \textbf{1021.98} & 367.91 \\
 & Task 22 ($\uparrow$) & -49.59 & -90.06 & -50.74 & \textbf{-16.72} \\
 & Task 23 ($\uparrow$) & \textbf{357.92} & 346.59 & 34.92 & 350.42 \\
\bottomrule
\end{tabular}
\end{table}

\begin{table}[htbp]
\centering
\caption{Comparison of optimization performance with one-shot prompting for \textit{Intermediate} difficulty and \textit{Small} budget.}
\label{tab:results_oneshot_intermediate_small}
\begin{tabular}{llrrrr}
\toprule
Domain & Task & Gemini 3 Flash & Gemini 3.1 Pro & GPT-5 mini & GPT-5.2 \\
\midrule
\multirow{7}{*}{YAHPO} & Task 1 ($\uparrow$) & 98.33 & 95.25 & 99.36 & \textbf{99.45} \\
 & Task 2 ($\uparrow$) & 0.95 & \textbf{0.96} & 0.94 & 0.92 \\
 & Task 3 ($\uparrow$) & 0.84 & 0.79 & \textbf{0.89} & 0.81 \\
 & Task 4 ($\uparrow$) & 0.90 & \textbf{0.92} & \textbf{0.92} & 0.91 \\
 & Task 5 ($\uparrow$) & 0.94 & 0.95 & \textbf{0.96} & 0.95 \\
 & Task 6 ($\uparrow$) & \textbf{0.99} & \textbf{0.99} & \textbf{0.99} & \textbf{0.99} \\
 & Task 7 ($\uparrow$) & 0.91 & 0.91 & 0.91 & \textbf{0.92} \\
\midrule
\multirow{9}{*}{Olympus} & Task 8 ($\downarrow$) & 9.06 & \textbf{3.26} & $\infty$ & $\infty$ \\
 & Task 9 ($\downarrow$) & 0.15 & \textbf{0.12} & 0.18 & 0.16 \\
 & Task 10 ($\downarrow$) & \textbf{0.31} & \textbf{0.31} & \textbf{0.31} & \textbf{0.31} \\
 & Task 11 ($\uparrow$) & \textbf{0.95} & 0.90 & 0.89 & 0.93 \\
 & Task 12 ($\downarrow$) & 0.34 & 0.34 & \textbf{0.32} & 0.33 \\
 & Task 13 ($\downarrow$) & 0.36 & 0.35 & \textbf{0.34} & 0.36 \\
 & Task 14 ($\downarrow$) & $\infty$ & $\infty$ & \textbf{0.80} & 0.91 \\
 & Task 15 ($\uparrow$) & $-\infty$ & $-\infty$ & $-\infty$ & \textbf{1274.15} \\
 & Task 16 ($\uparrow$) & $-\infty$ & $-\infty$ & \textbf{81.96} & 60.14 \\
\midrule
\multirow{2}{*}{MECHBench} & Task 17 ($\downarrow$) & \textbf{-8538.19} & -8528.84 & $\infty$ & 800.31 \\
 & Task 18 ($\downarrow$) & \textbf{3.61} & \textbf{3.61} & \textbf{3.61} & \textbf{3.61} \\
\midrule
\multirow{4}{*}{MuJoCo} & Task 20 ($\uparrow$) & 90.41 & 94.14 & 90.41 & \textbf{176.75} \\
 & Task 21 ($\uparrow$) & 998.72 & 138.52 & 138.52 & \textbf{1055.56} \\
 & Task 22 ($\uparrow$) & \textbf{-12.25} & \textbf{-12.25} & \textbf{-12.25} & $-\infty$ \\
 & Task 23 ($\uparrow$) & \textbf{0.02} & \textbf{0.02} & \textbf{0.02} & \textbf{0.02} \\
\bottomrule
\end{tabular}
\end{table}

\begin{table}[htbp]
\centering
\caption{Comparison of optimization performance with one-shot prompting for \textit{Hard} difficulty and \textit{Small} budget.}
\label{tab:results_oneshot_hard_small}
\begin{tabular}{llrrrr}
\toprule
Domain & Task & Gemini 3 Flash & Gemini 3.1 Pro & GPT-5 mini & GPT-5.2 \\
\midrule
\multirow{7}{*}{YAHPO} & Task 1 ($\uparrow$) & \textbf{97.78} & 96.52 & $-\infty$ & $-\infty$ \\
 & Task 2 ($\uparrow$) & \textbf{0.94} & 0.93 & 0.93 & 0.92 \\
 & Task 3 ($\uparrow$) & \textbf{0.86} & 0.76 & 0.84 & 0.83 \\
 & Task 4 ($\uparrow$) & \textbf{0.92} & \textbf{0.92} & \textbf{0.92} & 0.91 \\
 & Task 5 ($\uparrow$) & \textbf{0.95} & \textbf{0.95} & 0.94 & \textbf{0.95} \\
 & Task 6 ($\uparrow$) & \textbf{0.99} & $-\infty$ & \textbf{0.99} & \textbf{0.99} \\
 & Task 7 ($\uparrow$) & \textbf{0.91} & \textbf{0.91} & \textbf{0.91} & \textbf{0.91} \\
\midrule
\multirow{9}{*}{Olympus} & Task 8 ($\downarrow$) & $\infty$ & $\infty$ & $\infty$ & $\infty$ \\
 & Task 9 ($\downarrow$) & \textbf{0.03} & 0.05 & 0.04 & 0.04 \\
 & Task 10 ($\downarrow$) & 0.31 & \textbf{0.12} & 0.13 & 0.31 \\
 & Task 11 ($\uparrow$) & \textbf{0.95} & \textbf{0.95} & \textbf{0.95} & 0.80 \\
 & Task 12 ($\downarrow$) & \textbf{0.33} & 0.34 & \textbf{0.33} & \textbf{0.33} \\
 & Task 13 ($\downarrow$) & \textbf{0.35} & 0.36 & 0.36 & 0.36 \\
 & Task 14 ($\downarrow$) & 0.83 & 0.84 & \textbf{0.79} & 1.23 \\
 & Task 15 ($\uparrow$) & $-\infty$ & $-\infty$ & \textbf{1316.15} & 1257.63 \\
 & Task 16 ($\uparrow$) & $-\infty$ & $-\infty$ & $-\infty$ & $-\infty$ \\
\midrule
\multirow{2}{*}{MECHBench} & Task 17 ($\downarrow$) & \textbf{-8503.76} & -8482.05 & $\infty$ & 800.31 \\
 & Task 18 ($\downarrow$) & \textbf{3.61} & \textbf{3.61} & \textbf{3.61} & \textbf{3.61} \\
\midrule
\multirow{4}{*}{MuJoCo} & Task 20 ($\uparrow$) & 92.00 & 98.08 & \textbf{189.01} & 96.43 \\
 & Task 21 ($\uparrow$) & \textbf{138.52} & \textbf{138.52} & \textbf{138.52} & \textbf{138.52} \\
 & Task 22 ($\uparrow$) & \textbf{-12.25} & \textbf{-12.25} & \textbf{-12.25} & $-\infty$ \\
 & Task 23 ($\uparrow$) & \textbf{0.02} & \textbf{0.02} & \textbf{0.02} & \textbf{0.02} \\
\bottomrule
\end{tabular}
\end{table}

\begin{table}[htbp]
\centering
\caption{Comparison of optimization performance with one-shot prompting for \textit{Easy} difficulty and \textit{Large} budget. The symbol ``--'' indicates that the run was terminated due to more than 48 hours of computational time.}
\label{tab:results_oneshot_easy_large}
\begin{tabular}{llrrrr}
\toprule
Domain & Task & Gemini 3 Flash & Gemini 3.1 Pro & GPT-5 mini & GPT-5.2 \\
\midrule
\multirow{7}{*}{YAHPO} & Task 1 ($\uparrow$) & -- & \textbf{99.66} & $-\infty$ & 99.57 \\
 & Task 2 ($\uparrow$) & \textbf{0.96} & \textbf{0.96} & \textbf{0.96} & \textbf{0.96} \\
 & Task 3 ($\uparrow$) & \textbf{0.96} & 0.91 & \textbf{0.96} & \textbf{0.96} \\
 & Task 4 ($\uparrow$) & \textbf{0.92} & \textbf{0.92} & \textbf{0.92} & \textbf{0.92} \\
 & Task 5 ($\uparrow$) & 0.96 & \textbf{0.98} & \textbf{0.98} & 0.96 \\
 & Task 6 ($\uparrow$) & -- & 0.99 & \textbf{1.00} & 0.99 \\
 & Task 7 ($\uparrow$) & -- & \textbf{1.00} & -- & \textbf{1.00} \\
\midrule
\multirow{9}{*}{Olympus} & Task 8 ($\downarrow$) & \textbf{0.30} & 0.83 & 2.20 & 1.97 \\
 & Task 9 ($\downarrow$) & \textbf{0.02} & \textbf{0.02} & \textbf{0.02} & \textbf{0.02} \\
 & Task 10 ($\downarrow$) & \textbf{0.00} & 0.01 & 0.03 & 0.03 \\
 & Task 11 ($\uparrow$) & \textbf{0.96} & \textbf{0.96} & \textbf{0.96} & \textbf{0.96} \\
 & Task 12 ($\downarrow$) & \textbf{0.00} & \textbf{0.00} & -- & \textbf{0.00} \\
 & Task 13 ($\downarrow$) & \textbf{0.00} & \textbf{0.00} & \textbf{0.00} & \textbf{0.00} \\
 & Task 14 ($\downarrow$) & 0.22 & 0.23 & \textbf{0.21} & 0.27 \\
 & Task 15 ($\uparrow$) & 2532.92 & 2692.48 & \textbf{2704.45} & 2568.60 \\
 & Task 16 ($\uparrow$) & \textbf{100.00} & 97.88 & 97.58 & 99.92 \\
\midrule
\multirow{4}{*}{MuJoCo} & Task 20 ($\uparrow$) & 443.79 & 456.84 & 9359.43 & \textbf{9359.86} \\
 & Task 21 ($\uparrow$) & 1029.72 & \textbf{2589.99} & 1199.98 & 1347.54 \\
 & Task 22 ($\uparrow$) & -5.84 & -28.56 & -6.00 & \textbf{-5.04} \\
 & Task 23 ($\uparrow$) & 359.10 & 360.35 & 108.94 & \textbf{361.33} \\
\bottomrule
\end{tabular}
\end{table}

\begin{table}[htbp]
\centering
\caption{Comparison of optimization performance with one-shot prompting for \textit{Intermediate} difficulty and \textit{Large} budget. The symbol ``--'' indicates that the run was terminated due to more than 48 hours of computational time.}
\label{tab:results_oneshot_intermediate_large}
\begin{tabular}{llrrrr}
\toprule
Domain & Task & Gemini 3 Flash & Gemini 3.1 Pro & GPT-5 mini & GPT-5.2 \\
\midrule
\multirow{6}{*}{YAHPO} & Task 1 ($\uparrow$) & 98.53 & \textbf{99.57} & 99.50 & \textbf{99.57} \\
 & Task 2 ($\uparrow$) & -- & \textbf{0.96} & -- & -- \\
 & Task 3 ($\uparrow$) & -- & \textbf{0.91} & -- & 0.89 \\
 & Task 4 ($\uparrow$) & -- & \textbf{0.92} & \textbf{0.92} & -- \\
 & Task 5 ($\uparrow$) & -- & -- & -- & \textbf{0.95} \\
 & Task 6 ($\uparrow$) & \textbf{0.99} & \textbf{0.99} & -- & \textbf{0.99} \\
 & Task 7 ($\uparrow$) & -- & -- & -- & -- \\
\midrule
\multirow{9}{*}{Olympus} & Task 8 ($\downarrow$) & 0.22 & \textbf{0.16} & $\infty$ & 3.00 \\
 & Task 9 ($\downarrow$) & 0.15 & 0.15 & 0.15 & \textbf{0.03} \\
 & Task 10 ($\downarrow$) & \textbf{0.29} & 0.30 & 0.30 & 0.30 \\
 & Task 11 ($\uparrow$) & \textbf{0.96} & \textbf{0.96} & \textbf{0.96} & 0.93 \\
 & Task 12 ($\downarrow$) & \textbf{0.30} & 0.31 & \textbf{0.30} & 0.31 \\
 & Task 13 ($\downarrow$) & \textbf{0.34} & \textbf{0.34} & \textbf{0.34} & \textbf{0.34} \\
 & Task 14 ($\downarrow$) & \textbf{0.63} & $\infty$ & 0.68 & $\infty$ \\
 & Task 15 ($\uparrow$) & $-\infty$ & $-\infty$ & $-\infty$ & -- \\
 & Task 16 ($\uparrow$) & 81.46 & $-\infty$ & \textbf{83.81} & 83.39 \\
\midrule
\multirow{4}{*}{MuJoCo} & Task 20 ($\uparrow$) & 112.44 & 123.07 & \textbf{363.04} & 123.07 \\
 & Task 21 ($\uparrow$) & \textbf{1039.23} & 138.52 & 138.52 & 138.52 \\
 & Task 22 ($\uparrow$) & \textbf{-12.25} & \textbf{-12.25} & \textbf{-12.25} & $-\infty$ \\
 & Task 23 ($\uparrow$) & \textbf{0.02} & \textbf{0.02} & \textbf{0.02} & \textbf{0.02} \\
\bottomrule
\end{tabular}
\end{table}

\begin{table}[htbp]
\centering
\caption{Comparison of optimization performance with one-shot prompting for \textit{Hard} difficulty and \textit{Large} budget. The symbol ``--'' indicates that the run was terminated due to more than 48 hours of computational time.}
\label{tab:results_oneshot_hard_large}
\begin{tabular}{llrrrr}
\toprule
Domain & Task & Gemini 3 Flash & Gemini 3.1 Pro & GPT-5 mini & GPT-5.2 \\
\midrule
\multirow{7}{*}{YAHPO} & Task 1 ($\uparrow$) & -- & \textbf{99.57} & $-\infty$ & $-\infty$ \\
 & Task 2 ($\uparrow$) & -- & 0.93 & -- & \textbf{0.96} \\
 & Task 3 ($\uparrow$) & 0.86 & \textbf{0.91} & 0.87 & -- \\
 & Task 4 ($\uparrow$) & -- & \textbf{0.92} & \textbf{0.92} & -- \\
 & Task 5 ($\uparrow$) & -- & \textbf{0.95} & \textbf{0.95} & \textbf{0.95} \\
 & Task 6 ($\uparrow$) & -- & \textbf{0.99} & -- & \textbf{0.99} \\
 & Task 7 ($\uparrow$) & -- & -- & 0.91 & \textbf{0.96} \\
\midrule
\multirow{8}{*}{Olympus} & Task 8 ($\downarrow$) & -- & $\infty$ & -- & $\infty$ \\
 & Task 9 ($\downarrow$) & 0.02 & 0.02 & \textbf{0.01} & 0.02 \\
 & Task 10 ($\downarrow$) & 0.30 & \textbf{0.02} & 0.11 & 0.12 \\
 & Task 11 ($\uparrow$) & -- & \textbf{0.96} & -- & -- \\
 & Task 12 ($\downarrow$) & 0.31 & 0.32 & 0.31 & \textbf{0.30} \\
 & Task 13 ($\downarrow$) & \textbf{0.33} & \textbf{0.33} & 0.34 & \textbf{0.33} \\
 & Task 14 ($\downarrow$) & -- & \textbf{0.66} & -- & -- \\
 & Task 15 ($\uparrow$) & -- & -- & $-\infty$ & \textbf{1390.82} \\
 & Task 16 ($\uparrow$) & -- & -- & -- & -- \\
\midrule
\multirow{4}{*}{MuJoCo} & Task 20 ($\uparrow$) & 100.72 & 88.20 & \textbf{323.52} & 260.96 \\
 & Task 21 ($\uparrow$) & 138.52 & \textbf{1104.08} & 138.52 & 138.52 \\
 & Task 22 ($\uparrow$) & \textbf{-12.25} & \textbf{-12.25} & \textbf{-12.25} & $-\infty$ \\
 & Task 23 ($\uparrow$) & \textbf{0.02} & \textbf{0.02} & -0.26 & \textbf{0.02} \\
\bottomrule
\end{tabular}
\end{table}

\clearpage

\section{Search Spaces Designed by LLMs}
\label{app:searchspace}

Tables \ref{tab:search_space_svm_app} to \ref{tab:search_space_double_pendulum_easy_app} present representative examples of the search space designed by the LLMs for specified tasks and settings: \cref{tab:search_space_svm_app} corresponds to Task 5 (HPO of SVM in YAHPO) in \textit{Hard} difficulty under \textit{Small} budget; \cref{tab:search_space_snar_app} corresponds to Task 14 (S${}_N$Ar reaction in Olympus) in \textit{Hard} difficulty under \textit{Small} budget; \cref{tab:search_space_starcrashbox_app} corresponds to Task 17 (Star-shaped crash box, MECHBench) in \textit{Hard} difficulty under \textit{Small} budget; 
Tables \ref{tab:search_space_pendulum_hard_app} and \ref{tab:search_space_pendulum_easy_app} correspond to Task 19 (Inverted Pendulum, MuJoCo) in \textit{Hard} and \textit{Easy} difficulty under \textit{Small} budget; Tables \ref{tab:search_space_double_pendulum_hard_app} and \ref{tab:search_space_double_pendulum_easy_app} correspond to Task 20 (Inverted Double Pendulum, MuJoCo) in \textit{Hard} and \textit{Easy} difficulty under \textit{Small} budget, respectively.

For Task 5 (HPO of SVM in YAHPO), the best objective values of all LLMs and the human reference are comparable since all LLMs correctly inferred several important variables in their search space, such as \texttt{cost}, \texttt{kernel}, and \texttt{gamma}.
However, only GPT-5 mini yielded a narrow range restricted to small values of [1e-4, 1.0] for \texttt{cost} and \textit{gamma}, which could result in a slightly lower best objective value than those of the other LLMs.

For Task 14 (S${}_N$Ar reaction in Olympus), the best objective values achieved by the LLM-designed search spaces and selected algorithms were worse than those of the human reference.
This gap is attributable to the suboptimal search space designs proposed by the LLMs: although most models correctly identified one or both of \texttt{residence\_time} and \texttt{temperature} as design variables, the specified value ranges were poorly calibrated.
Excessively long residence times promote side reactions, thereby increasing impurity, while excessively low temperatures suppress the reaction rate and reduce substrate conversion, both of which are detrimental to impurity minimization. 

The best objective values achieved by the LLM-designed search spaces and selected algorithms also fell short of the human reference in Task 17 (Star-shaped crash box, MECHBench). In the LLM-designed search spaces, only the variable \texttt{thickness} was mapped to the canonical variables.
The best objective value of GPT-5 mini was $\infty$ because its value range of [5e-4, 2e-2] was outside the canonical variable range, and those of the other LLMs did not reach the human reference.
The LLMs designed their search spaces based on a polar coordinate system, while a rectangular coordinate system was expected in the canonical design space.
This variation in coordinate system conventions for designing search space presents a significant challenge that lies beyond the scope of the current framework.
This topic is one of the directions of future work.

For Task 19 (Inverted Pendulum, MuJoCo), Gemini 3 Flash restricts the gains for the pole angle and angular velocity to non-negative ranges based on PD-control principles, and achieves the best objective value among the LLMs and the human reference.
In the \textit{Hard} setting, where the variable names are not provided, GPT-5.2 named each gain after its corresponding observation and reached the maximum episode length.
In contrast, Gemini 3.1 Pro and GPT-5 mini used the variable names based on indices that did not indicate which observation each gain multiplied, resulting in incorrect mappings to the design variables.

For Task 20 (Inverted Double Pendulum, MuJoCo), some LLMs tended to design their search spaces with some or all variables asymmetric about zero in \textit{Hard} difficulty under \textit{Small} budget. This asymmetric design potentially demonstrates that LLMs leverage their knowledge about control engineering for designing search spaces.

Based on the analyses above, although identifying appropriate variables and value ranges remains challenging for LLMs, the best objective values achieved in some tasks are comparable to those of the human reference.
These findings suggest that LLMs can leverage their domain knowledge to design effective search spaces, while also highlighting the importance of accurately specifying variable ranges and representations.

\begin{table}[htbp]
\centering
\caption{Comparison of search space designed by human and LLMs, and the best achieved objective value for Task 5 (HPO of SVM in YAHPO) in \textit{Hard} difficulty and \textit{Small} budget. The symbol ``$-$'' indicates the variable was not included in the search space provided by the model.}
\label{tab:search_space_svm_app}
\small
{\setlength{\tabcolsep}{4pt}
\begin{tabularx}{\linewidth}{
  >{\raggedright\arraybackslash}p{3.0cm}
  >{\raggedright\arraybackslash}X
  >{\raggedright\arraybackslash}X
  >{\raggedright\arraybackslash}X
  >{\raggedright\arraybackslash}X
  >{\raggedright\arraybackslash}X
  >{\raggedright\arraybackslash}X}
\toprule
Variables & Canonical Domain & Human Reference & Gemini 3 Flash & Gemini 3.1 Pro & GPT-5 mini & GPT-5.2 \\
\midrule
cost  & [4.54e-5, 22026.47]  & [4.54e-5, 22026.47], log  & [1e-3, 1e3], log  & [1e-3, 1e3], log  & [1e-4, 1.0], log  & [1e-3, 1e3], log \\
\noalign{\vspace{2.5pt}}
kernel  & \{linear, radial, polynomial\}  & \{linear, radial, polynomial\}  & \{linear, radial, polynomial\}  & \{linear, radial\}  & \{linear, radial, polynomial\}  & \{linear, radial\} \\
\noalign{\vspace{2.5pt}}
tolerance  & [4.54e-5, 2.0]  & [4.54e-5, 2.0], log  & $-$  & $-$  & $-$  & [1e-5, 1e-2], log \\
\noalign{\vspace{2.5pt}}
degree
  & [2, 5], step=1
  & [2, 5], step=1  & $-$  & $-$  & [2, 5], step=2  & $-$ \\
\noalign{\vspace{2.5pt}}
gamma
  & [4.54e-5, 22026.47], log
  & [4.54e-5, 22026.47], log  & [1e-4, 10.0], log  & [1e-4, 10.0], log  & [1e-4, 1.0], log  & [1e-4, 10.0], log \\
\noalign{\vspace{2.5pt}}
num.impute.selected.cpo  & \{mean, hist, median\}  & hist  & $-$  & $-$  & $-$  & $-$ \\
\midrule
\multicolumn{7}{l}{Additional variables} \\
\midrule
probability  & $-$  & $-$  & True  & True  & $-$  & False \\
\noalign{\vspace{2pt}}
coef0        & $-$  & $-$  & $-$   & $-$   & 0.0  & $-$ \\
class\_weight & $-$  & $-$  & $-$   & $-$   & $-$  & \{none, balanced\} \\
\noalign{\vspace{2.5pt}}
shrinking    & $-$  & $-$  & $-$   & $-$   & $-$  & \{True, False\} \\
max\_iter    & $-$  & $-$  & $-$   & $-$   & $-$  & -1 \\
\midrule
Best obj. ($\uparrow$)  & $-$  & 0.9765 {\scriptsize $\pm$ 0.0055} & 0.9498 {\scriptsize $\pm$ 0.0004}  & 0.9483 {\scriptsize $\pm$ 0.0055}  & 0.9334 {\scriptsize $\pm$ 0.0091}  & 0.9503 {\scriptsize $\pm$ 0.0002} \\

\bottomrule
\end{tabularx}}
\end{table}

\begin{table}[htbp]
\centering
\caption{Comparison of search space designed by human and LLMs, and the best achieved objective value for Task 14 (S${}_N$Ar reaction in Olympus) in \textit{Hard} difficulty and \textit{Small} budget. The symbol ``$-$'' indicates the variable was not included in the search space provided by the model.}
\label{tab:search_space_snar_app}
\small
{\setlength{\tabcolsep}{4pt}
\begin{tabularx}{\linewidth}{
  >{\raggedright\arraybackslash}p{2.8cm}
  >{\raggedright\arraybackslash}X
  >{\raggedright\arraybackslash}X
  >{\raggedright\arraybackslash}X
  >{\raggedright\arraybackslash}X
  >{\raggedright\arraybackslash}X
  >{\raggedright\arraybackslash}X}
\toprule
Variables & Canonical Domain & Human Reference & Gemini 3 Flash & Gemini 3.1 Pro & GPT-5 mini & GPT-5.2 \\
\midrule
residence\_time  & [0.5, 2.0]   & [0.5, 2.0]   & [0.0, 1440]  & [1.0, 24.0] & $-$     & $-$ \\
ratio            & [1.0, 5.0]   & [1.0, 5.0]   & $-$  & $-$ & $-$     & $-$ \\
concentration    & [0.1, 0.5]   & [0.1, 0.5]   & [0.01, 2.0], log  & $-$ & [0.01, 1.0], log & [0.05, 1.0], log \\
\noalign{\vspace{2.5pt}}
temperature      & [60.0, 140.0]  & [60.0, 140.0]  & [0.0, 250.0] & [20.0, 140.0] & [20, 150], step=1 & [20.0, 140.0] \\
\midrule
\multicolumn{7}{l}{Additional variables} \\
\midrule
\noalign{\vspace{2.5pt}}
solvent                & $-$  & $-$  & \{dmf, dmso, nmp, thf, acetonitrile\} & \{dmf, dmso, nmp\}     & \{ dimethylformamide, dimethylsulfoxide, acetonitrile\}        & \{dmf, dmso, acetonitrile, nmp\} \\
\noalign{\vspace{2.5pt}}
base\_equivalents      & $-$  & $-$  & [0.5, 5.0]          & $-$  & $-$     & [0.5, 3.0] \\
\noalign{\vspace{2.5pt}}
base\_type             & $-$  & $-$  & \{k2co3, cs2co3, tea, naoh, koh\}     & \{k2co3, cs2co3, dipea\} & \{potassium carbonate, cesium carbonate, triethylamine\} & \{k2co3, cs2co3, dipea\} \\
\noalign{\vspace{2.5pt}}
nucleophile \newline equivalents & $-$ & $-$ & $-$ & [1.0, 3.0] & [1.0, 3.0] & [0.8, 2.0] \\
\noalign{\vspace{2.5pt}}
reaction\_time\_min    & $-$  & $-$  & $-$                 & $-$  & [15, 480], step=5         & $-$ \\
\noalign{\vspace{2.5pt}}
reaction\_time\_h      & $-$  & $-$  & $-$                 & $-$  & $-$     & [0.5, 24.0] \\
\midrule
Best obj. ($\downarrow$)  & $-$  & 0.1988 {\scriptsize $\pm$ 0.0056}  & 0.9060 {\scriptsize $\pm$ 0.0145}  & 1.5634 {\scriptsize $\pm$ 0.7547}  & 0.7980 {\scriptsize $\pm$ 0.0067} & 0.7843 {\scriptsize $\pm$ 0.0073} \\
\bottomrule
\end{tabularx}}
\end{table}
 
\begin{table}[htbp]
\centering
\caption{Comparison of search space designed by human and LLMs, and the best achieved objective value for Task 17 (Star-shaped crash box, MECHBench) in \textit{Hard} difficulty and \textit{Small} budget. The symbol ``$-$'' indicates the variable was not included in the search space provided by the model.}
\label{tab:search_space_starcrashbox_app}
\small
{\setlength{\tabcolsep}{4pt}
\begin{tabularx}{\linewidth}{
  >{\raggedright\arraybackslash}p{2.8cm}
  >{\raggedright\arraybackslash}X
  >{\raggedright\arraybackslash}X
  >{\raggedright\arraybackslash}X
  >{\raggedright\arraybackslash}X
  >{\raggedright\arraybackslash}X
  >{\raggedright\arraybackslash}X}
\toprule
Variables & Canonical Domain & Human Reference & Gemini 3 Flash & Gemini 3.1 Pro & GPT-5 mini & GPT-5.2 \\
\midrule
width         & [60, 120]   & [60, 120]   & $-$  & $-$  & $-$  & $-$ \\
depth         & [60, 120]   & [60, 120]   & $-$  & $-$  & $-$  & $-$ \\
notch\_width  & [0, 30]     & 15          & $-$  & $-$  & $-$  & $-$ \\
notch\_depth  & [0, 30]     & 15          & $-$  & $-$  & $-$  & $-$ \\
\noalign{\vspace{2.5pt}}
thickness     & [0.7, 3.0]  & [1.1, 2.4]  & [1.0, 3.0]  & [1.0, 5.0]  & [5e-4, 2e-2]  & [0.8, 4.0], log \\
\midrule
\multicolumn{7}{l}{Additional variables} \\
\midrule
inner\_radius       & $-$  & $-$  & [20.0, 40.0]   & [20.0, 60.0]   & [1e-2, 0.2]       & [0.35, 0.85] \\
outer\_radius       & $-$  & $-$  & [41.0, 60.0]   & [65.0, 120.0]  & [5e-2, 0.3]       & [25.0, 75.0] \\
corner\_radius      & $-$  & $-$  & [2.0, 10.0]    & $-$            & $-$               & [0.5, 8.0] \\
number\_of\_corners & $-$  & $-$  & 4              & $-$            & $-$               & $-$ \\
length              & $-$  & $-$  & $-$            & 200.0          & $-$               & $-$ \\
num\_points         & $-$  & $-$  & $-$            & $-$            & [3, 12], step=1   & [4, 10], step=1 \\
\noalign{\vspace{2.5pt}}
crash\_box\_height  & $-$  & $-$  & $-$            & $-$            & $-$               & [80.0, 220.0] \\
inner\_radius\_ratio & $-$ & $-$  & $-$            & $-$            & $-$               & [0.35, 0.85] \\
taper\_angle\_deg   & $-$  & $-$  & $-$            & $-$            & $-$               & [0.0, 6.0] \\
\midrule
Best obj. ($\downarrow$)  & $-$  & \mbox{-11513.2551} {\scriptsize $\pm$ 534.0972}  & -8528.0561 {\scriptsize $\pm$ 6.8648}  & -8471.5581 {\scriptsize $\pm$ 98.1745}  & $\infty$  & -8525.2024 {\scriptsize $\pm$ 15.6285} \\
\bottomrule
\end{tabularx}}
\end{table}

\begin{table}[htbp]
\centering
\caption{Comparison of search space designed by human and LLMs, and the best achieved objective value for Task 19 (Inverted Pendulum in MuJoCo) in \textit{Hard} difficulty and \textit{Small} budget. The symbol ``$-$'' indicates the variable was not included in the search space provided by the model.}
\label{tab:search_space_pendulum_hard_app}
\small
{\setlength{\tabcolsep}{4pt}
\begin{tabularx}{\linewidth}{
  >{\raggedright\arraybackslash}p{2.8cm}
  >{\raggedright\arraybackslash}X
  >{\raggedright\arraybackslash}X
  >{\raggedright\arraybackslash}X
  >{\raggedright\arraybackslash}X
  >{\raggedright\arraybackslash}X
  >{\raggedright\arraybackslash}X}
\toprule
Variables & Canonical Domain & Human Reference & Gemini 3 Flash & Gemini 3.1 Pro & GPT-5 mini & GPT-5.2 \\
\midrule
position\_to\_force  & [$-$1000, 1000] & 0.0 & [$-$2.0, 2.0] \newline {\scriptsize (x\_weight)} & [$-$10.0, 10.0] \newline {\scriptsize (w\_1)} & $-$ & [$-$5.0, 5.0] \newline {\scriptsize (w\_position)} \\
\noalign{\vspace{2.5pt}}
angle\_to\_force     & [$-$1000, 1000] & [$-$150.0, 150.0] & [$-$2.0, 2.0] \newline {\scriptsize (theta\_weight)} & [$-$10.0, 10.0] \newline {\scriptsize (w\_2)} & $-$ & [$-$10.0, 10.0] \newline {\scriptsize (w\_angle)} \\
\noalign{\vspace{2.5pt}}
velocity\_to\_force  & [$-$1000, 1000] & 0.0  & [$-$2.0, 2.0] \newline {\scriptsize (x\_dot\_weight)} & [$-$10.0, 10.0] \newline {\scriptsize (w\_3)} & $-$ & [$-$5.0, 5.0] \newline {\scriptsize (w\_velocity)} \\
\noalign{\vspace{2.5pt}}
angular\_velocity\_ \newline to\_force & [$-$1000, 1000] & [$-$30.0, 30.0] & $-$ & $-$ & $-$ & [$-$5.0, 5.0] \newline {\scriptsize (w\_angular\_\allowbreak velocity)} \\
\noalign{\vspace{2.5pt}}
bias\_to\_force      & [$-$1000, 1000] & 0.0  & $-$ & [$-$10.0, 10.0] {\scriptsize (w\_0)} \newline [$-$10.0, 10.0] {\scriptsize (bias)} & $-$ & [$-$2.0, 2.0] \newline {\scriptsize (bias)} \\
\midrule
\multicolumn{7}{l}{Additional variables} \\
\midrule
theta\_dot\_weight & $-$ & $-$ & [$-$2.0, 2.0] & $-$ & $-$ & $-$ \\
\noalign{\vspace{2.5pt}}
weight\_$i$\_mag \newline ($i=0,\dots,3$) & $-$ & $-$ & $-$ & $-$ & [0.0001, 1.0], log & $-$ \\
\noalign{\vspace{2.5pt}}
weight\_$i$\_sign \newline ($i=0,\dots,3$) & $-$ & $-$ & $-$ & $-$ & bool & $-$ \\
\noalign{\vspace{2.5pt}}
bias\_mag          & $-$ & $-$ & $-$ & $-$ & [0.0001, 1.0], log & $-$ \\
\noalign{\vspace{2.5pt}}
bias\_sign         & $-$ & $-$ & $-$ & $-$ & bool & $-$ \\
\midrule
Best obj. ($\uparrow$)  & $-$  & 577.4 {\scriptsize $\pm$ 370.1}  & 191.4 {\scriptsize $\pm$ 149.5}  & $-\infty$  & 24.0 {\scriptsize $\pm$ 0.0} & 1000.0 {\scriptsize $\pm$ 0.0} \\
\bottomrule
\end{tabularx}}
\end{table}

\begin{table}[htbp]
\centering
\caption{Comparison of search space designed by human and LLMs, and the best achieved objective value for Task 19 (Inverted Pendulum in MuJoCo) in \textit{Easy} difficulty and \textit{Small} budget.}
\label{tab:search_space_pendulum_easy_app}
\small
{\setlength{\tabcolsep}{4pt}
\begin{tabularx}{\linewidth}{
  >{\raggedright\arraybackslash}p{2.8cm}
  >{\raggedright\arraybackslash}X
  >{\raggedright\arraybackslash}X
  >{\raggedright\arraybackslash}X
  >{\raggedright\arraybackslash}X
  >{\raggedright\arraybackslash}X
  >{\raggedright\arraybackslash}X}
\toprule
Variables & Canonical Domain & Human Reference & Gemini 3 Flash & Gemini 3.1 Pro & GPT-5 mini & GPT-5.2 \\
\midrule
position\_to\_force  & [$-$1000, 1000] & 0.0      & [$-$50.0, 50.0] & [$-$100.0, 100.0] & [$-$50.0, 50.0] & [$-$300.0, 300.0] \\
\noalign{\vspace{2.5pt}}
angle\_to\_force     & [$-$1000, 1000] & [$-$150.0, 150.0] & [0.0, 500.0]    & [$-$100.0, 100.0] & [$-$50.0, 50.0] & [$-$300.0, 300.0] \\
\noalign{\vspace{2.5pt}}
velocity\_to\_force  & [$-$1000, 1000] & 0.0      & [$-$50.0, 50.0] & [$-$100.0, 100.0] & [$-$50.0, 50.0] & [$-$300.0, 300.0] \\
\noalign{\vspace{2.5pt}}
angular\_velocity\_ \newline to\_force & [$-$1000, 1000] & [$-$30.0, 30.0] & [0.0, 100.0] & [$-$100.0, 100.0] & [$-$50.0, 50.0] & [$-$300.0, 300.0] \\
\noalign{\vspace{2.5pt}}
bias\_to\_force      & [$-$1000, 1000] & 0.0      & [$-$10.0, 10.0] & 0.0        & [$-$10.0, 10.0] & [$-$3.0, 3.0] \\
\midrule
Best obj. ($\uparrow$)  & $-$  & 577.4 {\scriptsize $\pm$ 370.1}  & 1000.0 {\scriptsize $\pm$ 0.0}  & 427.4 {\scriptsize $\pm$ 511.6}  & 421.9 {\scriptsize $\pm$ 501.2} & 180.6 {\scriptsize $\pm$ 7.7} \\
\bottomrule
\end{tabularx}}
\end{table}

\begin{table}[htbp]
\centering
\caption{Comparison of search space designed by human and LLMs, and the best achieved objective value for Task 20 (Inverted Double Pendulum, MuJoCo) in \textit{Hard} difficulty and \textit{Small} budget. The symbol ``$-$'' indicates the variable was not included in the search space provided by the model.}
\label{tab:search_space_double_pendulum_hard_app}
\small
{\setlength{\tabcolsep}{4pt}
\begin{tabularx}{\linewidth}{
  >{\raggedright\arraybackslash}p{3cm}
  >{\raggedright\arraybackslash}X
  >{\raggedright\arraybackslash}X
  >{\raggedright\arraybackslash}X
  >{\raggedright\arraybackslash}X
  >{\raggedright\arraybackslash}X
  >{\raggedright\arraybackslash}X}
\toprule
Variables & Canonical Domain & Human Reference & Gemini 3 Flash & Gemini 3.1 Pro & GPT-5 mini & GPT-5.2 \\
\midrule
position\_to\_force           & [-1e3, 1e3]  & 0  & [-50, 0]  & $-$  & [0, 10]  & $-$ \\
sin\_pole1\_to\_force         & [-1e3, 1e3]  & [-10, 10]  & $-$  & $-$  & $-$  & $-$ \\
sin\_pole2\_to\_force         & [-1e3, 1e3]  & [-10, 10]  & $-$  & $-$  & $-$  & $-$ \\
cos\_pole1\_to\_force         & [-1e3, 1e3]  & 0  & $-$  & $-$  & $-$  & $-$ \\
cos\_pole2\_to\_force         & [-1e3, 1e3]  & 0  & $-$  & $-$  & $-$  & $-$ \\
velocity\_to\_force           & [-1e3, 1e3]  & 0  & [-20, 0]  & $-$  & $-$  & $-$ \\
\noalign{\vspace{2.5pt}}
angular\_velocity\newline pole1\_to\_force  & [-1e3, 1e3]  & [-50, 50]  & $-$  & $-$  & $-$  & $-$ \\
\noalign{\vspace{2.5pt}}
angular\_velocity\newline pole2\_to\_force  & [-1e3, 1e3]  & [-50, 50]  & $-$  & $-$  & $-$  & $-$ \\
\noalign{\vspace{2.5pt}}
constraint\_force\newline \_to\_force  & [-1e3, 1e3]  & 0  & $-$  & $-$  & $-$  & $-$ \\
bias\_to\_force               & [-1e3, 1e3]  & 0  & $-$  & $-$  & [-10, 10]  & [-2, 2] \\
\midrule
\multicolumn{7}{l}{Additional variables} \\
\midrule
k\_pole1\_ang  & $-$  & $-$  & [10, 300]    & $-$  & $-$  & $-$ \\
k\_pole1\_vel  & $-$  & $-$  & [2, 100]     & $-$  & $-$  & $-$ \\
k\_pole2\_ang  & $-$  & $-$  & [10, 300]    & $-$  & $-$  & $-$ \\
k\_pole2\_vel  & $-$  & $-$  & [2, 100]     & $-$  & $-$  & $-$ \\
w0             & $-$  & $-$  & $-$  & [-10, 10]  & $-$  & $-$ \\
w1             & $-$  & $-$  & $-$  & [-10, 10]  & $-$  & $-$ \\
w2             & $-$  & $-$  & $-$  & [-10, 10]  & $-$  & $-$ \\
w3             & $-$  & $-$  & $-$  & [-10, 10]  & $-$  & $-$ \\
w4             & $-$  & $-$  & $-$  & [-10, 10]  & $-$  & $-$ \\
w5             & $-$  & $-$  & $-$  & [-10, 10]  & $-$  & $-$ \\
w6             & $-$  & $-$  & $-$  & [-10, 10]  & $-$  & $-$ \\
w7             & $-$  & $-$  & $-$  & [-10, 10]  & $-$  & $-$ \\
w8             & $-$  & $-$  & $-$  & [-10, 10]  & $-$  & $-$ \\
w9             & $-$  & $-$  & $-$  & [-10, 10]  & $-$  & $-$ \\
w10            & $-$  & $-$  & $-$  & [-10, 10]  & $-$  & $-$ \\
kp\_pole1      & $-$  & $-$  & $-$  & $-$  & [0, 50]  & $-$ \\
kp\_pole2      & $-$  & $-$  & $-$  & $-$  & [0, 50]  & $-$ \\
kd\_angles     & $-$  & $-$  & $-$  & $-$  & [0, 20]  & $-$ \\
w\_cart\_position    & $-$  & $-$  & $-$  & $-$  & $-$  & [-5, 5] \\
w\_cart\_velocity    & $-$  & $-$  & $-$  & $-$  & $-$  & [-5, 5] \\
w\_pole\_1\_angle    & $-$  & $-$  & $-$  & $-$  & $-$  & [-10, 10] \\
w\_pole\_1 \newline angular\_velocity & $-$  & $-$  & $-$  & $-$  & $-$  & [-5, 5] \\
w\_pole\_2\_angle    & $-$  & $-$  & $-$  & $-$  & $-$  & [-10, 10] \\
w\_pole\_2\_angular \newline velocity & $-$  & $-$  & $-$  & $-$  & $-$  & [-5, 5] \\
\midrule
Best obj. ($\uparrow$)  & $-$  & 139.9769 {\scriptsize $\pm$ 52.6278}  & 101.2238 {\scriptsize $\pm$ 5.0037}  & $-\infty$  & 93.8434 {\scriptsize $\pm$ 1.0817}  & 90.4095 {\scriptsize $\pm$ 0.0000} \\
\bottomrule
\end{tabularx}}
\end{table}

\begin{table}[htbp]
\centering
\caption{Comparison of search space designed by human and LLMs, and the best achieved objective value for Task 20 (Inverted Double Pendulum, MuJoCo, maximization) in \textit{Easy} difficulty and \textit{Small} budget. The symbol ``$-$'' indicates the variable was not included in the search space provided by the model.}
\label{tab:search_space_double_pendulum_easy_app}
\small
{\setlength{\tabcolsep}{4pt}
\begin{tabularx}{\linewidth}{
  >{\raggedright\arraybackslash}p{3cm}
  >{\raggedright\arraybackslash}X
  >{\raggedright\arraybackslash}X
  >{\raggedright\arraybackslash}X
  >{\raggedright\arraybackslash}X
  >{\raggedright\arraybackslash}X
  >{\raggedright\arraybackslash}X}
\toprule
Variables & Canonical Domain & Human Reference & Gemini 3 Flash & Gemini 3.1 Pro & GPT-5 mini & GPT-5.2 \\
\midrule
position\_to\_force           & [-1e3, 1e3]  & 0   & [-5e2, 5e2]  & [-50, 50]  & [-2e2, 2e2]  & [-1e2, 1e2] \\
sin\_pole1\_to\_force         & [-1e3, 1e3]  & [-10, 10] & [-1e3, 1e3]  & [-50, 50]  & [-2e2, 2e2]  & [-1e2, 1e2] \\
sin\_pole2\_to\_force         & [-1e3, 1e3]  & [-10, 10] & [-1e3, 1e3]  & [-50, 50]  & [-2e2, 2e2]  & [-1e2, 1e2] \\
cos\_pole1\_to\_force         & [-1e3, 1e3]  & 0   & [-5e2, 5e2]  &      0     & [-2e2, 2e2]  & [-1e2, 1e2] \\
cos\_pole2\_to\_force         & [-1e3, 1e3]  & 0   & [-5e2, 5e2]  &      0     & [-2e2, 2e2]  & [-1e2, 1e2] \\
velocity\_to\_force           & [-1e3, 1e3]  & 0   & [-2e2, 2e2]  & [-50, 50]  & [-2e2, 2e2]  & [-1e2, 1e2] \\
\noalign{\vspace{2.5pt}}
angular\_velocity\newline pole1\_to\_force  & [-1e3, 1e3]  & [-50, 50]  & [-2e2, 2e2]  & [-50, 50]  & [-2e2, 2e2]  & [-1e2, 1e2] \\
\noalign{\vspace{2.5pt}}
angular\_velocity\newline pole2\_to\_force  & [-1e3, 1e3]  & [-50, 50]  & [-2e2, 2e2]  & [-50, 50]  & [-2e2, 2e2]  & [-1e2, 1e2] \\
\noalign{\vspace{2.5pt}}
constraint\_force\newline \_to\_force  & [-1e3, 1e3]  & 0  & [-10, 10]  & 0  & [-2e2, 2e2]  & [-1e2, 1e2] \\
bias\_to\_force               & [-1e3, 1e3]  & 0  & [-10, 10]  & 0 & [-2e2, 2e2]  & [-50, 50] \\
\midrule
Best obj. ($\uparrow$)  & $-$  & 139.9769 {\scriptsize $\pm$ 52.6278}  & 139.1437 {\scriptsize $\pm$ 14.0981}  & 135.3829 {\scriptsize $\pm$ 15.2964}  & 81.5348 {\scriptsize $\pm$ 6.5960}  & 81.7984 {\scriptsize $\pm$ 2.7418} \\
\bottomrule
\end{tabularx}}
\end{table}

\clearpage

\section{Results of FRS Metric} \label{app:results_frs}
We calculate the FRS for 22 tasks under the \textit{Small} budget. We exclude Task 17 as the evaluations required to construct its reference set take a long time.
For each task, we first compute the FRS for each optimization run; each reported value then represents the mean and standard deviation over three runs.
The complete results under \textit{Easy}, \textit{Intermediate} and \textit{Hard} difficulties are provided in Table \ref{tab:frs_easy_small}, Table \ref{tab:frs_intermediate_small} and Table \ref{tab:frs_hard_small}, respectively.

\begin{table}[htbp]
\centering
\small
\caption{Comparison of FRS between different models in \textit{Easy} difficulty and \textit{Small} budget. Values are mean ± standard deviation over three runs.}
\label{tab:frs_easy_small}
\begin{tabular}{lccccc}
\toprule
Domain & \# Task & Gemini 3 Flash & Gemini 3.1 Pro & GPT-5 mini & GPT-5.2 \\
\midrule
\multirow{7}{*}{YAHPO} & 1 & 0.984 {\scriptsize $\pm$ 0.022} & \textbf{0.985} {\scriptsize $\pm$ 0.006} & 0.955 {\scriptsize $\pm$ 0.070} & 0.787 {\scriptsize $\pm$ 0.019} \\
 & 2 & 0.969 {\scriptsize $\pm$ 0.030} & 0.969 {\scriptsize $\pm$ 0.030} & 0.969 {\scriptsize $\pm$ 0.030} & \textbf{0.981} {\scriptsize $\pm$ 0.020} \\
 & 3 & 0.998 {\scriptsize $\pm$ 0.003} & 0.815 {\scriptsize $\pm$ 0.032} & \textbf{1.000} {\scriptsize $\pm$ 0.000} & \textbf{1.000} {\scriptsize $\pm$ 0.000} \\
 & 4 & 0.999 {\scriptsize $\pm$ 0.001} & 0.997 {\scriptsize $\pm$ 0.005} & 0.998 {\scriptsize $\pm$ 0.002} & \textbf{1.000} {\scriptsize $\pm$ 0.001} \\
 & 5 & \textbf{0.988} {\scriptsize $\pm$ 0.000} & 0.623 {\scriptsize $\pm$ 0.146} & 0.962 {\scriptsize $\pm$ 0.026} & 0.784 {\scriptsize $\pm$ 0.147} \\
 & 6 & 0.989 {\scriptsize $\pm$ 0.019} & 0.968 {\scriptsize $\pm$ 0.045} & \textbf{1.000} {\scriptsize $\pm$ 0.000} & 0.986 {\scriptsize $\pm$ 0.014} \\
 & 7 & 0.971 {\scriptsize $\pm$ 0.006} & 0.417 {\scriptsize $\pm$ 0.025} & 0.924 {\scriptsize $\pm$ 0.034} & \textbf{0.975} {\scriptsize $\pm$ 0.005} \\
\midrule
\multirow{9}{*}{Olympus} & 8 & \textbf{0.998} {\scriptsize $\pm$ 0.000} & \textbf{0.998} {\scriptsize $\pm$ 0.001} & 0.997 {\scriptsize $\pm$ 0.002} & \textbf{0.998} {\scriptsize $\pm$ 0.000} \\
 & 9 & \textbf{1.000} {\scriptsize $\pm$ 0.000} & \textbf{1.000} {\scriptsize $\pm$ 0.000} & 0.958 {\scriptsize $\pm$ 0.058} & 0.907 {\scriptsize $\pm$ 0.006} \\
 & 10 & 0.999 {\scriptsize $\pm$ 0.002} & \textbf{1.000} {\scriptsize $\pm$ 0.000} & 0.993 {\scriptsize $\pm$ 0.004} & 0.990 {\scriptsize $\pm$ 0.005} \\
 & 11 & 0.982 {\scriptsize $\pm$ 0.009} & \textbf{0.990} {\scriptsize $\pm$ 0.007} & 0.873 {\scriptsize $\pm$ 0.019} & 0.983 {\scriptsize $\pm$ 0.020} \\
 & 12 & \textbf{1.000} {\scriptsize $\pm$ 0.000} & \textbf{1.000} {\scriptsize $\pm$ 0.000} & 0.620 {\scriptsize $\pm$ 0.135} & 0.886 {\scriptsize $\pm$ 0.197} \\
 & 13 & \textbf{1.000} {\scriptsize $\pm$ 0.000} & \textbf{1.000} {\scriptsize $\pm$ 0.000} & \textbf{1.000} {\scriptsize $\pm$ 0.000} & \textbf{1.000} {\scriptsize $\pm$ 0.000} \\
 & 14 & \textbf{1.000} {\scriptsize $\pm$ 0.000} & \textbf{1.000} {\scriptsize $\pm$ 0.000} & 0.996 {\scriptsize $\pm$ 0.003} & \textbf{1.000} {\scriptsize $\pm$ 0.000} \\
 & 15 & \textbf{0.989} {\scriptsize $\pm$ 0.016} & 0.983 {\scriptsize $\pm$ 0.018} & 0.908 {\scriptsize $\pm$ 0.025} & 0.902 {\scriptsize $\pm$ 0.025} \\
 & 16 & \textbf{1.000} {\scriptsize $\pm$ 0.000} & \textbf{1.000} {\scriptsize $\pm$ 0.000} & 0.000 {\scriptsize $\pm$ 0.000} & \textbf{1.000} {\scriptsize $\pm$ 0.000} \\
\midrule
MECHBench & 18 & 0.676 {\scriptsize $\pm$ 0.041} & \textbf{0.897} {\scriptsize $\pm$ 0.070} & 0.614 {\scriptsize $\pm$ 0.011} & \textbf{0.897} {\scriptsize $\pm$ 0.070} \\
\midrule
\multirow{5}{*}{MuJoCo} & 19 & \textbf{1.000} {\scriptsize $\pm$ 0.000} & 0.693 {\scriptsize $\pm$ 0.409} & 0.839 {\scriptsize $\pm$ 0.143} & 0.813 {\scriptsize $\pm$ 0.007} \\
 & 20 & \textbf{0.898} {\scriptsize $\pm$ 0.032} & 0.886 {\scriptsize $\pm$ 0.041} & 0.518 {\scriptsize $\pm$ 0.063} & 0.519 {\scriptsize $\pm$ 0.029} \\
 & 21 & \textbf{0.871} {\scriptsize $\pm$ 0.066} & 0.832 {\scriptsize $\pm$ 0.086} & 0.864 {\scriptsize $\pm$ 0.165} & 0.862 {\scriptsize $\pm$ 0.007} \\
 & 22 & 0.445 {\scriptsize $\pm$ 0.012} & 0.404 {\scriptsize $\pm$ 0.024} & \textbf{0.751} {\scriptsize $\pm$ 0.000} & 0.631 {\scriptsize $\pm$ 0.064} \\
 & 23 & 0.933 {\scriptsize $\pm$ 0.014} & 0.716 {\scriptsize $\pm$ 0.283} & 0.933 {\scriptsize $\pm$ 0.014} & \textbf{0.969} {\scriptsize $\pm$ 0.001} \\
\midrule
\multicolumn{2}{c}{Average} & \textbf{0.940} {\scriptsize $\pm$ 0.003} & 0.871 {\scriptsize $\pm$ 0.038} & 0.849 {\scriptsize $\pm$ 0.011} & 0.903 {\scriptsize $\pm$ 0.018} \\
\bottomrule
\end{tabular}
\end{table}

\begin{table}[htbp]
\centering
\small
\caption{Comparison of FRS between different models in \textit{Intermediate} difficulty and \textit{Small} budget. Values are mean ± standard deviation over three runs.}
\label{tab:frs_intermediate_small}
\begin{tabular}{lccccc}
\toprule
Domain & \# Task & Gemini 3 Flash & Gemini 3.1 Pro & GPT-5 mini & GPT-5.2 \\
\midrule
\multirow{7}{*}{YAHPO} & 1 & \textbf{0.867} {\scriptsize $\pm$ 0.057} & 0.417 {\scriptsize $\pm$ 0.014} & 0.503 {\scriptsize $\pm$ 0.139} & 0.742 {\scriptsize $\pm$ 0.049} \\
 & 2 & 0.919 {\scriptsize $\pm$ 0.045} & \textbf{0.996} {\scriptsize $\pm$ 0.004} & 0.646 {\scriptsize $\pm$ 0.051} & 0.272 {\scriptsize $\pm$ 0.000} \\
 & 3 & 0.740 {\scriptsize $\pm$ 0.206} & \textbf{0.929} {\scriptsize $\pm$ 0.008} & 0.916 {\scriptsize $\pm$ 0.009} & 0.249 {\scriptsize $\pm$ 0.082} \\
 & 4 & 0.966 {\scriptsize $\pm$ 0.017} & \textbf{1.000} {\scriptsize $\pm$ 0.000} & 0.990 {\scriptsize $\pm$ 0.016} & 0.996 {\scriptsize $\pm$ 0.001} \\
 & 5 & 0.636 {\scriptsize $\pm$ 0.091} & \textbf{0.679} {\scriptsize $\pm$ 0.020} & 0.577 {\scriptsize $\pm$ 0.090} & 0.578 {\scriptsize $\pm$ 0.090} \\
 & 6 & 0.638 {\scriptsize $\pm$ 0.004} & 0.623 {\scriptsize $\pm$ 0.009} & 0.619 {\scriptsize $\pm$ 0.007} & \textbf{0.721} {\scriptsize $\pm$ 0.079} \\
 & 7 & 0.089 {\scriptsize $\pm$ 0.000} & 0.000 {\scriptsize $\pm$ 0.000} & 0.090 {\scriptsize $\pm$ 0.000} & \textbf{0.403} {\scriptsize $\pm$ 0.005} \\
\midrule
\multirow{9}{*}{Olympus} & 8 & 0.000 {\scriptsize $\pm$ 0.000} & \textbf{0.785} {\scriptsize $\pm$ 0.014} & 0.000 {\scriptsize $\pm$ 0.000} & 0.000 {\scriptsize $\pm$ 0.000} \\
 & 9 & 0.179 {\scriptsize $\pm$ 0.012} & 0.212 {\scriptsize $\pm$ 0.053} & 0.182 {\scriptsize $\pm$ 0.008} & \textbf{0.983} {\scriptsize $\pm$ 0.005} \\
 & 10 & \textbf{0.172} {\scriptsize $\pm$ 0.013} & 0.169 {\scriptsize $\pm$ 0.005} & 0.169 {\scriptsize $\pm$ 0.014} & \textbf{0.172} {\scriptsize $\pm$ 0.005} \\
 & 11 & 0.201 {\scriptsize $\pm$ 0.097} & 0.270 {\scriptsize $\pm$ 0.467} & 0.000 {\scriptsize $\pm$ 0.000} & \textbf{0.416} {\scriptsize $\pm$ 0.036} \\
 & 12 & 0.056 {\scriptsize $\pm$ 0.001} & 0.056 {\scriptsize $\pm$ 0.002} & \textbf{0.057} {\scriptsize $\pm$ 0.003} & \textbf{0.057} {\scriptsize $\pm$ 0.002} \\
 & 13 & \textbf{0.051} {\scriptsize $\pm$ 0.000} & 0.050 {\scriptsize $\pm$ 0.001} & \textbf{0.051} {\scriptsize $\pm$ 0.001} & \textbf{0.051} {\scriptsize $\pm$ 0.001} \\
 & 14 & 0.087 {\scriptsize $\pm$ 0.150} & 0.000 {\scriptsize $\pm$ 0.000} & \textbf{0.577} {\scriptsize $\pm$ 0.040} & 0.443 {\scriptsize $\pm$ 0.009} \\
 & 15 & 0.000 {\scriptsize $\pm$ 0.000} & 0.000 {\scriptsize $\pm$ 0.000} & 0.000 {\scriptsize $\pm$ 0.000} & \textbf{0.400} {\scriptsize $\pm$ 0.003} \\
 & 16 & 0.488 {\scriptsize $\pm$ 0.481} & 0.000 {\scriptsize $\pm$ 0.000} & 0.000 {\scriptsize $\pm$ 0.000} & \textbf{0.885} {\scriptsize $\pm$ 0.053} \\
\midrule
MECHBench & 18 & \textbf{0.139} {\scriptsize $\pm$ 0.000} & \textbf{0.139} {\scriptsize $\pm$ 0.000} & \textbf{0.139} {\scriptsize $\pm$ 0.000} & \textbf{0.139} {\scriptsize $\pm$ 0.000} \\
\midrule
\multirow{5}{*}{MuJoCo} & 19 & 0.863 {\scriptsize $\pm$ 0.011} & \textbf{1.000} {\scriptsize $\pm$ 0.000} & \textbf{1.000} {\scriptsize $\pm$ 0.000} & \textbf{1.000} {\scriptsize $\pm$ 0.000} \\
 & 20 & 0.721 {\scriptsize $\pm$ 0.099} & \textbf{0.735} {\scriptsize $\pm$ 0.022} & 0.583 {\scriptsize $\pm$ 0.053} & 0.471 {\scriptsize $\pm$ 0.062} \\
 & 21 & 0.634 {\scriptsize $\pm$ 0.000} & \textbf{0.944} {\scriptsize $\pm$ 0.015} & 0.634 {\scriptsize $\pm$ 0.000} & 0.634 {\scriptsize $\pm$ 0.000} \\
 & 22 & \textbf{0.751} {\scriptsize $\pm$ 0.000} & \textbf{0.751} {\scriptsize $\pm$ 0.000} & 0.000 {\scriptsize $\pm$ 0.000} & 0.000 {\scriptsize $\pm$ 0.000} \\
 & 23 & \textbf{0.052} {\scriptsize $\pm$ 0.000} & \textbf{0.052} {\scriptsize $\pm$ 0.000} & \textbf{0.052} {\scriptsize $\pm$ 0.000} & \textbf{0.052} {\scriptsize $\pm$ 0.000} \\
\midrule
\multicolumn{2}{c}{Average} & 0.420 {\scriptsize $\pm$ 0.019} & \textbf{0.446} {\scriptsize $\pm$ 0.024} & 0.354 {\scriptsize $\pm$ 0.004} & 0.439 {\scriptsize $\pm$ 0.002} \\
\bottomrule
\end{tabular}
\end{table}

\begin{table}[htbp]
\centering
\small
\caption{Comparison of FRS between different models in \textit{Hard} difficulty and \textit{Small} budget. Values are mean ± standard deviation over three runs.}
\label{tab:frs_hard_small}
\begin{tabular}{lccccc}
\toprule
Domain & \# Task & Gemini 3 Flash & Gemini 3.1 Pro & GPT-5 mini & GPT-5.2 \\
\midrule
\multirow{7}{*}{YAHPO} & 1 & 0.578 {\scriptsize $\pm$ 0.059} & \textbf{0.994} {\scriptsize $\pm$ 0.001} & 0.000 {\scriptsize $\pm$ 0.000} & 0.471 {\scriptsize $\pm$ 0.096} \\
 & 2 & 0.555 {\scriptsize $\pm$ 0.021} & \textbf{0.705} {\scriptsize $\pm$ 0.006} & 0.618 {\scriptsize $\pm$ 0.012} & 0.301 {\scriptsize $\pm$ 0.038} \\
 & 3 & \textbf{0.910} {\scriptsize $\pm$ 0.018} & 0.375 {\scriptsize $\pm$ 0.016} & 0.553 {\scriptsize $\pm$ 0.002} & 0.344 {\scriptsize $\pm$ 0.032} \\
 & 4 & 0.990 {\scriptsize $\pm$ 0.010} & 0.998 {\scriptsize $\pm$ 0.003} & \textbf{1.000} {\scriptsize $\pm$ 0.000} & \textbf{1.000} {\scriptsize $\pm$ 0.000} \\
 & 5 & 0.697 {\scriptsize $\pm$ 0.004} & 0.661 {\scriptsize $\pm$ 0.107} & 0.353 {\scriptsize $\pm$ 0.219} & \textbf{0.703} {\scriptsize $\pm$ 0.003} \\
 & 6 & 0.913 {\scriptsize $\pm$ 0.009} & 0.906 {\scriptsize $\pm$ 0.117} & 0.919 {\scriptsize $\pm$ 0.011} & \textbf{0.927} {\scriptsize $\pm$ 0.066} \\
 & 7 & 0.089 {\scriptsize $\pm$ 0.000} & \textbf{0.159} {\scriptsize $\pm$ 0.002} & 0.081 {\scriptsize $\pm$ 0.000} & 0.000 {\scriptsize $\pm$ 0.000} \\
\midrule
\multirow{9}{*}{Olympus} & 8 & \textbf{0.000} {\scriptsize $\pm$ 0.000} & \textbf{0.000} {\scriptsize $\pm$ 0.000} & \textbf{0.000} {\scriptsize $\pm$ 0.000} & \textbf{0.000} {\scriptsize $\pm$ 0.000} \\
 & 9 & \textbf{1.000} {\scriptsize $\pm$ 0.000} & \textbf{1.000} {\scriptsize $\pm$ 0.000} & 0.990 {\scriptsize $\pm$ 0.013} & 0.994 {\scriptsize $\pm$ 0.009} \\
 & 10 & 0.173 {\scriptsize $\pm$ 0.012} & \textbf{1.000} {\scriptsize $\pm$ 0.000} & 0.745 {\scriptsize $\pm$ 0.008} & 0.685 {\scriptsize $\pm$ 0.023} \\
 & 11 & 0.666 {\scriptsize $\pm$ 0.047} & \textbf{0.972} {\scriptsize $\pm$ 0.030} & 0.886 {\scriptsize $\pm$ 0.051} & 0.119 {\scriptsize $\pm$ 0.001} \\
 & 12 & 0.056 {\scriptsize $\pm$ 0.001} & \textbf{0.057} {\scriptsize $\pm$ 0.003} & 0.056 {\scriptsize $\pm$ 0.002} & \textbf{0.057} {\scriptsize $\pm$ 0.002} \\
 & 13 & \textbf{0.051} {\scriptsize $\pm$ 0.000} & 0.050 {\scriptsize $\pm$ 0.001} & 0.050 {\scriptsize $\pm$ 0.002} & 0.050 {\scriptsize $\pm$ 0.001} \\
 & 14 & 0.469 {\scriptsize $\pm$ 0.013} & 0.219 {\scriptsize $\pm$ 0.322} & 0.561 {\scriptsize $\pm$ 0.008} & \textbf{0.576} {\scriptsize $\pm$ 0.007} \\
 & 15 & 0.000 {\scriptsize $\pm$ 0.000} & 0.000 {\scriptsize $\pm$ 0.000} & 0.000 {\scriptsize $\pm$ 0.000} & \textbf{0.407} {\scriptsize $\pm$ 0.008} \\
 & 16 & \textbf{0.583} {\scriptsize $\pm$ 0.124} & 0.000 {\scriptsize $\pm$ 0.000} & 0.000 {\scriptsize $\pm$ 0.000} & 0.000 {\scriptsize $\pm$ 0.000} \\
\midrule
MECHBench & 18 & \textbf{0.139} {\scriptsize $\pm$ 0.000} & \textbf{0.139} {\scriptsize $\pm$ 0.000} & \textbf{0.139} {\scriptsize $\pm$ 0.000} & \textbf{0.139} {\scriptsize $\pm$ 0.000} \\
\midrule
\multirow{5}{*}{MuJoCo} & 19 & 0.769 {\scriptsize $\pm$ 0.092} & 0.000 {\scriptsize $\pm$ 0.000} & 0.341 {\scriptsize $\pm$ 0.000} & \textbf{1.000} {\scriptsize $\pm$ 0.000} \\
 & 20 & \textbf{0.719} {\scriptsize $\pm$ 0.043} & 0.000 {\scriptsize $\pm$ 0.000} & 0.640 {\scriptsize $\pm$ 0.012} & 0.607 {\scriptsize $\pm$ 0.000} \\
 & 21 & \textbf{0.634} {\scriptsize $\pm$ 0.000} & \textbf{0.634} {\scriptsize $\pm$ 0.000} & \textbf{0.634} {\scriptsize $\pm$ 0.000} & \textbf{0.634} {\scriptsize $\pm$ 0.000} \\
 & 22 & \textbf{0.751} {\scriptsize $\pm$ 0.000} & \textbf{0.751} {\scriptsize $\pm$ 0.000} & \textbf{0.751} {\scriptsize $\pm$ 0.000} & \textbf{0.751} {\scriptsize $\pm$ 0.000} \\
 & 23 & 0.052 {\scriptsize $\pm$ 0.000} & \textbf{0.115} {\scriptsize $\pm$ 0.058} & 0.052 {\scriptsize $\pm$ 0.000} & 0.052 {\scriptsize $\pm$ 0.000} \\
\midrule
\multicolumn{2}{c}{Average} & \textbf{0.491} {\scriptsize $\pm$ 0.010} & 0.443 {\scriptsize $\pm$ 0.018} & 0.426 {\scriptsize $\pm$ 0.013} & 0.446 {\scriptsize $\pm$ 0.008} \\
\bottomrule
\end{tabular}
\end{table}

\clearpage

\section{Results of Search Space Quality} \label{app:results_sample_space_quality}

\begin{figure}[htbp]
    \centering
    \includegraphics[width=\linewidth]{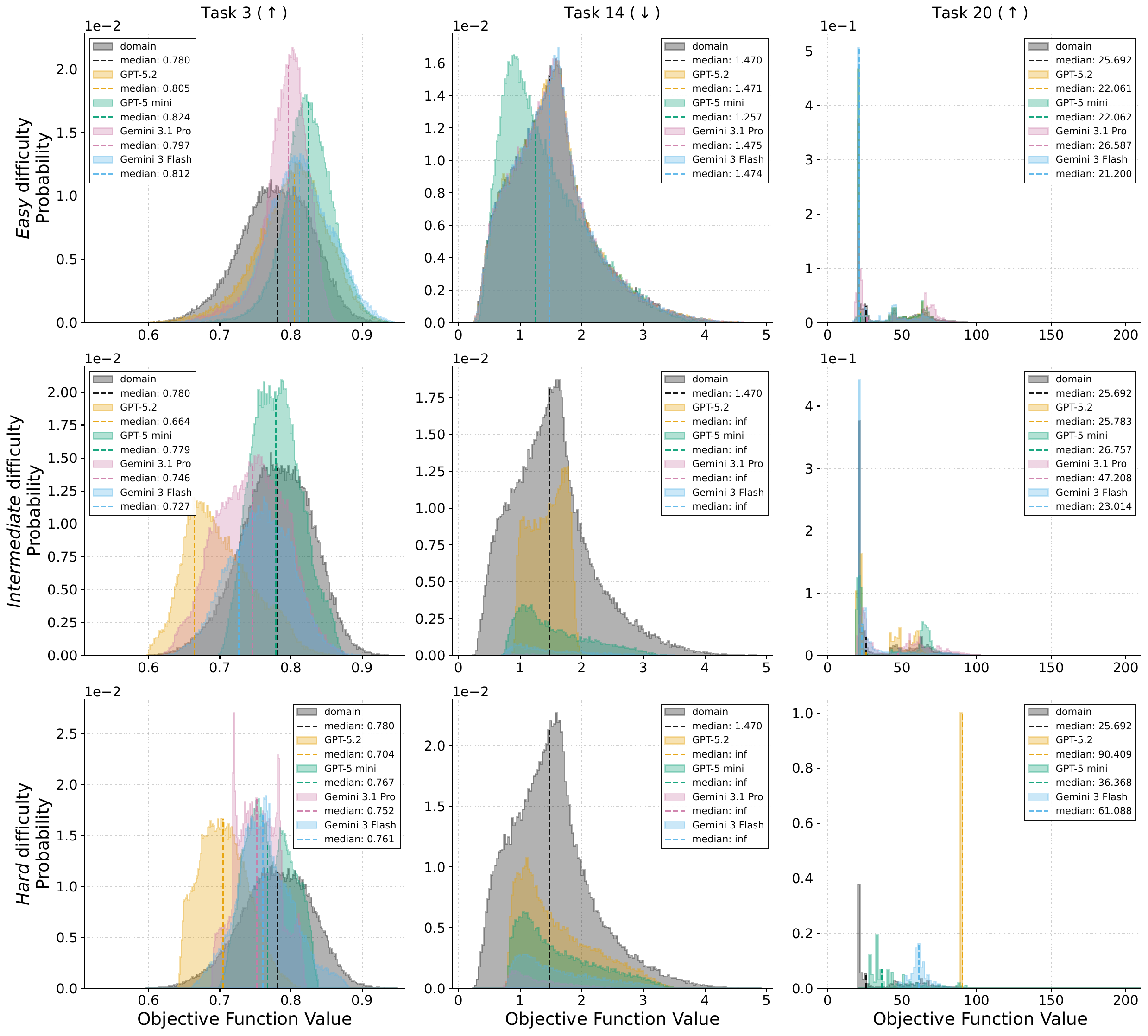}
    \caption{EPDF of objective function values obtained by uniformly sampling 100,000 points from the canonical domain (gray) versus the search spaces generated by four LLMs. This result is the case of zero-shot prompting. The worst objective function value is omitted in the figure.}
    \label{fig:zeroshot_search_space_hist}
\end{figure}

\begin{figure}[htbp]
    \centering
    \includegraphics[width=\linewidth]{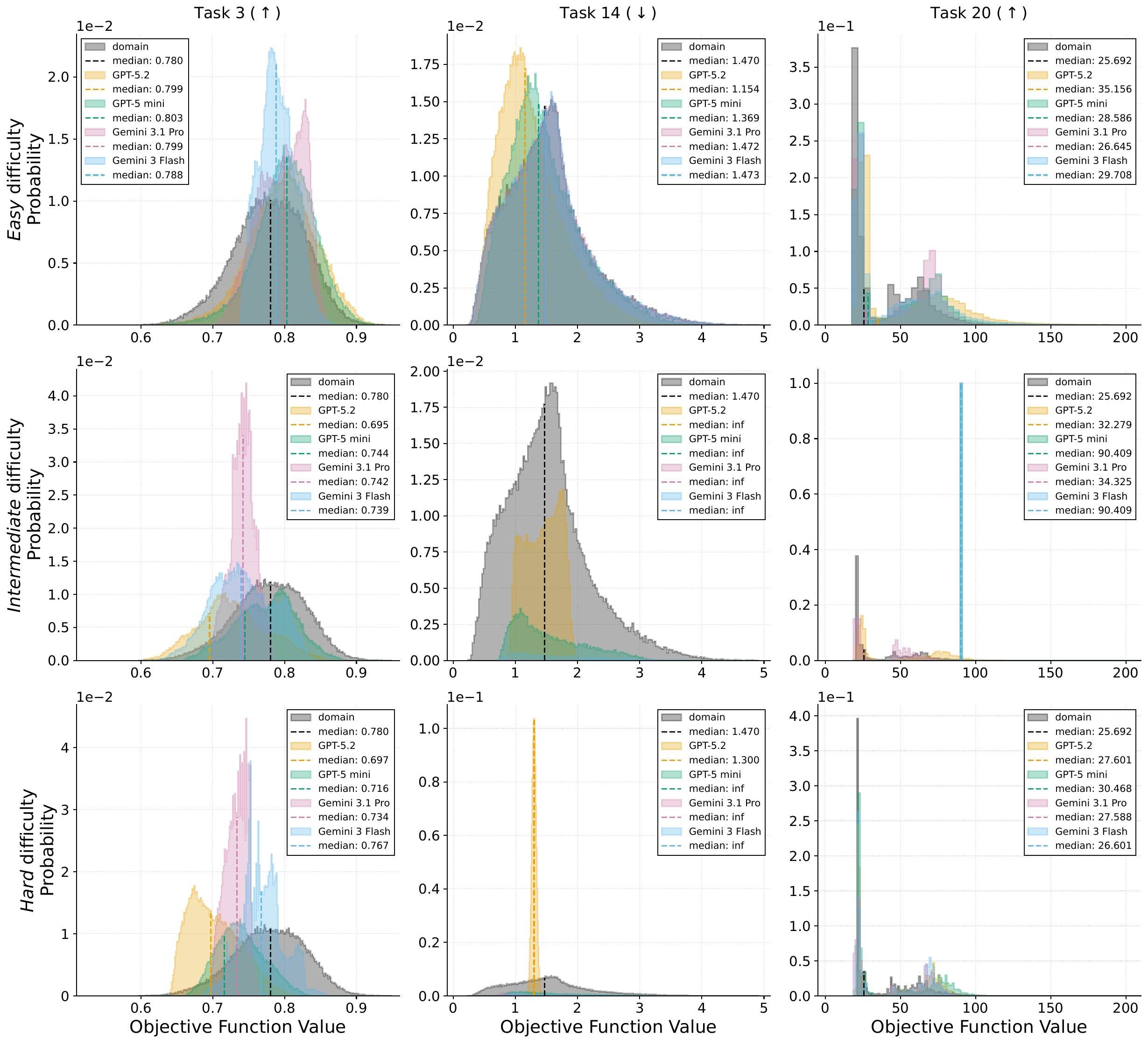}
    \caption{EPDF of objective function values obtained by uniformly sampling 100,000 points from the canonical domain (gray) versus the search spaces generated by four LLMs. This result is the case of one-shot prompting. The worst objective function value is omitted in the figure.}
    \label{fig:oneshot_search_space_hist}
\end{figure}

Figures \ref{fig:zeroshot_search_space_hist} and \ref{fig:oneshot_search_space_hist} visualize how the objective function values are distributed when sampling 100,000 points from the canonical domain of the task and search spaces designed by the LLMs with zero-shot and one-shot prompting, respectively.

In both zero-shot and one-shot settings, for Task 3 in \textit{Easy} difficulty, the median values of the samples from the search spaces designed by LLMs are better than those from the canonical domains.
Conversely, in \textit{Intermediate} and \textit{Hard} difficulty, where more implicit descriptions are given, the median values for the search spaces by LLMs deteriorate even with one-shot prompting.

Comparing zero-shot and one-shot settings, one-shot prompting improves the median values for most models in \textit{Easy} and \textit{Intermediate} difficulties for Task 20, which shares its domain with the one-shot example.
However, the effectiveness of one-shot prompting is limited as seen in \textit{Hard} difficulty for Task 20 and all difficulties for Tasks 3 and 14.

\clearpage

\section{Ablation of Search Space and Algorithm}
\label{app:ss_algo_ablation}

\begin{table}[t]
\centering
\small
\caption{2$\times$2 ablation of search space (SS) and algorithm (Algo) (zero-shot, \textit{Small} budget). $\text{FRS}_\text{LLM}$ is shown for the setting where the LLM designs both, and each $\Delta$ is the change from it (fixed $-$ $\text{FRS}_\text{LLM}$): SS is fixed to the canonical domain, and Algo is fixed to CatCMA with margin. A negative $\Delta$ means that the LLM's choice contributed. Each value is averaged over 3 optimization runs of a single LLM generation. Fixing both axes does not depend on the LLM output, so $\Delta_{\mathrm{Both}}$ uses the Easy runs for every difficulty.}
\label{tab:ss_algo_ablation_delta}
\begin{tabular}{lllrrrr}
\toprule
Model & Task & Diff. & $\text{FRS}_\text{LLM}$ & $\Delta_{\mathrm{SS}}$ & $\Delta_{\mathrm{Algo}}$ & $\Delta_{\mathrm{Both}}$ \\
\midrule
\multirow{6}{*}[-1.3ex]{Gemini 3 Flash} & \multirow{2}{*}{Task 3} & Easy & 0.998 & $-$0.007 & $+$0.002 & $+$0.002 \\
 &  & Hard & 0.910 & $+$0.082 & $+$0.013 & $+$0.090 \\
\cmidrule(l){2-7}
 & \multirow{2}{*}{Task 14} & Easy & 1.000 & 0.000 & 0.000 & 0.000 \\
 &  & Hard & 0.469 & $+$0.531 & $+$0.105 & $+$0.531 \\
\cmidrule(l){2-7}
 & \multirow{2}{*}{Task 20} & Easy & 0.898 & $-$0.386 & $-$0.134 & $-$0.447 \\
 &  & Hard & 0.719 & $-$0.206 & $-$0.266 & $-$0.267 \\
\midrule
\multirow{6}{*}[-1.3ex]{Gemini 3.1 Pro} & \multirow{2}{*}{Task 3} & Easy & 0.815 & $+$0.178 & $-$0.016 & $+$0.185 \\
 &  & Hard & 0.375 & $+$0.618 & $-$0.004 & $+$0.625 \\
\cmidrule(l){2-7}
 & \multirow{2}{*}{Task 14} & Easy & 1.000 & 0.000 & 0.000 & 0.000 \\
 &  & Hard & 0.219 & $+$0.781 & $-$0.193 & $+$0.781 \\
\cmidrule(l){2-7}
 & \multirow{2}{*}{Task 20} & Easy & 0.886 & $-$0.374 & $-$0.035 & $-$0.435 \\
 &  & Hard & 0.000 & $+$0.638 & 0.000 & $+$0.452 \\
\midrule
\multirow{6}{*}[-1.3ex]{GPT-5 mini} & \multirow{2}{*}{Task 3} & Easy & 1.000 & $-$0.008 & 0.000 & 0.000 \\
 &  & Hard & 0.553 & $+$0.439 & $-$0.002 & $+$0.447 \\
\cmidrule(l){2-7}
 & \multirow{2}{*}{Task 14} & Easy & 0.996 & $+$0.002 & $+$0.004 & $+$0.004 \\
 &  & Hard & 0.561 & $+$0.439 & $-$0.016 & $+$0.439 \\
\cmidrule(l){2-7}
 & \multirow{2}{*}{Task 20} & Easy & 0.518 & $-$0.005 & $-$0.051 & $-$0.066 \\
 &  & Hard & 0.640 & $-$0.128 & $-$0.018 & $-$0.189 \\
\midrule
\multirow{6}{*}[-1.3ex]{GPT-5.2} & \multirow{2}{*}{Task 3} & Easy & 1.000 & $-$0.009 & 0.000 & 0.000 \\
 &  & Hard & 0.344 & $+$0.647 & $-$0.119 & $+$0.656 \\
\cmidrule(l){2-7}
 & \multirow{2}{*}{Task 14} & Easy & 1.000 & 0.000 & 0.000 & 0.000 \\
 &  & Hard & 0.576 & $+$0.423 & $-$0.002 & $+$0.424 \\
\cmidrule(l){2-7}
 & \multirow{2}{*}{Task 20} & Easy & 0.519 & $-$0.007 & $+$0.099 & $-$0.068 \\
 &  & Hard & 0.607 & $-$0.094 & 0.000 & $-$0.155 \\
\bottomrule
\end{tabular}
\end{table}

We conduct a 2$\times$2 ablation to separate the contributions of the search space and the algorithm. 
On Tasks 3, 14, and 20 under the zero-shot setting with the \textit{Small} budget, we take the configuration generated by each LLM for the \textit{Easy} and Hard descriptions and replace (i) its search space with the canonical domain, (ii) its algorithm with CatCMAwM, or (iii) both. 
We choose CatCMAwM because it is generally less suited to the \textit{Small} budget.
FRS is averaged over three optimization runs, and \cref{tab:ss_algo_ablation_delta} reports its change from the original configuration, $\Delta_{\mathrm{SS}}$, $\Delta_{\mathrm{Algo}}$, and $\Delta_{\mathrm{Both}}$ for variants (i)--(iii).

In the Easy setting of Tasks 3 and 14, most original configurations already attain FRS close to 1, and all changes are within $0.009$.
The exception is Gemini 3.1 Pro on Task 3: replacing its search space increases FRS by $0.178$, whereas replacing its algorithm changes it by only $0.016$.
In the Hard setting of Tasks 3 and 14, replacing the search space increases FRS by $0.082$--$0.781$, which exceeds the change from replacing the algorithm for all four LLMs.
The performance degradation under the Hard descriptions thus stems mainly from the generated search spaces.
In contrast, for Task 20, $\Delta_{\mathrm{SS}}$ and $\Delta_{\mathrm{Algo}}$ are each positive in only one of the eight cases.
It shows that the generated search spaces outperform the canonical domain, and replacing the algorithm also lowers FRS in most cases.

\clearpage

\section{Analysis of Generation Variance}
\label{app:genvar}

To analyze the effect of variability in LLM outputs, we performed three independent generations for three selected tasks.
For each generated formulation, we averaged the FRS over the three optimization runs.
Each reported value represents the mean and standard deviation of these averaged FRS values across the three generated formulations.
The result is shown in \cref{tab:frs_genvar_overall_small}.
In the \textit{Easy} setting, the FRS is generally stable across generations, particularly for Tasks 3 and 14.
In contrast, the variability in FRS across generations tends to be greater in the \textit{Intermediate} and \textit{Hard} settings.
For example, GPT-5.2 exhibits standard deviations of 0.287 and 0.280 for Task 3 in the \textit{Intermediate} and \textit{Hard} settings, respectively, compared with 0.0 in the \textit{Easy} setting.
This suggests that the quality of generated formulations can be sensitive to LLM generation particularly when problem descriptions are less informative.
Further analysis with more generations and tasks is left for future work.

\begin{table}[htbp]
\centering
\small
\caption{FRS across three independent LLM generations for three selected tasks. For each generation, the FRS is averaged over three optimization runs, and each reported value shows the mean and standard deviation across the three generations.}
\label{tab:frs_genvar_overall_small}
\begin{tabular}{clcccc}
\toprule
Difficulty & Task & Gemini 3 Flash & Gemini 3.1 Pro & GPT-5 mini & GPT-5.2 \\
\midrule
\multirow{3}{*}{Easy} & Task 3 (YAHPO) & 0.999 {\scriptsize $\pm$ 0.001} & 0.938 {\scriptsize $\pm$ 0.107} & 1.000 {\scriptsize $\pm$ 0.000} & 1.000 {\scriptsize $\pm$ 0.000} \\
& Task 14 (Olympus) & 1.000 {\scriptsize $\pm$ 0.000} & 1.000 {\scriptsize $\pm$ 0.000} & 0.993 {\scriptsize $\pm$ 0.008} & 1.000 {\scriptsize $\pm$ 0.000} \\
& Task 20 (MuJoCo) & 0.913 {\scriptsize $\pm$ 0.025} & 0.830 {\scriptsize $\pm$ 0.060} & 0.586 {\scriptsize $\pm$ 0.118} & 0.536 {\scriptsize $\pm$ 0.029} \\
\midrule
\multirow{3}{*}{Intermediate} & Task 3 (YAHPO) & 0.786 {\scriptsize $\pm$ 0.046} & 0.718 {\scriptsize $\pm$ 0.240} & 0.742 {\scriptsize $\pm$ 0.177} & 0.564 {\scriptsize $\pm$ 0.287} \\
& Task 14 (Olympus) & 0.258 {\scriptsize $\pm$ 0.156} & 0.000 {\scriptsize $\pm$ 0.000} & 0.525 {\scriptsize $\pm$ 0.046} & 0.335 {\scriptsize $\pm$ 0.201} \\
& Task 20 (MuJoCo) & 0.755 {\scriptsize $\pm$ 0.117} & 0.672 {\scriptsize $\pm$ 0.064} & 0.497 {\scriptsize $\pm$ 0.170} & 0.529 {\scriptsize $\pm$ 0.179} \\
\midrule
\multirow{3}{*}{Hard} & Task 3 (YAHPO) & 0.614 {\scriptsize $\pm$ 0.257} & 0.403 {\scriptsize $\pm$ 0.055} & 0.480 {\scriptsize $\pm$ 0.069} & 0.497 {\scriptsize $\pm$ 0.280} \\
& Task 14 (Olympus) & 0.488 {\scriptsize $\pm$ 0.025} & 0.224 {\scriptsize $\pm$ 0.226} & 0.518 {\scriptsize $\pm$ 0.037} & 0.556 {\scriptsize $\pm$ 0.033} \\
& Task 20 (MuJoCo) & 0.725 {\scriptsize $\pm$ 0.043} & 0.202 {\scriptsize $\pm$ 0.350} & 0.700 {\scriptsize $\pm$ 0.062} & 0.727 {\scriptsize $\pm$ 0.170} \\
\bottomrule
\end{tabular}
\end{table}

\clearpage

\section{Analysis of Alias Mapping}
\label{app:alias_mapping}

BBOWP-Bench evaluates a generated search space by mapping its variables to the canonical design space through alias dictionaries.
This section quantifies how the alias mapping affects the evaluation results.

\subsection{Effects on Optimization}\label{app:alias_effects}
We examine every optimization run whose best objective value equals the worst value $\omega_d$ in the zero-shot setting.
We manually classify each of these runs into the following three categories according to its cause:
\textit{Out of bounds}, where the designed ranges extend beyond the canonical design space $\mathcal{X}_d$, either being wider than it or lying entirely outside it, and the evaluated points consequently fall outside $\mathcal{X}_d$;
\textit{Ambiguous mapping}, where two or more designed variables are matched to the same canonical variable (i.e., $|M_i| > 1$ in Eq.~\eqref{eq:wrapper_obj}); and \textit{Type mismatch}, which includes conflicts in variable types due to underspecified problem descriptions.
\footnote{Due to a labeling error in our aggregation script, the ``Ambiguous mapping'' and ``Others'' columns in the rebuttal were swapped (the latter is now ``Type mismatch''); the totals and our conclusions are unchanged.}

Tables~\ref{tab:failure_difficulty} and \ref{tab:failure_domain} report the proportions of the worst-value runs by difficulty of problem description and by domain, respectively, each broken down into the three categories.
By difficulty, the worst-value runs become more frequent as the problem description becomes less informative.
By domain, Olympus shows the highest proportion, mainly due to Out of bounds.
This reflects a mismatch in ranges between canonical and designed variables, which may result from a failure to infer appropriate ranges from the description.
As shown in these tables, although the alias mapping can affect the optimization results, its effect is limited, and most of the worst-value trials originate from poorly designed search spaces.

\begin{table}[htbp]
\centering
\small
\caption{Proportions (\%) of trials resulting in the worst objective value by difficulty of problem description, broken down by cause of failure.}
\label{tab:failure_difficulty}
\begin{tabular}{lcccc}
\toprule
Difficulty & Out of bounds & Ambiguous mapping & Type mismatch & Total\\
\midrule
Easy & 0.6 & 0.0 & 0.0 & 0.6 \\
Intermediate & 11.5 & 1.9 & 0.6 & 14.0 \\
Hard & 5.4 & 3.9 & 7.8 & 17.2 \\
\bottomrule
\end{tabular}

\end{table}

\begin{table}[htbp]
\centering
\small
\caption{Proportions (\%) of trials resulting in the worst objective value by domain, broken down by cause of failure.}
\label{tab:failure_domain}
\begin{tabular}{lcccc}
\toprule
Domain & Out of bounds & Ambiguous mapping & Type mismatch & Total\\
\midrule
YAHPO & 1.5 & 2.2 & 0.0 & 3.7 \\
Olympus & 11.9 & 0.5 & 6.3 & 18.7 \\
MECHBench & 4.2 & 0.0 & 0.0 & 4.2 \\
MuJoCo & 0.0 & 4.2 & 0.0 & 4.2 \\
\bottomrule
\end{tabular}

\end{table}

\subsection{Coverage of Alias Dictionaries}\label{app:alias_coverage}
We further measured how often the designed variables are left unmapped solely because the alias dictionaries lack an appropriate entry.
We manually identified such variables among all 4,474 variables designed by the LLMs in the zero-shot experiments, pooled over the models, tasks, and budgets.
Table~\ref{tab:unmapped_alias} reports the number of these unmapped variables by difficulty of problem description and domain, together with their proportion relative to the total number of designed variables.
Overall, 457 variables were left unmapped for this reason.
By difficulty, the proportion is only 1.1\% for \textit{Easy}, while it reaches 21.3\% for \textit{Intermediate}, most of which comes from MuJoCo (204 variables).
By domain, the proportion is low in MECHBench (1.0\%) and YAHPO (3.7\%) but higher in Olympus (12.8\%) and MuJoCo (16.3\%).

When a designed variable is left unmapped, the corresponding canonical variable is assigned its predefined default value, so the trial still executes and returns a valid objective value.
That is, unmapped variables do not by themselves lead to the worst objective value; instead, the corresponding canonical variables are held fixed rather than optimized, and the reported metrics, including FRS, may be underestimated or overestimated for the configurations in which such variables occur.

\begin{table}[htbp]
\centering
\small
\caption{Number of LLM-designed variables left unmapped due to a missing alias-dictionary entry, by difficulty of problem description and domain. Values in parentheses indicate the proportion relative to the total number of variables designed by the LLMs.}
\label{tab:unmapped_alias}
\begin{tabular}{lccccc}
\toprule
Difficulty & YAHPO & Olympus & MECHBench & MuJoCo & Total \\
\midrule
Easy & 0 & 0 & 0 & 18 & 18 (1.1\%) \\ 
Intermediate & 37 & 67 & 1 & 204 & 309 (21.3\%) \\ 
Hard & 19 & 63 & 2 & 46 & 130 (9.8\%) \\ 
\midrule
Total & 56 (3.7\%) & 130 (12.8\%) & 3 (1.0\%) & 268 (16.3\%) & 457 (10.2\%) \\ 
\bottomrule
\end{tabular}
\end{table}

\subsection{Representative Mapping Cases}\label{app:alias_cases}
Table~\ref{tab:mapping_cases} presents representative examples of designed variables that were mapped to canonical variables and those that were not.
The mapped examples show that the alias dictionaries absorb typical naming variations, such as alternative names (\texttt{hidden\_units} for \texttt{max\_units}) and additional prefixes or suffixes (\texttt{dye\_orange} and \texttt{orange\_ratio}).

The unmapped examples arise from two types of causes.
The first is missing alias entries.
The variables \texttt{mtry\_fraction} and \texttt{warm\_orange\_ratio} were manually judged to be semantically equivalent to the corresponding canonical variables but are not included in the alias dictionaries; such cases can be addressed by extending the dictionaries.
The second is differences in the coordinate system or parameterization.
In Task~17, the LLM describes the cross-section of the crash box with outer and inner radii based on a polar coordinate system, whereas the canonical design space uses the width and depth based on a rectangular coordinate system.
In Task~20, the LLM uses a single gain \texttt{k\_pole1\_ang} for the angle of the first pole, whereas the canonical design space uses separate gains for its sine and cosine.
Such differences cannot be resolved by name matching and are beyond the scope of the current framework; addressing them is left for future work.

\begin{table}[htbp]
\centering
\footnotesize
\caption{Representative examples of designed variables mapped and unmapped to canonical variables. Setting indicates the difficulty, budget, and model under which the design was generated in zero-shot.}
\label{tab:mapping_cases}
\begin{tabular}{lllll}
\toprule
Task & Canonical variable & Designed variable & Setting & Result \\
\midrule
1 & \texttt{max\_units} & \texttt{hidden\_units} & Hard / Large / GPT-5.2 & Mapped \\
\midrule
3 & \texttt{mtry.power} & \texttt{mtry\_fraction} & Hard / Large / GPT-5.2 & Unmapped \\
\midrule
\multirow{5}{*}{9} & \multirow{5}{*}{\texttt{orange}} & \multirow{3}{*}{\texttt{orange\_ratio}}  & Hard / Large / GPT-5 mini & \multirow{3}{*}{Mapped} \\
 & &  & Hard / Large / GPT-5.2 &  \\
 & &  & Hard / Small / Gemini 3 Flash &  \\
 & & \texttt{dye\_orange} & Hard / Small / GPT-5 mini & Mapped \\
 & & \texttt{warm\_orange\_ratio} & Intermediate / Large / GPT-5.2 & Unmapped \\
\midrule
14 & \texttt{temperature} & \texttt{temperature\_celsius} & Hard / Large / Gemini 3 Flash & Mapped \\
\midrule
\multirow{2}{*}{17} & \multirow{2}{*}{\texttt{width}, \texttt{depth}} & \texttt{outer\_radius} & \multirow{2}{*}{Hard / Small / GPT-5 mini} & \multirow{2}{*}{Unmapped} \\
   & & \texttt{inner\_radius} & \\
\midrule
\multirow{2}{*}{20} & \texttt{sin\_pole1\_to\_force} & \multirow{2}{*}{\texttt{k\_pole1\_ang}} & \multirow{2}{*}{Hard / Large / Gemini 3 Flash} & \multirow{2}{*}{Unmapped} \\
 & \texttt{cos\_pole1\_to\_force} & & & \\
\bottomrule
\end{tabular}
\end{table}

\clearpage
\section{Potential Societal Impact}
\label{app:impact}
We believe that our proposed benchmark can support research on automated formulation for black-box optimization, potentially reducing the expertise required to apply optimization methods. A possible negative societal impact is that more automated optimization could be misused in harmful applications if combined with inappropriate objectives or unsafe evaluation environments. However, this paper introduces a benchmark and does not release high-risk models, scraped datasets, or deployed optimization systems.


\clearpage
\section*{NeurIPS Paper Checklist}

\begin{enumerate}

\item {\bf Claims}
    \item[] Question: Do the main claims made in the abstract and introduction accurately reflect the paper's contributions and scope?
    \item[] Answer: \answerYes{} 
    \item[] Justification: We have summarized the main contributions and scope in the abstract and introduction. Our contributions are (1) introducing Black-Box Optimization Word Problem (BBOWP), (2) proposing BBOWP-Bench to evaluate the output for BBOWP, (3) designing the evaluation protocol and framework on BBOWP, and (4) providing the first benchmark study of LLMs on BBOWP-Bench. The main claim is supported by our proposed benchmark and experimental evaluation described in Sections \ref{sec:bbowp_bench} and \ref{sec:experiment}.

\item {\bf Limitations}
    \item[] Question: Does the paper discuss the limitations of the work performed by the authors?
    \item[] Answer: \answerYes{} 
    \item[] Justification: The limitations of the work are discussed in \cref{sec:limitation}.

\item {\bf Theory assumptions and proofs}
    \item[] Question: For each theoretical result, does the paper provide the full set of assumptions and a complete (and correct) proof?
    \item[] Answer: \answerNA{} 
    \item[] Justification: The paper does not include theoretical results.

    \item {\bf Experimental result reproducibility}
    \item[] Question: Does the paper fully disclose all the information needed to reproduce the main experimental results of the paper to the extent that it affects the main claims and/or conclusions of the paper (regardless of whether the code and data are provided or not)?
    \item[] Answer: \answerYes{} 
    \item[] Justification: We describe our dataset and benchmark in \cref{sec:bbowp_bench} and the experimental setup in \cref{sec:exp_setup}. In addition, we provide more details in the appendix and share all our assets and code.

\item {\bf Open access to data and code}
    \item[] Question: Does the paper provide open access to the data and code, with sufficient instructions to faithfully reproduce the main experimental results, as described in supplemental material?
    \item[] Answer: \answerYes{} 
    \item[] Justification: Our code is available under the MIT License at \url{https://github.com/shiralab/bbowp-bench}. 
Our benchmark and dataset are also released under the Creative Commons Attribution 4.0 International License (CC BY 4.0) at \url{https://github.com/shiralab/bbowp-bench/tree/main/benchmark}.

\item {\bf Experimental setting/details}
    \item[] Question: Does the paper specify all the training and test details (e.g., data splits, hyperparameters, how they were chosen, type of optimizer) necessary to understand the results?
    \item[] Answer: \answerYes{} 
    \item[] Justification: We describe the experimental design and setup in \cref{sec:exp_setup} and \cref{app:input_output}.

\item {\bf Experiment statistical significance}
    \item[] Question: Does the paper report error bars suitably and correctly defined or other appropriate information about the statistical significance of the experiments?
    \item[] Answer: \answerYes{} 
    \item[] Justification: We report the optimization results as the mean and standard deviation over three optimization runs with different random seeds (\cref{sec:results_optimization,app:experiment}). For the FRS, we compute the values for each of the three seeds with a single generated search space and report their mean and standard deviation (\cref{sec:results_metric,app:results_frs}). Owing to API cost considerations, each LLM output is generated once per experimental condition. The variability across LLM generations is analyzed separately for three tasks (\cref{app:genvar}). In addition, several settings under the \textit{Large} budget require more than 48 hours of computation; we terminated such runs.

\item {\bf Experiments compute resources}
    \item[] Question: For each experiment, does the paper provide sufficient information on the computer resources (type of compute workers, memory, time of execution) needed to reproduce the experiments?
    \item[] Answer: \answerYes{} 
    \item[] Justification: We describe the information about our computer resources used in the experiments in \cref{app:resource}.
    
\item {\bf Code of ethics}
    \item[] Question: Does the research conducted in the paper conform, in every respect, with the NeurIPS Code of Ethics \url{https://neurips.cc/public/EthicsGuidelines}?
    \item[] Answer: \answerYes{} 
    \item[] Justification: We have carefully read the NeurIPS Code of Ethics and believe our work conforms to it.

\item {\bf Broader impacts}
    \item[] Question: Does the paper discuss both potential positive societal impacts and negative societal impacts of the work performed?
    \item[] Answer: \answerYes{} 
    \item[] Justification: \cref{app:impact} discusses the potential positive and negative societal impacts of our research. We believe that our proposed benchmark can support research on automated formulation for black-box optimization, which is a positive societal impact of our work. A possible negative societal impact is that more automated optimization could be misused in harmful applications if combined with inappropriate objectives or unsafe evaluation environments. However, this paper introduces a benchmark and does not release high-risk models, scraped datasets, or deployed optimization systems. 
    
\item {\bf Safeguards}
    \item[] Question: Does the paper describe safeguards that have been put in place for responsible release of data or models that have a high risk for misuse (e.g., pre-trained language models, image generators, or scraped datasets)?
    \item[] Answer: \answerNA{} 
    \item[] Justification: We did not train any model and do not release new models. Our benchmark, dataset, and code have no perceivable risk of misuse because they can be used only for evaluating models and systems for BBOWP.

\item {\bf Licenses for existing assets}
    \item[] Question: Are the creators or original owners of assets (e.g., code, data, models), used in the paper, properly credited and are the license and terms of use explicitly mentioned and properly respected?
    \item[] Answer: \answerYes{} 
    \item[] Justification: In \cref{sec:data_description}, \cref{sec:exp_setup}, and \cref{app:code_dataset}, we describe the license of existing assets used in our work and refer to the original papers. We properly respect the license and terms of use for the existing assets.

\item {\bf New assets}
    \item[] Question: Are new assets introduced in the paper well documented and is the documentation provided alongside the assets?
    \item[] Answer: \answerYes{} 
    \item[] Justification: We detail our new dataset asset in \cref{sec:data_description} and the Appendix, and provide the documentation alongside it.
    The licenses and existing assets used to construct BBOWP-Bench are detailed in Appendix~\ref{app:code_dataset}. The benchmark and dataset are released under CC BY 4.0, and the code is released under the MIT License.

\item {\bf Crowdsourcing and research with human subjects}
    \item[] Question: For crowdsourcing experiments and research with human subjects, does the paper include the full text of instructions given to participants and screenshots, if applicable, as well as details about compensation (if any)? 
    \item[] Answer: \answerNA{} 
    \item[] Justification: The paper does not involve crowdsourcing nor research with human subjects.

\item {\bf Institutional review board (IRB) approvals or equivalent for research with human subjects}
    \item[] Question: Does the paper describe potential risks incurred by study participants, whether such risks were disclosed to the subjects, and whether Institutional Review Board (IRB) approvals (or an equivalent approval/review based on the requirements of your country or institution) were obtained?
    \item[] Answer: \answerNA{} 
    \item[] Justification: The paper does not involve crowdsourcing nor research with human subjects.

\item {\bf Declaration of LLM usage}
    \item[] Question: Does the paper describe the usage of LLMs if it is an important, original, or non-standard component of the core methods in this research? Note that if the LLM is used only for writing, editing, or formatting purposes and does \emph{not} impact the core methodology, scientific rigor, or originality of the research, declaration is not required.
    \item[] Answer: \answerYes{} 
    \item[] Justification: We evaluate the performance of the LLMs in our experiments. Detailed information about the models we use is provided in \cref{sec:exp_setup}.

\end{enumerate}

\end{document}